\documentclass[pdflatex,sn-mathphys-num]{sn-jnl}% Math and Physical Sciences Numbered Reference Style
\usepackage{graphicx}%
\usepackage{longtable}
\usepackage{array}
\usepackage{multirow}
\usepackage{rotating}
\usepackage{amsmath,amssymb,amsfonts}%
\usepackage{amsthm}%
\usepackage{mathrsfs}%
\usepackage[title]{appendix}%
\usepackage{xcolor}%
\usepackage{textcomp}%
\usepackage{manyfoot}%
\usepackage{booktabs}%
\usepackage{algorithm}%
\usepackage{algorithmicx}%
\usepackage{algpseudocode}%
\usepackage{listings}%
\usepackage{amssymb}
\usepackage{tikz}
\usepackage{graphicx}
\usepackage{caption}
\usetikzlibrary{positioning,shapes,arrows}
\usepackage{longtable}
\usepackage{lscape}
\usepackage{url}  
\usepackage{array}       % Required for custom column types
\usepackage{longtable}   % Since you're using longtable

\newcolumntype{P}[1]{>{\centering\arraybackslash}p{#1}}
 
\theoremstyle{thmstyleone}%
\theoremstyle{thmstyletwo}%

\theoremstyle{thmstylethree}%

\begin{document}

\title[The Evolution of Natural Language Processing: How Prompt Optimization and Language Models Are Shaping the Future]{The Evolution of Natural Language Processing: How Prompt Optimization and Language Models are Shaping the Future}

\author*[1,2]{\fnm{Summra } \sur{Saleem}}\email{summra.saleem@rptu.de}
\equalcont{These authors contributed equally to this work.}

\author[2]{\fnm{Muhammad Nabeel } \sur{Asim}}\email{muhammad\_nabeel.asim@dfki.de}
\equalcont{These authors contributed equally to this work.}

\author[3]{\fnm{Shaista } \sur{Zulfiqar}}\email{shaistazulfiqar65@gmail.com}

\author[1,2]{\fnm{Andreas} \sur{Dengel}}\email{andreas.dengel@dfki.de}

\affil*[1]{ \orgname{RPTU Rheinland-Pfälzische Technische Universität Kaiserslautern-Landau},  \postcode{67663},\country{Germany}}

\affil[2]{\orgname{German Research Centre for Artificial Intelligence}, \city{Kaiserslautern}, \postcode{67663},  \country{Germany}}

\affil[3]{\orgdiv{Department of Computer Science}, \orgname{University of Management and Technology}, \city{Lahore}, \country{Pakistan}}


\abstract{Large Language Models (LLMs) have revolutionized  the field of Natural Language Processing (NLP) by automating traditional labor-intensive tasks and consequently accelerated the development of computer-aided applications. As researchers continue to advance this field with the introduction of novel language models and more efficient training/finetuning methodologies, the idea of prompt engineering and subsequent optimization strategies with LLMs has emerged as a particularly impactful trend to yield a substantial performance boost across diverse NLP tasks. To best of our knowledge numerous review articles have explored prompt engineering, however, a critical gap exists in comprehensive analyses of prompt optimization strategies. To bridge this gap this paper provides unique and comprehensive insights about the potential of diverse prompt optimization strategies. It analyzes their underlying working paradigms and based on these principles, categorizes them into 11 distinct classes. Moreover, the paper provides details about various NLP tasks where these prompt optimization strategies have been employed, along with details of different LLMs and benchmark datasets used for evaluation. This comprehensive compilation lays a robust foundation for future comparative studies and enables rigorous assessment of prompt optimization and LLM-based predictive pipelines under consistent experimental settings: a critical need in the current landscape. Ultimately, this research will centralize diverse strategic knowledge to facilitate the adaptation of existing prompt optimization strategies for development of innovative predictors across unexplored tasks.}




\maketitle

\section{Introduction}

\begin{figure}[htbp]
    \centering
    \includegraphics[width=0.8\textwidth]{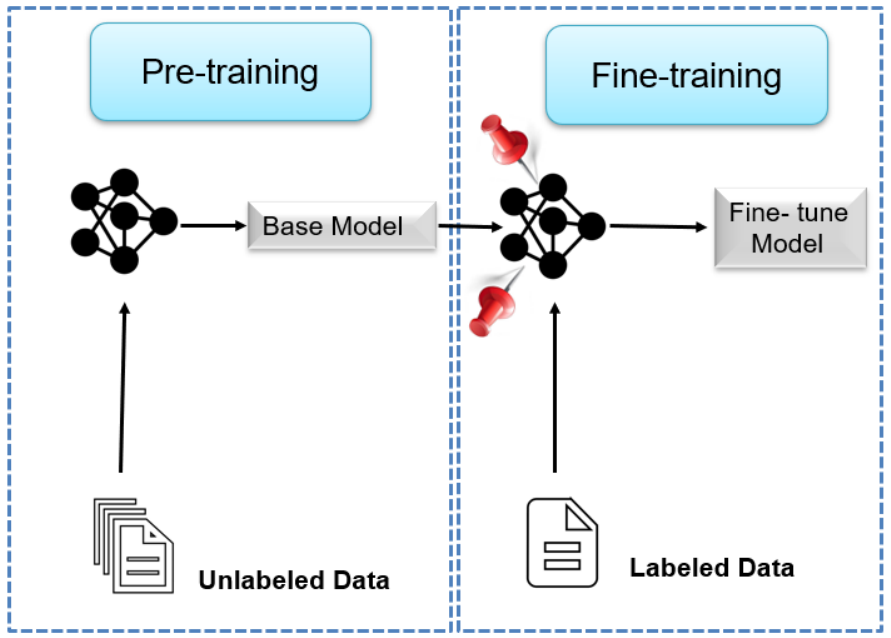}
    \caption{Different Stages of Training Large Language Models}
    \label{LLM-training}
\end{figure}

The emergence of Large language models (LLMs) has 
marked a paradigm shift in the Natural Language Processing (NLP) domain and offered ground-breaking solutions to address complex linguistic tasks such as code generation \cite{fakhoury2024llm, huang2024bias}, machine translation \cite{enis2024llm, elshin2024general}, style transfer \cite{wu2025implementing}, sequence analysis \cite{asim2022lgca}, requirement engineering \cite{saleem2025generative, saleem2024passionnet} etc. However, the remarkable achievements of LLMs are accompanied by certain trade-offs, which stem from two stage training process of LLMs: 1) Pre-training, 2) Fine-tuning (as shown in Figure \ref{LLM-training}). During the pre-training stage, LLMs are exposed to extensive corpora of unlabelled text to learn general linguistic patterns and  semantic relations between different words. The sheer size of LLMs requires substantial computational sources, massive dataset and extremely large time frame for pre-training \cite{raiaan2024review}. For instance, pre-training of well-known BERT model required 16 Google TPUs for 4 full days \cite{devlin2018bert}. Similarly, pre-training of the Megatron-Turing NLG 530B model took 2,000 NVIDIA A100 GPUs for 9.2 consecutive days \cite{nvidiamegatronturing}. On the other hand, during the fine-tuning stage, the pre-trained LLMs are adapted for downstream tasks by training them in a supervised manner on labelled dataset \cite{bonfigli2024pre}. 
 While fine-tuning requires fewer resources than pre-training, it still demands careful dataset curation and parameter optimization to avoid overfitting or degrading the general capabilities learned during pre-training. 
 In addition to financial cost, the environmental damage caused by LLM training also raises critical concerns about the sustainability of current development practices. According to a recent study \cite{forbes_ai_carbon}, the training of a single LLM generates carbon emission (around 600,000 pounds of CO2), which is equivalent to 125 round-trip flights between New York and Beijing. Hence, LLMs are constrained by their dependency on large-scale resources and data, which hinders LLMs' applicability across diverse tasks.

\begin{figure}[htbp]
    \centering
    \includegraphics[width=0.9\textwidth]{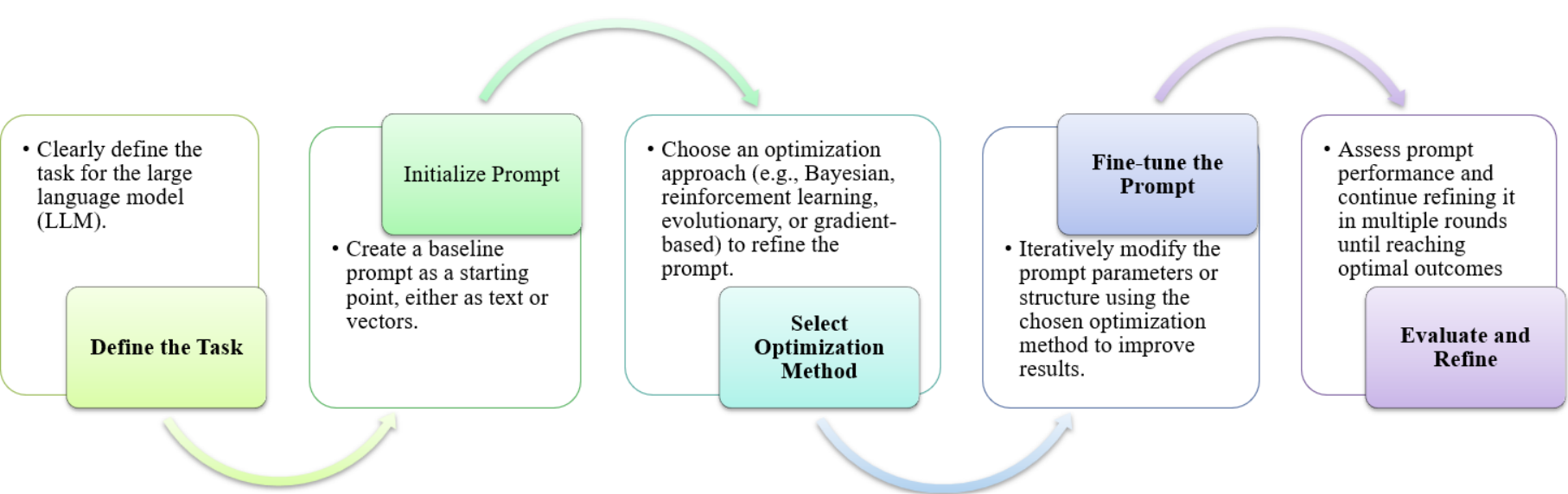}
    \caption{Key stages of prompt optimization}
    \label{stages-of-prompt-optimization}
\end{figure}

 To mitigate computational constraints of LLMs, in 2021  prompt-based LLMs (e.g. GPT-Neo, OPT, Flan-T5) emerged as efficient alternatives. These models provide a flexible approach to leverage vast knowledge encoded in pre-trained LLMs without any need for extensive retraining. Specifically, the prompt is an instruction given to LLM to guide its behaviour to perform a certain task. Although prompt-based models are capable of developing rapid solutions across a diverse array of tasks, however, their performance is heavily dependent on the quality of input prompt. A well-crafted prompt can drastically improve model output, while poorly constructed prompt can severely deteriorate models' performance. This considerable performance contrast based on prompt quality led to the evolution of prompt optimization as a distinct field, which focuses on techniques to craft prompts to achieve optimal performance. This includes techniques such as zero-shot prompting, chain-of-thought prompting, meta prompting, instructional prompting, contextual prompting, etc. However, as the importance of prompt quality became evident, the field of prompt optimization emerged as a natural evolution. While prompt engineering focuses on the initial creation of effective prompts, prompt optimization centres on refining and improving these prompts iteratively. This ensures that the prompts are fine-tuned to elicit the best possible outputs for specific tasks, addressing challenges and performance gaps identified through testing and feedback. Figure \ref{stages-of-prompt-optimization} graphically illustrates key stages of prompt optimization. 

The advent of prompt optimization has opened up new possibilities for improving model performance across a wide range of tasks \cite{chen2023unleashing}. By focusing on enhancing the input rather than modifying the model itself, researchers and practitioners can achieve significant improvements in output quality, task accuracy, and overall model efficiency. In recent years, 45 different prompt optimization strategies have been proposed. These prompts range from simple methods to complex approaches which leverage reinforcement learning, evolutionary algorithms, Bayesian optimization and gradient based back-propagation techniques. Based on structure and interaction with the LLM, prompt optimization strategies can be broadly categorized into two categories: soft prompts and hard prompts \cite{qin2021learning, wen2023hard}. Hard prompts are discrete human-readable instructions given to language models, while soft prompts are learnable continuous vectors that are not interpretable. To date, 3 soft prompts and 42 hard prompt based optimization strategies have been proposed for diverse NLP tasks. However, despite exponential growth the current landscape of prompt optimization research has a significant gap. To the best of our knowledge the field lacks a comprehensive, systematic comparison of existing strategies, which makes it challenging to deeply understand their relative strengths, weaknesses, and optimal use cases across distinct types of tasks. Considering the need, this manuscript encompasses a comprehensive review of 45 existing prompt optimization strategies. Furthermore, this review sheds light on the broader impact of these strategies on the field of NLP to highlighting how these methods have strengthened capabilities of LLMs. 

The rest of the paper is structured as follows. Section 2 outlines the research methodology adopted for this review study. Section 3 presents a journal, conference, and publisher-wise analysis of the distribution of relevant articles. Section 4 provides a structured review of prompt optimization strategies categorized by methodological approach. Section 5 offers an overview of various NLP tasks used to evaluate these techniques. In Section 6, we analyze the application of prompt optimization strategies across different NLP tasks and benchmark datasets. Section 7 evaluates the performance and effectiveness of these techniques across a range of pretrained language models. Finally, Section 8 concludes the paper with a discussion of the findings and key takeaways.
\section{Research Methodology}
This section outlines the methodology used to select research articles focusing on prompt optimization strategies for diverse types of NLP applications. Following the article searching and selection criteria of existing review articles \cite{dalibor2022generating, di2022low}, our research methodology ensures reliability of this paper by carefully selecting the most relevant articles through two  distinct stages: 1) Article identification, 2) Article screening and
filtering. 
\begin{figure*}[htbp]
 \centering 
 \includegraphics[width=1\textwidth]{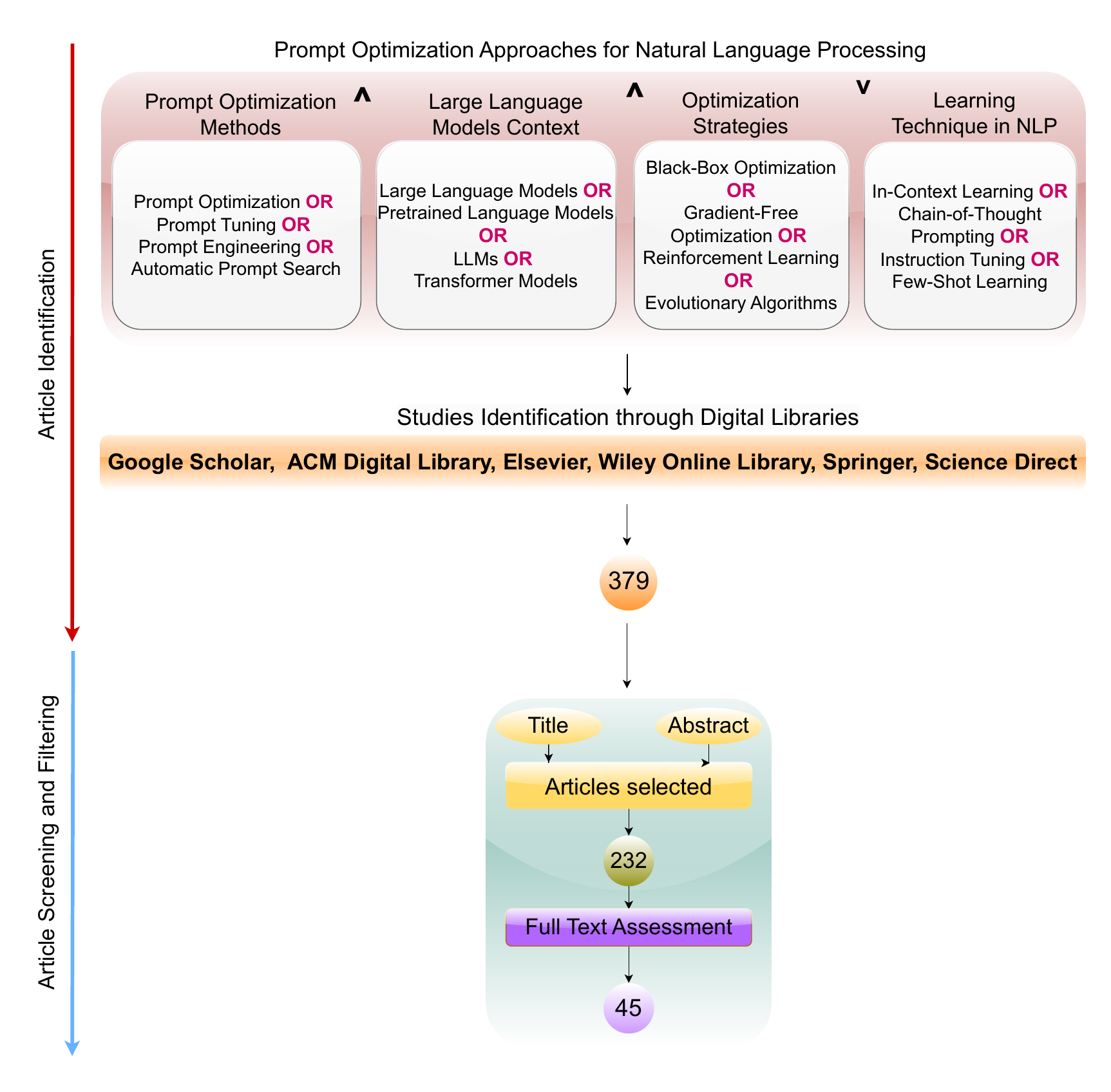}
 \caption{Graphical illustration of two-stage research methodology}
\label{research-methodolgy-pipeline}
\end{figure*}
\subsection{Article Searching} 
In Figure \ref{research-methodolgy-pipeline}, article identification module contains three cells for different kinds of keywords, namely prompt optimization strategies, Language models and NLP tasks. To formulate search queries, keywords within same cell are combined using OR $\vee$ operator while keywords from different cells are combined using AND $\wedge$ operator. For instance, few sample queries include: prompt optimization strategies for pretrained large language model across NLP tasks, prompt design for language model in NLP doamin, etc. To extract search paper, these search queries are executed on academic search engines
such as Google Scholar,  
ScienceDirect and the archive.
ACM Digital Library \footnote{https://dl.acm.org/},
IEEE Xplore \footnote{https://ieeexplore.ieee.org/},
ScienceDirect \footnote{https://www.sciencedirect.com/},
Elsevier\footnote{https://www.elsevier.com/},
Springer\footnote{https://www.springer.com/gp},
Wiley Online Library\footnote{https://www.wiley.com/en-us}. Furthermore, snowballing is employed to identify more research articles by examining the reference list. 
\subsection{Article Screening and Filtering}
Second stage consists of two steps process to select the most relevant articles. In the first step, titles and abstracts of 379 articles were reviewed by domain experts, resulting in the identification of 232 relevant articles. In the second step, a full-text assessment of these articles discards 185 irrelevant research articles and keeps only 45 articles related to prompt optimization strategies for diverse types of NLP tasks for the detailed systematic review.

\section{Journal/Conference and Publisher-wise Analysis of Article Distribution}
\begin{figure*}[htbp]
 \centering 
 \includegraphics[width=1\textwidth]{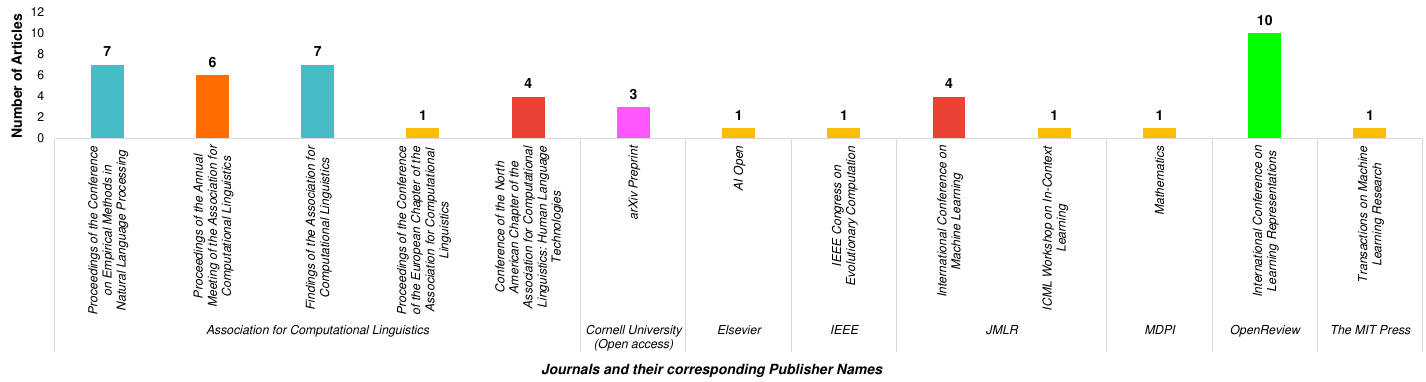}
 \caption{Graphical illustration of article distribution by journal/conference and publisher}
\label{publisher-distribution}
\end{figure*}

This section presents the distribution of prompt optimization literature across different journals, conferences and publishers. The analysis not only helps researchers strategically position their work but also highlights the dynamic and evolving landscape of research for prompt optimization in NLP.
As illustrated in Figure \ref{publisher-distribution}, the reviewed studies span 8 different publishers, including the Association for Computational Linguistics, Cornell University, Elsevier, IEEE, JMLR, MDPI, OpenReview and The MIT Press. Particularly, 25 out of 45 studies have been published by the Association for Computational Linguistics, which highlights its contribution to prompt optimization research. Additionally, OpenReview has contributed significantly, publishing 10 relevant papers in recent years. Overall, these studies have appeared in over 10 distinct journals and conferences, which reflects the interest in this area of NLP research.

 \section{Analysis of Prompt Optimization Strategies Through the Lens of their Working Paradigms}
\begin{figure*}[htbp]
 \centering 
 \includegraphics[width=1\textwidth]{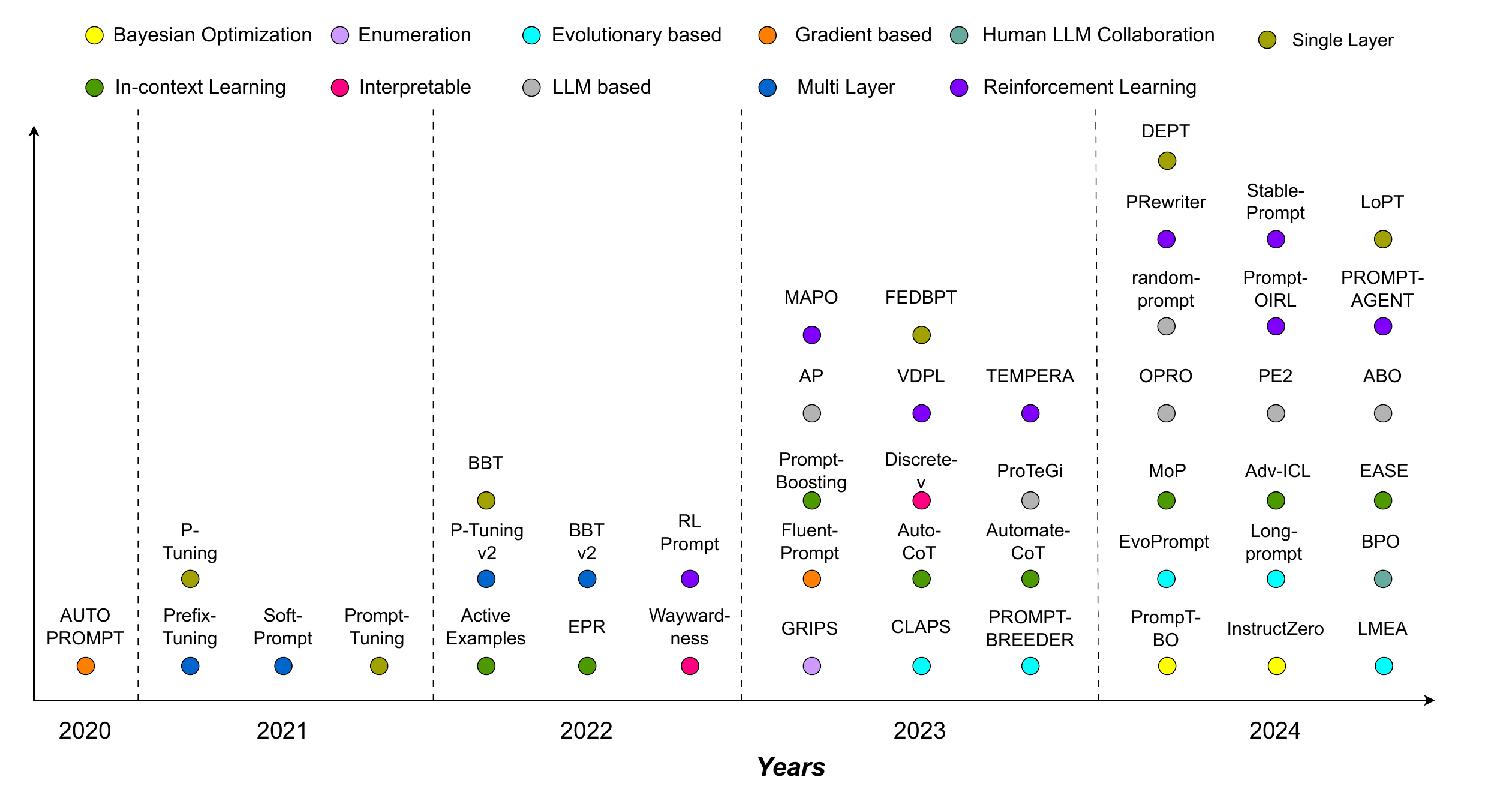}
 \caption{A year-wise look at prompt optimization research from 2020 through 2024, highlighting the field’s rapid innovation and methodological growth}
\label{timeline-of-prompt-optimization}
\end{figure*}

 To provide a comprehensive understanding of diverse strategies used in prompt optimization, this section presents a structured review of recent methods based on their underlying optimization technique. These categories demonstrate the methodological distinctions between different techniques that exhibit continuous prompt embeddings (soft prompting) and those that rely on discrete prompt text (hard prompting). This classification enables a clearer comparison of techniques employed for prompt optimization to reflect the breadth of strategies explored for prompt optimisation within NLP research. Based on underlying algorithmic approaches, prompt optimization strategies can be categorized into 11 distinct types, as shown in Figure \ref{prompt-types}. Furthermore, Figure \ref{timeline-of-prompt-optimization} graphically illustrates the evolution timeline of prompt optimization strategies.
\begin{enumerate}
\begin{figure*}[htbp]
 \centering 
 \includegraphics[width=1\textwidth]{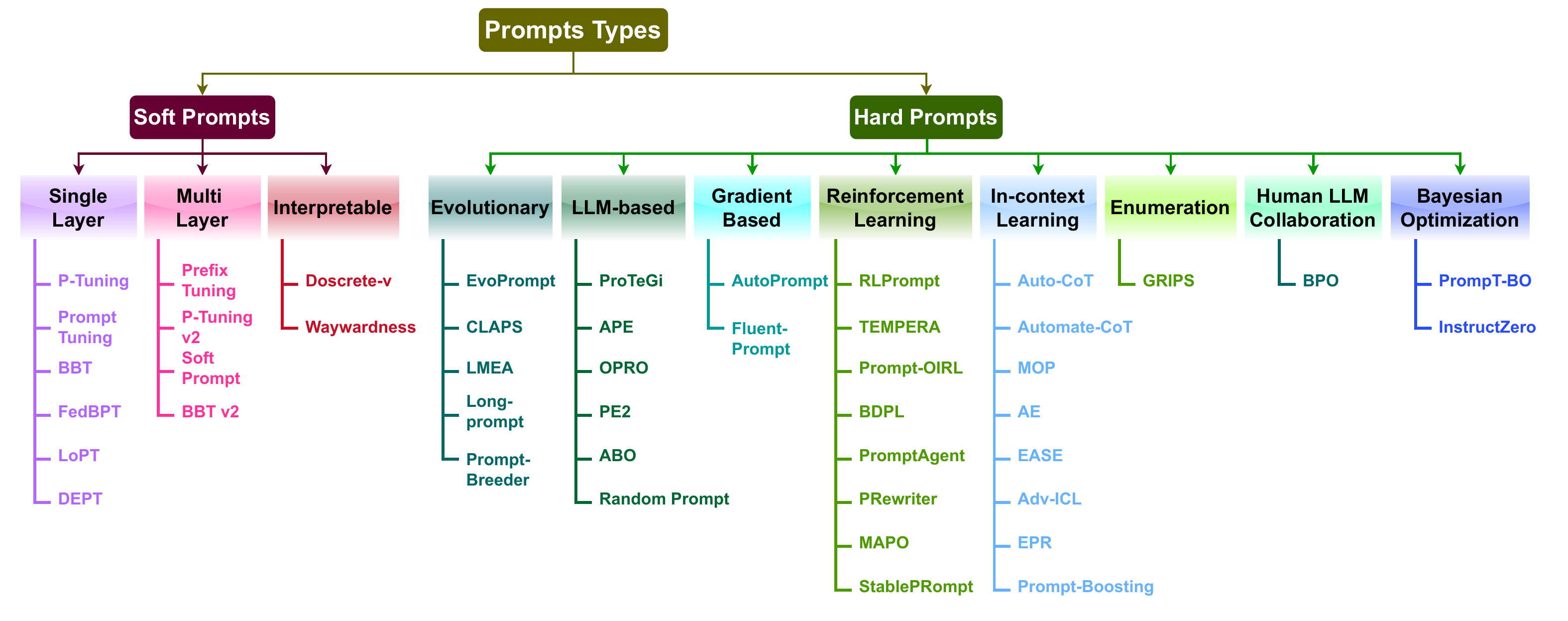}
 \caption{Classification of prompt optimization strategies based on their working paradigm}
\label{prompt-types}
\end{figure*}
\item \textbf{\textit{\underline{Gradient based}}} 
Gradient-based prompt optimization employs textual gradients of the input tokens to optimize discrete textual prompts, without altering the model’s parameters. The first method in this category, \textbf{AutoPrompt} (2020) \cite{shin2020autoprompt}, leverages gradient signals to automatically discover effective textual prompts. Specifically, through iterative gradient evaluation, it selects which input tokens to substitute to boost the model’s performance.  The algorithm iteratively updates a prompt template by identifying tokens from the vocabulary whose gradients maximize task-specific likelihood. Prompt tuning facilitates effective adaptation of pretrained models' weights and eliminates the need to modify the model’s internal parameters. AutoPrompt approach is evaluated on various tasks including sentiment analysis, fact retrieval and natural language inference tasks. In 2023, \textbf{FluentPrompt} \cite{shi2022toward} addressed the key shortcomings of AutoPrompt the inability to generate linguistically coherent prompts. FluentPrompt integrates gradient guidance with linguistic constraints to ensure that the generated prompts are effective and semantically correct.  These approaches demonstrate the potential of textual gradient-based optimization to bridge the gap between automated prompt engineering and human-crafted prompts, particularly in low-resource or few-shot learning settings.\newline

     \item \textbf{\textit{\underline{Single Layer }}}
Single-layer prompt optimization strategies involve the insertion of a learnable prompt representation at the input layer of a pretrained language model. These approaches aim to steer the model’s behavior for downstream tasks by optimizing a small set of task-specific parameters, without updating the core weights of the model. The goal of these methods is to guide the pretrained language model for the downstream task by fine-tuning a limited number of task-specific parameters, without updating main model's weight. \textbf{Prompt-Tuning }introduced in 2021 \cite{lester2021power} stands as the foundation method of single layer prompt optimization strategies. This method learns a continuous, task-aware vector representation (''soft prompts") that is fed to a pre-trained model along with input embeddings of sequence \cite{qin2021learning}. Afterwards, \textbf{P-Tuning} \cite{liu2024gpt} extended this by incorporating a deep prompt module, which enables integration of learnable representation at multiple layers while preserving the core model's weights.
 Later, in 2022, \textbf{Black-Box Tuning} (BBT)  \cite{sun2022black} marked a significant shift by optimizing continuous prompts in black-box manner without any gradient information. Mainly, it leverages evolutionary strategies and bandit algorithms to optimize the prompts layer-by-layer while treating the model as black-box API. This approach adheres to single-layer principle and emphasizes adaptability in closed-source settings. In 2023, \textbf{Efficient Federated Black-box Prompt Tuning} (FedBPT) \cite{sun2023fedbpt} extended the concept of black box optimization to decentralized setting with multiple clients having private data. Specifically, FedBPT incorporates a federated learning framework where multiple clients collaboratively tune prompts using only their local data, without sharing raw data or model parameters
In 2024, \textbf{Decoupled Prompt Tuning} (DEPT)  \cite{shi2023dept} introduced a modular approach to segregate prompts into shared and task-specific components. The shared prompts encompass general knowledge about the task, while the task-specific prompts capture individual task specific features.  Meanwhile, \textbf{Low-rank Prompt Tuning} (LoPT)  \cite{guo2024lopt},  proposed a low-rank factorization of the prompt space to drastically reduce the number of tunable parameters.  Instead of optimizing the full prompt matrix, only smaller matrices are updated during training.\newline

\item \textbf{\textit{\underline{Multi Layer }}} 
Unlike single layer methods that only modify the input embeddings for prompt optimization, multi-layer approaches interact with internal hidden states of the network across various levels. This deeper integration helps more powerful adaptation to complex task-specific patterns. The first method of the multi-layer paradigm is \textbf{PrefixTuning} (2021) \cite{li2021prefix}, which adds trainable prefix vectors to the key and value matrices attention layer of every transformer block. In 2022, \textbf{P-Tuning v2} \cite{liu2021p} advanced this paradigm to enable richer interactions with internal representations by introducing trainable prompts across all layers of the model, not limited only to the attention layer. Also in 2022, \textbf{Black-Box Tuning} (BBTv2) \cite{sun2022bbtv2} extended BBT concept to multi-layer prompt tuning, which allows trainable prompts at multiple transformer layers while supporting black-box optimization.\newline

\item \textbf{\textit{\underline{Interpretable}}}

Interpretable prompt optimization strategies aim not only to enhance model behaviour but also to make the effect of prompts interpretable to humans. Mainly, these methods focus on transparency and explainability by using discrete prompts that allow researchers to trace how distinct parts of the prompt influence model’s behaviour. In contrast to black-box approaches that solely emphasize performance, interpretable methods aids deeper understanding of model's interaction with guided input.
The first interpretable approach is \textbf{Waywardness} (2022) \cite{khashabi-etal-2022-prompt}, which aims to analyze and quantify the ''waywardness” of model i.e how subtle prompt modifications can yield unexpected shifts in a model's outputs.  The method significantly uncovers the prompt segments which lack robustness and offers valuable insights into the model's generalizability. Subsequently, \textbf{DiscreteV} (2023) proposed discrete prompt learning by optimizing over constrained vocabulary through reinforcement learning or search-based techniques. It produces human-readable prompts, which allow users to manually inspect, modify, or analyze. The model aims to achieve robust performance with semantic clarity to ensure that each token of optimized prompt provides intuitive meaning in the context of the task.\newline

 \item \textbf{\textit{\underline{Reinforcement Learning }}} 
   Prompt optimization strategies in this category utilize sequential decision-making process to discover optimal prompts. Primarily, these methods leverage reward signals such as, model performance or feedback quality) to refine prompts. Following the working paradigm of traditional Reinforcement Learning (RL), these techniques also treat prompt generation as a trainable policy. They obtain feedback through interaction with LLMs to guide optimization of prompts towards maximizing sustained reward.
   The first method of this category, \textbf{Reinforcement Learning Prompt} (RLPrompt) was introduced in 2022 \cite{deng2022rlprompt}, which employs  policy gradients to tune discrete prompts but struggles with scalability and gradient issues. To address these challenges, \textbf{Black-box Discrete Prompt Learning } (BDPL) (2023) \cite{diao2023black} advanced RLPrompt to black-box strategy that optimizes discrete prompts based on performance feedback. Later in 2023,  \textbf{Test-Time Prompt Editing via Reinforcement Learning } (TEMPERA) \cite{zhang2022tempera} expanded the field by introducing dynamic prompt modification during inference to enable a shift from static to real-time optimization. In the same year, \textbf{Model Adaptive Prompt Optimizer} (MAPO) \cite{chen2024mapo} incoporated a memory buffer for high-reward prompts to facilitate intelligent exploration of prompt space. In 2024, \textbf{Prompt Rewriter} (PRewriter) \cite{kong2024prewrite} leveraged RL to enable highly specific edits of existing prompts, consequently boosting interpretability and control.  \textbf{StablePRompt} (2024) \cite{kwon2024stableprompt} introduced reward smoothing and regularization strategies to yield robust prompts and deal with instability challenges in RL training. PromptOIRL (2024) \cite{sun2023query} innovated further by applying inverse RL to learn reward functions from high-quality prompt samples. Finally, \textbf{PROMPT-AGENT} (2024) \cite{wang2023promptagent} proposed a general-purpose RL agent that interacts with LLMs and external tools across multiple steps to facilitate adaptive prompt optimization for complex reasoning and decision-making.\newline
   
    \item \textbf{\textit{\underline{Enumeration}}}
 Enumeration-based prompt optimization comprehensively investigates the prompt domain through direct generation, followed by performance evaluation of a diverse collection of discrete prompts. These method are model-agnostic and focus on structured exploration of prompt candidate to discover task effective solution. Notably, this category is exclusively represented by  \textbf{Gradient-free Instructional Prompt Search} (GRIPS)  method \cite{prasad2022grips}, proposed in 2023  which is edit-based prompt optimization strategy and doesn't rely on gradients. Initially, it takes from a human-written instruction as base prompt and subsequently applies a series of heuristic modifications (e.g. insertion, deletion, or word replacement). Afterwards, it computes performance of each variants on a small set of labelled samples. The method continues iteratively and greedily chose the optimal-performing modification at each step until no improvement is detected.\newline
    \item \textbf{\textit{\underline{Evolutionary based  }}}
Evolutionary-based prompt optimization strategies draw inspiration from biological concepts such as mutation, crossover, and natural selection to iteratively refine prompts. Fundamentally, these methods maintain a pool of candidate prompts and improve them over generations by choosing the most effective variants and produce new variants through stochastic alterations. The first method of this category: \textbf{Cl}ustering \textbf{a}nd \textbf{P}runing for Efficient Black-box
Prompt \textbf{S}earch (CLAPS, 2023) \cite{zhou2023survival}, reaps combined benefits of contrastive learning and evolutionary search to enhance. In the same year, \textbf{PROMPTBREEDER} (2023) \cite{fernando2023promptbreeder} adopted traditional genetic algorithms to initialize a diverse population of prompts followed by mutation and crossover operations guided by performance score. Expanding on prior research, EvoPrompt (2024) \cite{guo12connecting} proposed fitness-aware mutation, where prompts alteration are influenced by task-specific score.  Here's a comprehensive rephrasing of your statement, followed by five candidate sentences: Afterwards, LongPrompt (2024) significantly extended the evolutionary optimization paradigm to address the challenges of scenarios requiring rich contextual information or complex chain-of-thought reasoning. Specifically, LongPrompt utilizes hierarchical structures within both the prompt encoding and the evolutionary process to preserve coherence and logical flow. Finally, \textbf{Language Model Evolutionary Algorithm} (LMEA, 2024) \cite{liu2024large} adopted a hybrid strategy which directly integrates LLM as co-optimizer into the evolutionary process to generate, mutate and refine prompts.\newline
    \item \textbf{\textit{\underline{In-context Learning  }}}
   In-context learning based prompt optimization strategies aim to boost a model's task performance by improving the selection or formulation of demonstration examples directly within the input prompt. These methods do not update model’s gradients; instead, they focus on optimization of information presented to the model during inference, which makes them highly adaptable for zero-shot or few-shot tasks. The initial methods of this category, like EPR (2022) \cite{rubin2021learning} and Active Example (2022) \cite{zhang2022active}, emphasize intelligently choosing examples. EPR employs similarity metrics to identify task-relevant instances, while Active Example uses active learning to select the most informative examples for prompts. Later in 2023, Auto-CoT \cite{zhang2022automatic} proposed automatic chain-of-thought generation, in which model is introduced with intermediate reasoning steps in instances to improve task performance. This was extended by Automate-CoT (2023) \cite{shum2023automatic}, which automatically improves and selects the most effective reasoning steps. To further boost performance, Prompt-Boosting (2023) \cite{hou2023promptboosting} proposed to aggregate outputs from multiple ICL prompts. In 2024, MOP (Mixture of Prompting)  \cite{wang2023mixture} emerged as a more advanced method which constructs prompt population and weighs them based on confidence, to mitigate the risk of single prompt dependence. EASE (2024) \cite{wu2024prompt} introduced instance-aware scoring to assess the individual contribution of each instance for prompt optimization. Finally, Adv-ICL (2024) \cite{long2024prompt} investigates robustness by introducing adversarial modification in examples to test prompt stability and improve generalization.\newline

    \item \textbf{\textit{\underline{LLM based}}}
    This category encompasses prompt optimization strategies that leverage the inherent capabilities of LLMs to generate, refine, and evaluate prompts. \textbf{Pr}ompt \textbf{O}ptimization with \textbf{Te}xtual (ProTeGi, 2023) \cite{pryzant2023automatic} leveraged textual gradients to guide prompt refinement using the model's feedback. Automatic Prompt Engineer (APE, 2023) formulates prompt optimizaion as a search task and iteratively employs LLMs to create and score prompts. In 2024 \textbf{ Optimization by PROmpting} (OPRO)  \cite{yang2023large} presented optimization as a prompting task, which enables LLMs to utilize chain-of-thought style reasoning for refined instructions. PE2 (2024) \cite{ye2023prompt} enhances this by systematically incorporating structural feedback to improve prompt quality. \textbf{Automatic Behaviour Optimization} (ABO) (2024) focuses on behaviour optimization to guide the LLM toward desired outputs through self-generated feedback loops. Random Prompt (2024)  \cite{lu2023strings} method demonstrates that LLM scoring-based filtering of randomly generated prompts can yield competitive results. \newline
     \item \textbf{\textit{\underline{Human LLM Collaboration }}}
     This category represents prompt optimization strategies that empower LLMS with human expertise and feedback to refine prompts. These methods bridge the gap between fully automated LLM-based strategies and traditional human prompt engineering by harnessing the combined power of human intellect and LLMs extensive capabilities. \textbf{Bayesian Prompt Optimization} (BPO) proposed in 2024 \cite{cheng2023black}, stands as the sole approach of this category. Initially, humans craft prompt candidates based on their domain knowledge. Then, Bayesian optimization is applied to effectively explores and exploit these prompts. Subsequently, LLM assess the performance of all candidate prompts and results are fed back into the optimization loop. This interactive approach leverages both human intuition and automated search for systematic prompt enhancement.\newline
     \item \textbf{\textit{\underline{Bayesian Optimization}}} 
    This category of prompt optimization strategies employs Bayesian Optimization (BO) to identify optimal prompt configurations. Moreover, BO is a sequential model-based optimization strategy, unlike gradient-based methods that randomly explore prompts. Specifically, it builds a probabilistic model of the objective function (e.g., task performance) which guides the discovery of the next prompt for evaluation using a minimal number of expensive language model evaluations. The pioneering method of this category  \textbf{Prompt-BO} (2024) \cite{sabbatella2024prompt} aims to identify the most effective discrete tokens to append with a query to boost the model's behaviour. Primarily, it utilizes Bayesian optimization in a continuous embedding space and projects the solutions to their discrete prompts.
During the same year \textbf{InstructZero} \cite{chen2023instructzero} was proposed, which optimizes prompts for black-box LLMs using a two-stage process. Firstly, it refines a soft prompt on an open-source model to yield human-readable instructions. Subsequently, these instructions are evaluated on the black-box model in zero-shot manner. This performance feedback guides the Bayesian optimizer to tune the soft prompt.  
\end{enumerate}

\section{Taxonomy and Overview of NLP Tasks for Evaluating Prompt Optimization Strategies}
This section sheds light on nine key NLP tasks used to assess the effectiveness of different prompt optimization strategies.
\subsection{Classification}
Classification is a fundamental task of NLP that involves categorizing input sentences, documents or sequences into predefined classes. Based on the working paradigm, classification can be categorized into 3 types: binary, multi-class, multilable classification. 1) Binary classification assigns inputs to one of two classes, while multi-class classification categorizes inputs into one of several mutually exclusive classes. On the other hand, in multi-label classification, a input can belong to multiple classes simultaneously. The ability to accurately classify input enables automated decision making and information processing across various domains.
The real-world applications of classification are diverse and far-reaching, spanning multiple fields and industries. To ensure online safety and integrity of digital platforms, classification techniques are utilized for hate speech detection and fake news identification. The cybersecurity domain leverages from fake URL detection to protect users from phishing attempts and malicious websites. In the field of bio-informatics, classification plays a crucial role to advance genomic and proteomics research such as DNA modification prediction, enhancer identification and soluble proteins identification. These applications indicates the versatility and significance of classification techniques in addressing complex challenges from social media and journalism to scientific research and public health.

\subsection{Question Answering}
Question answering is the task of processing and understanding the input question to retrieve relevant information from a knowledge source to formulate an appropriate response. Question answering systems rely on difference sources such as structured databases, unstructured document collections, pre-trained knowledge within LLMs to generate answers. In the financial sector, QA systems can assist to provide information about digital billing. In the aviation industry QA systems aid to extract answers from extensive manuals for pilots. Furthermore, QA systems are also employed in customer service chatbots and real-time information retrieval to obtain up-to-date answers about current events and rapidly changing information. 

\subsection{Natural Language Inference}
Natural Language Inference (NLI) is a fundamental task of NLP that determines the logical relationship between two input sentences: a premise and a hypothesis. The goal is to classify whether the hypothesis can be inferred from the premise (entailment), contradicts the premise (contradiction), or is neither entailed nor contradicted by the premise (neutral). NLI has various practical applications and implications for advancing AI technology. It is crucial for improving question-answering systems, information retrieval and text summarization tools. Additionally, NLI serves as a stepping stone towards more complex language understanding tasks, such as reading comprehension and commonsense reasoning. 
\subsection{Natural Language Generation}
Natural Language Generation (NLG) is the task of generating grammatically correct human-readable descriptions from structured data or other input types. The complexity of NLG tasks includes diverse array of tasks ranging from simple template-based text generation to more complex approaches that involve deep learning and LLMs. The real-world applications of NLG are numerous and 
 reshaping across various industries. For instance, text summarization condenses large volumes of information into concise formats to benefit fields like journalism, research and business intelligence. Another significant application is machine translation which facilitates global communication by automatically converting text from one language to another. Other significant applications include automated report generation in finance and healthcare, personalized content creation for marketing, chatbots for customer service, and even creative writing assistance. 
\subsection{Semantic Similarity Detection}
Semantic similarity detection is a key task in natural language processing (NLP) that computes the degree of similarity between two input pairs (such as sentences, questions, or documents). This process involves to capture the underlying meaning and context of the text. Semantic similarity detection utilizes advanced techniques to identify conceptual relationships even when the exact words differ. Semantic similarity detection has impactful applications across diverse real-world problems. A key application is in plagiarism detection, where it aids to identify and highlight copied content by comparing documents to indicate unauthorized use of original work. Another significant application is duplicate detection, which is crucial for managing large databases to improve search engine results by eliminating redundant information. Moreover, semantic similarity detection plays a vital role in recommendation engines, information retrieval and question-answering system to enable context-aware responses to user queries and preferences.
\subsection{Information Extraction}
Information extraction (IE) is the task of automatically extracting specific pieces of information such as entities and relationships from unstructured or semi-structured sources. There are numerous real-world applications of IE across various industries. In healthcare, IE systems can extract patient information from clinical notes to diagnosis and treat different diseases. Financial institutions use IE to analyze market trends and extract key data from financial reports and news articles to support investment decisions and risk management. Additionally, IE is crucial in scientific research for extracting relevant information from academic papers and reports, facilitating knowledge discovery and accelerating the research process. These applications demonstrate IE enhances data usability, automate complex data processing tasks and supports decision-making across diverse fields.

\subsection{Information Retrieval}
Information Retrieval (IR) is the task of locating and extracting relevant information from large collections of unstructured or semi-structured data in response to a user's query. IR systems process user queries, identify relevant terms and use various techniques to rank and retrieve documents or information items based on their relevance. Generally, these systems involve indexing of data objects, utilizing IR models to calculate relevance scores and present results in a ranked list. In e-commerce, IR techniques enhance product search and recommendations to improve the shopping experience. Enterprise search systems utilize IR to help employees quickly find relevant documents across various data stores, boosting productivity and knowledge sharing. In customer service, IR powers self-service portals and assists support teams in rapidly accessing information to resolve queries. Web search engines, digital libraries and media search platforms all rely on IR techniques to provide users with relevant results. 
\subsection{Linguistic and Semantic Understanding }
Linguistic and Semantic Understanding involves interpretation of the meaning of language in context. Mainly, it analyzes phrases and sentences to identify their actual meaning based on different factors such as syntax and semantics.  In search engines, it enhances relevance by comprehending user queries more accurately. Chatbots or virtual assistants utilize it to identify user intent and provide accurate responses. Sentiment analysis leverage it to identify emotions in the text to aid customer feedback analysis. Additionally, recommendation systems use this technique to deliver personalized content by analysing user preferences. 
\subsection{Reasoning based Tasks}
Reasoning based tasks draw conclusion based on available description, knowledge or logic. It is a simple cognitive process that assists problem-solving, decision-making and understanding of complex problems. There are multiple subtypes of reasoning, which are briefly described below:
\begin{enumerate}
    \item Commonsense Reasoning: This involves using everyday knowledge and experiences to understand and navigate the world. It enables AI systems to make intuitive judgments and handle situations that require general knowledge such as conversational AI agents like Siri or Alexa and Smart home devices that anticipate user needs.
    \item Symbolic Reasoning: This type uses abstract symbols and rules to represent and manipulate knowledge. Few applications of symbolic reasoning are theorem provers in mathematics, Formal logic operations and Knowledge representation.
    \item Logical Reasoning: This involves using structured arguments and rules of inference to draw valid conclusions, A few applications of logical reasoning are legal analysis and argumentation, scientific hypothesis testing, automated planning and scheduling systems.
    \item Temporal Reasoning: This focuses on understanding and reasoning about time-related concepts and events such as weather forecasting systems, financial market prediction models and scheduling and logistics optimization.
    \item Spatial and Geometric Reasoning: This involves understanding and manipulating spatial relationships and geometric properties. It has many real-world application including: computer-aided design (CAD) software, robotics and autonomous navigation systems and virtual and augmented reality applications.
    \item Numerical and Arithmetic Reasoning: This type deals with numerical computations and mathematical problem-solving. A few real-world applications of Numerical and Arithmetic Reasoning are Financial analysis and forecasting, scientific computing and simulations and educational software for mathematics.

\end{enumerate}

These various types of reasoning have numerous real-world applications across different industries and domains. As AI continues to advance, these reasoning capabilities are becoming more sophisticated, enabling systems to handle increasingly complex tasks and make more nuanced decisions in real-world scenarios.
\section{A Comprehensive Analysis of Prompt Optimization Strategies across Diverse NLP Tasks, Pretrained Model, Benchmark Datasets and Performance Insights}
This section provides a comprehensive analysis of 45 different prompt optimization strategies across 9 different classes of NLP Tasks, in terms of benchmark dataset, LLM used for evaluation and performance values of underlying tasks.
\subsection{Classification}
Classification plays a crucial role in Natural Language Understanding (NLU) tasks by enabling systems to categorize text into predefined labels and facilitates efficient information retrieval and decision-making. Table \ref{Classification} explores a variety of classification sub-tasks including both binary and multiclass scenarios. It demonstrates various techniques and their influence on performance across various datasets and domains. Table \ref{Classification} categorizes nine distinct tasks under binary classification including fake news detection, coreference resolution, hate speech detection, sarcasm detection, hyperpartisan detection, acceptability judgments, jail break detection and word sense disambiguation. \\

Table \ref{Classification} illustrates that ProTeGi along with GPT-3.5-turbo pre-trained model is employed for fake news detection, hate speech detection and sarcasm detection. It indicates higher accuracy for hate speech detection followed by sarcasm detection. However, for fake news detection, it demonstrates a significant decrease in accuracy. ProTeGi along with an updated version of GPT, GPT-4 is also employed for jailbreak detection. However, it did not surpass performance on hate speech detection task. Moreover, Table \ref{Classification} exhibits performance of 12 unique approaches along with their best pre-trained models on 3 distinct data for coreference resolution task. Disambiguation-qa (BBH) is the most commonly used data to evaluate 8 distinct approaches. \\

Table \ref{Classification} shows a significant variation in its performance across various approaches ranges from about 50\% to 76\% in terms of accuracy. Among EvoPrompt, APE \cite{zhou2022large}, StablePRompt and Adv-ICL approaches, EvoPrompt paired with GPT-3.5 model outperform all other approaches. Contrarily, Adv-ICL along with GPT-3.5 model and StablePRompt with Gemma-7B pre-trained model underperforms and exhibit low accuracy score of less than 65\%. For WSC (Winograd Schema Challenge) data, P-tuning v2 with GLM-xxlarge model exhibits higher accuracy of above 90\% while P-tuning with GPT-2 Med model demonstrates about 20\% decrease in accuracy. Despite utilizing same amount of training data, DEPT, along with T5-base model exhibit the lowest performance. Whereas, Prompt-Tuning along with T5-xxl pre-trained model utilized comparatively smaller training dataset, illustrates an excellent aaccuracy score of more than 95\%, indicates its strong generalization capability. PROMPTBREEDER approach shows comparative results (nearly 90\%) for hate speech detection and subjectivity classification task with PaLM2-L and GPT-4 pre-trained model respectively. To assess performance of subjectivity classification, subj data is widely used. Notably, performance of various approaches depends on the size of training data, performance increases with the size of training data and vice versa. Additionally, Snarks (BBH) is utilized to evaluate various approaches for sarcasm detection. Despite the least amount of training data, OPRO along with PaLM2-L pre-trained model shows highest accuracy followed by ABO approach \cite{ma2024large} with Llama-2-70B-chat model. Additionally, Word-in-Context (WiC) and Word-in-Context of IIT are widely used across various approaches for word sense disambiguation task. Table \ref{Classification} shows that various approaches demonstrate comparable results. Moreover, PROMPTBREEDER along with PaLM2-L and APE with text-davinci-002 model conducted experiments on few-shot settings. Although both approaches demonstrate comparable results, PROMPTBREEDER still perform better than APE. \\
\scriptsize
% \usepackage{multirow}
% \renewcommand{\arraystretch}{1}
\begin{longtable}{|p{0.65cm}|p{1.85cm}|p{1,8cm}|p{1.8cm}|p{1.35cm}|p{1.45cm}|p{1.5cm}|p{1.85cm}|}
\caption{Overview of existing classification predictors. ``*", ``**" and ``***" symbol following a dataset name signifies the identical train/dev/test split was utilized for the reported results.}
\label{Classification}\\
\hline
\textbf{Task} & \textbf{Sub-Task} & \textbf{Method} & \textbf{Data} & \textbf{Domain} & \textbf{Pre-trained Model} & \textbf{Evaluation Metric} & \textbf{Performance} \\ \hline
& {\begin{tabular}[c]{@{}c@{}}Fake \\News \\ Detection \end{tabular}} & ProTeGi & {\begin{tabular}[c]{@{}c@{}}LIAR\\\cite{wang2017liar}\end{tabular}}  & {\begin{tabular}[c]{@{}c@{}}Political\\ Debate,\\ TV Ads,\\ Facebook\\ Posts,\\ Tweets,\\ Interview, \\News\end{tabular}}   & GPT-3.5-turbo & Accuracy & 0.64 \\ \cmidrule(lr){2-8} 
& Coreference Resolution & Prompt-Tuning & {\begin{tabular}[c]{@{}c@{}}WSC\\\cite{levesque2012winograd}\end{tabular}}  & {\begin{tabular}[c]{@{}c@{}}Fiction \\ Books \\News\end{tabular}} & T5-xxl & Accuracy & 96.2 \\ \cmidrule(lr){3-8} 
& & {\begin{tabular}[c]{@{}c@{}}P-tuning \\ V2 \end{tabular}}& {\begin{tabular}[c]{@{}c@{}}WSC\\\cite{levesque2012winograd} *\end{tabular}}  & {\begin{tabular}[c]{@{}c@{}}Fiction \\ Books \\News\end{tabular}} & GLM-xxlarge & Accuracy & 93.3 \\ \cmidrule(lr){3-8} 
& & P-tuning &  {\begin{tabular}[c]{@{}c@{}}WSC\\\cite{levesque2012winograd} *\end{tabular}} & {\begin{tabular}[c]{@{}c@{}}Fiction \\ Books \\News\end{tabular}} & GPT-2-Med & Accuracy & 74 \\ \cmidrule(lr){3-8} 
& & \multirow{2}{*}{DEPT} &  {\begin{tabular}[c]{@{}c@{}}WSC\\\cite{levesque2012winograd} \end{tabular}} & {\begin{tabular}[c]{@{}c@{}}Fiction \\ Books \\News\end{tabular}} & T5-base & Accuracy & 67.3 \\ \cmidrule(lr){4-8} 
& &  & WG \cite{sakaguchi2021winogrande} & {\begin{tabular}[c]{@{}c@{}}Social \\ Common\\sense, \\Physical, \\ Common\\sense\end{tabular}}  & T5-base & Accuracy & 59 \\ \cmidrule(lr){3-8} 
& & OPRO &  {\begin{tabular}[c]{@{}c@{}}disambiguation \\qa (BBH) \cite{suzgun2022bigbenchhard}\end{tabular}}   & \_ & GPT-3.5-turbo, text bison scorer & Accuracy & 75.5 \\ \cmidrule(lr){3-8} 
& & EvoPrompt & {\begin{tabular}[c]{@{}c@{}}disambiguation \\qa (BBH) \cite{suzgun2022bigbenchhard} *\end{tabular}} & \_ & GPT-3.5 & Accuracy & 71.2 \\ \cmidrule(lr){3-8} 
& &  APE & {\begin{tabular}[c]{@{}c@{}}disambiguation \\qa (BBH) \cite{suzgun2022bigbenchhard} *\end{tabular}} & \_ & text-davinci-002 & Normalized Score & 5.6 \\ \cmidrule(lr){3-8} 
& & MoP & {\begin{tabular}[c]{@{}c@{}}disambiguation \\qa (BBH) \cite{suzgun2022bigbenchhard} \end{tabular}} & \_ & \_ & \_ & \_ \\ \cmidrule(lr){3-8} 
& & AEO & {\begin{tabular}[c]{@{}c@{}}disambiguation \\qa (BBH) \cite{suzgun2022bigbenchhard} \end{tabular}} \cite{suzgun2022bigbenchhard} & \_ & text-bison task model and PaLM2-L optimizer & Accuracy & 68 \\ \cmidrule(lr){3-8} 
& & {\begin{tabular}[c]{@{}c@{}}Stable\\PRompt \end{tabular}} & {\begin{tabular}[c]{@{}c@{}}disambiguation \\qa (BBH) \cite{suzgun2022bigbenchhard} *\end{tabular}} & \_ & Gemma-7B & Accuracy & 64.04 \\ \cmidrule(lr){3-8} 
\rotatebox{90}{Binary Classification} & & PE2 & {\begin{tabular}[c]{@{}c@{}}disambiguation \\qa (BBH) \cite{suzgun2022bigbenchhard} \end{tabular}} & \_ & GPT-3.5-turbo-instruct  as the task model and GPT-4  turbo as the prompt proposal model. & Accuracy & 50 \\ \cmidrule(lr){3-8} 
& & \multirow{2}{*}{Adv-ICL} & {\begin{tabular}[c]{@{}c@{}}disambiguation \\qa (BBH) \cite{suzgun2022bigbenchhard} *\end{tabular}} & \_ & GPT-3.5-turbo-0613 & Accuracy & less than 64 \\ \cmidrule(lr){4-8} 
& &  & WSC \cite{levesque2012winograd} & {\begin{tabular}[c]{@{}c@{}}Fiction \\ Books \end{tabular}} & text-davinci-002 & Accuracy & 73.8 \\ \cmidrule(lr){2-8} 
& {\begin{tabular}[c]{@{}c@{}}Hate \\ Speech \\ Detection \end{tabular}} & ProTeGi & Ethos \cite{mollas2022ethos} & {\begin{tabular}[c]{@{}c@{}}YouTube \\ and \\Reddit \\Comments \\News\end{tabular}}   & GPT-3.5-turbo & Accuracy & 0.95 \\ \cmidrule(lr){3-8} 
&  & {\begin{tabular}[c]{@{}c@{}}Prompt\\ BREEDER\end{tabular}}  & Ethos \cite{mollas2022ethos} & {\begin{tabular}[c]{@{}c@{}}YouTube \\ and \\Reddit \\Comments \\News\end{tabular}}& PaLM2-L & Accuracy & 89\% \\ \cmidrule(lr){2-8} 
& {\begin{tabular}[c]{@{}c@{}}Subjectivity \\ Classification\end{tabular}} & RLPrompt & Subj \cite{pang2004sentimental} & Movie Reviews & RoBERTa-large & Accuracy & 81.9 (2 discrete tokens) \\ \cmidrule(lr){3-8}
& & EvoPrompt & Subj \cite{pang2004sentimental} & Movie Reviews & Alpaca-7b & Accuracy & 75.55 (DE) \\ \cmidrule(lr){3-8} 
& & {\begin{tabular}[c]{@{}c@{}}Prompt\\ AGENT\end{tabular}}  & Subj \cite{pang2004sentimental} & Movie Reviews & GPT-4 & Accuracy & 0.879 \\ \cmidrule(lr){3-8} 
& & Waywardness & Subj \cite{pang2004sentimental} & Movie Reviews & GPT-2 & Prompt F1 & 100 \\ \cmidrule(lr){3-8} 
& & Random-prompt & Subj \cite{pang2004sentimental} & Movie Reviews & Mistral 7B & Accuracy & 84.4 (Random with Context) \\ \cmidrule(lr){2-8} 
& {\begin{tabular}[c]{@{}c@{}}Sarcasm \\ Detection \end{tabular}} & ProTeGi & ArSarcasm \cite{farha2020arabic} & Tweet & GPT-3.5-turbo & Accuracy & 0.87 \\ \cmidrule(lr){3-8} 
& & OPRO & Snarks (BBH) \cite{suzgun2022bigbenchhard} & \_ & \begin{tabular}[c]{@{}l@{}}PaLM2-L\\ scorer, \\ PaLM2-\\L-IT \\Optimizer\end{tabular} & Accuracy & 83.2 \\ \cmidrule(lr){3-8} 
& & EvoPrompt & Snarks (BBH) \cite{suzgun2022bigbenchhard} * & \_ & GPT-3.5 & Accuracy & 77.53 \\ \cmidrule(lr){3-8} 
& & APE & Snarks (BBH) \cite{suzgun2022bigbenchhard} * & \_ & text-davinci-002 & Normalized Score & 4 \\ \cmidrule(lr){3-8} 
& & MoP & Snarks (BBH) \cite{suzgun2022bigbenchhard} & \_ & \_ & \_ & \_ \\ \cmidrule(lr){3-8} 
& & ABO & Snarks (BBH) \cite{suzgun2022bigbenchhard} & \_ & Llama-2-70B-chat &  & 0.811 \\ \cmidrule(lr){3-8} 
& & {\begin{tabular}[c]{@{}c@{}}Stable \\PRompt\end{tabular}} & Snarks (BBH) \cite{suzgun2022bigbenchhard} * & \_ & Gemma-7B & Accuracy & 52.32 \\ \cmidrule(lr){3-8} 
& & PE2 & Snarks (BBH) \cite{suzgun2022bigbenchhard} & \_ & GPT-3.5-turbo-instruct as the task model and GPT-4 as the prompt proposal model. & Accuracy & 73.08 \\ \cmidrule(lr){3-8} 
& & Adv-ICL & Snarks (BBH) \cite{suzgun2022bigbenchhard} * & \_ & GPT-3.5-turbo-0613 & Accuracy & less than 75 \\ \cmidrule(lr){2-8}
& {\begin{tabular}[c]{@{}c@{}}Hyperpartisan \\ Detection \end{tabular}} & BDPL & HP \cite{kiesel2019semeval} & \begin{tabular}[c]{@{}c@{}}News \\ Articles \end{tabular} & GPT-3 Davinci & F1-Score & 58.5 \\ \cmidrule(lr){2-8}
& {\begin{tabular}[c]{@{}c@{}}Acceptability \\ Judgments \end{tabular}} & BDPL & CoLA \cite{warstadt2019neural} & Books, Articles & GPT-3 Davinci & Matthews Correlation & 58.4 \\ \cmidrule(lr){3-8} 
& & DEPT & CoLA \cite{warstadt2019neural} & Books, Articles & T5-base & Accuracy & 63.8 \\ \cmidrule(lr){2-8}
& {{\begin{tabular}[c]{@{}c@{}}Jail \\ Break \\ Detection \end{tabular}}} & ProTeGi & Multilingual jailbreak & \_ & GPT-4 & Accuracy & 0.88 \\ \cmidrule(lr){2-8}
& {\begin{tabular}[c]{@{}c@{}}Word \\Sense \\ Disambiguation \end{tabular}} & P-tuning v2 & WiC \cite{pilehvar2018wic} * & WordNet, VerbNet, Wiktionary & BERT-large & Accuracy & 75.1 \\ \cmidrule(lr){3-8} 
& & P-tuning & WiC \cite{pilehvar2018wic} * & WordNet, VerbNet, Wiktionary & BERT-Large & Accuracy & 72.7 \\ \cmidrule(lr){3-8} 
& & LoPT & WiC \cite{pilehvar2018wic} * & WordNet, VerbNet, Wiktionary & T5-base & Accuracy & 62.7(LoPT-2) \\ \cmidrule(lr){3-8} 
& & DEPT & WiC \cite{pilehvar2018wic} & WordNet, VerbNet, Wiktionary & T5-base & Accuracy & 68.7 \\ \cmidrule(lr){3-8} 
& & Prompt-Tuning & WiC \cite{pilehvar2018wic} & WordNet, VerbNet, Wiktionary & T5-xxl & Accuracy & 76.6 \\ \cmidrule(lr){3-8} 
& & {\begin{tabular}[c]{@{}c@{}}Prompt\\ BREEDER\end{tabular}}  & {\begin{tabular}[c]{@{}c@{}}Wic-IIT \\ \cite{honovich2022instruction} *\end{tabular}}   & WordNet, VerbNet, Wiktionary & PaLM2-L & Execution Accuracy & 65(fewshot) \\ \cmidrule(lr){3-8}
& & APE & {\begin{tabular}[c]{@{}c@{}}Wic-IIT \\ \cite{honovich2022instruction} *\end{tabular}} & WordNet, VerbNet, Wiktionary & text-davinci-002 & Execution Accuracy & 63 (fewshot) \\ \cmidrule(lr){3-8} 
& & MoP & {\begin{tabular}[c]{@{}c@{}}Wic-IIT \\ \cite{honovich2022instruction} \end{tabular}}  & WordNet, VerbNet, Wiktionary & \_ & \_ & \_ \\ \cmidrule(lr){3-8} 
& & PE2 & {\begin{tabular}[c]{@{}c@{}}Wic-IIT \\ \cite{honovich2022instruction} \end{tabular}}  & WordNet, VerbNet, Wiktionary & GPT-3.5-turbo-instruct as the task model and GPT-4 as the prompt proposal model. & Accuracy & 61 \\ \hline
& {{\begin{tabular}[c]{@{}c@{}}Abstract \\Sentence \\ Roles \\ Sentence\\ Classification\end{tabular}}} & BDPL & RCT \cite{dernoncourt2017pubmed} & Medical & GPT-3 Curie & F1-Score & 49.6 \\ \cmidrule(lr){2-8}
& {{\begin{tabular}[c]{@{}c@{}}Citation\\ Intent\end{tabular}}}  & BDPL & CI & Computer Science & GPT-3 Babbage & F1-Score & 40.2 \\ \cmidrule(lr){2-8}
\rotatebox{90}{Multiclass Classification} & {{\begin{tabular}[c]{@{}c@{}}Topic \\ Classification\end{tabular}}} & {RLPrompt} & {\begin{tabular}[c]{@{}c@{}}AG’s \\News  \cite{zhang2015character} *\end{tabular}}    & {\begin{tabular}[c]{@{}c@{}}News\\ Articles\end{tabular}} & RoBERTa-large & Accuracy & 80.2 (5 tokens) \\ \cmidrule(lr){4-8} 
& &  & TREC \cite{voorhees2000building} & Question Types & RoBERTa-large & Accuracy & 57.6  (5 tokens) \\ \cmidrule(lr){4-8} 
& &  & Yahoo  \cite{zhang2015character} & Various & RoBERTa-large & Accuracy & 48.6 (both tokens) \\ \cmidrule(lr){4-8} 
& & & DBPedia  \cite{lehmann2015dbpedia} * & Wikipedia Ontologies & RoBERTa-large & Accuracy & 84.6 (5 tokens) \\ \cmidrule(lr){3-8} 
& & TEMPERA & {\begin{tabular}[c]{@{}c@{}}AG’s \\News  \cite{zhang2015character} *\end{tabular}}  & {\begin{tabular}[c]{@{}c@{}}News\\ Articles\end{tabular}} & RoBERTa-large & Accuracy & 85.5 \\ \cmidrule(lr){3-8} 
& & {Active Examples} & {\begin{tabular}[c]{@{}c@{}}AG’s \\News  \cite{zhang2015character} \end{tabular}}  & {\begin{tabular}[c]{@{}c@{}}News\\ Articles\end{tabular}} & BABBAGE (Calibration) & Accuracy & 78.1 \\ \cmidrule(lr){4-8} 
& &  & TREC \cite{voorhees2000building} & Question Types & CURIE (Calibration) & Accuracy & 47 \\ \cmidrule(lr){3-8} 
& & Discrete-v & {\begin{tabular}[c]{@{}c@{}}AG’s \\News  \cite{zhang2015character} \end{tabular}}  & {\begin{tabular}[c]{@{}c@{}}News\\ Articles\end{tabular}} & BERT-base-cased & Accuracy & 79.8 \\ \cmidrule(lr){3-8} 
& & {Waywardness} & {\begin{tabular}[c]{@{}c@{}}AG’s \\News  \cite{zhang2015character} **\end{tabular}}  &{\begin{tabular}[c]{@{}c@{}}News\\ Articles\end{tabular}} & GPT-2 & Prompt F1 & 95.7 \\ \cmidrule(lr){4-8} 
& &  & TREC \cite{voorhees2000building} & Question Types & GPT-2 & Prompt F1 & 99.3 \\ \cmidrule(lr){3-8} 
& & {BBT} & {\begin{tabular}[c]{@{}c@{}}AG’s \\News  \cite{zhang2015character} *\end{tabular}} & {\begin{tabular}[c]{@{}c@{}}News\\ Articles\end{tabular}} & RoBERTa-large & Accuracy & 81.51 \\ \cmidrule(lr){4-8} 
& &  & DBPedia  \cite{lehmann2015dbpedia} * & Wikipedia Ontologies & RoBERTa-large & Accuracy & 79.99 \\ \cmidrule(lr){3-8} 
& & CLAPS & {\begin{tabular}[c]{@{}c@{}}AG’s \\News  \cite{zhang2015character} *\end{tabular}}  & {\begin{tabular}[c]{@{}c@{}}News\\ Articles\end{tabular}} & Flan-T5large & Accuracy & 84.24 (genetics) \\ \cmidrule(lr){3-8} 
& & FEDBPT & {\begin{tabular}[c]{@{}c@{}}AG’s \\News  \cite{zhang2015character} \end{tabular}}  & {\begin{tabular}[c]{@{}c@{}}News\\ Articles\end{tabular}} & Llama-2 & Accuracy & 90 \\ \cmidrule(lr){3-8} 
& & {BBTv2} & {\begin{tabular}[c]{@{}c@{}}AG’s \\News  \cite{zhang2015character} *\end{tabular}}  & {\begin{tabular}[c]{@{}c@{}}News\\ Articles\end{tabular}} & RoBERTa-large & Accuracy & 85.28 \\ \cmidrule(lr){4-8} 
& &  & DBPedia  \cite{lehmann2015dbpedia} * & Wikipedia Ontologies & RoBERTa-large & Accuracy & 93.64 \\ \cmidrule(lr){3-8} 
& & {EvoPrompt} & AG’s News  \cite{zhang2015character} & {\begin{tabular}[c]{@{}c@{}}News\\ Articles\end{tabular}} & Alpaca-7b & Accuracy & 73.82 (DE) \\ \cmidrule(lr){4-8} 
& &  & TREC \cite{voorhees2000building} & Question Types & Alpaca-7b & Accuracy & 64.00 (GA) \\ \cmidrule(lr){3-8} 
& & {\begin{tabular}[c]{@{}c@{}}Prompt\\ AGENT\end{tabular}}  & TREC \cite{voorhees2000building} & Question Types & GPT-3.5 & Accuracy & 0.886 \\ \cmidrule(lr){3-8} 
& & LoPT & {\begin{tabular}[c]{@{}c@{}}AG’s \\News  \cite{zhang2015character} **\end{tabular}}  & {\begin{tabular}[c]{@{}c@{}}News\\ Articles\end{tabular}} & GPT-2-large & Accuracy & 91.9 (LoPT-1) \\ \cmidrule(lr){3-8} 
& & PRewriter & {\begin{tabular}[c]{@{}c@{}}AG’s \\News  \cite{zhang2015character} \end{tabular}}  & {\begin{tabular}[c]{@{}c@{}}News\\ Articles\end{tabular}} & PaLM2-S & Accuracy & 85.2 (PRewrite-S) \\ \cmidrule(lr){3-8} 
& & {\begin{tabular}[c]{@{}c@{}}Fluent\\Prompt\end{tabular}}  & {\begin{tabular}[c]{@{}c@{}}AG’s \\News  \cite{zhang2015character} **\end{tabular}}  & {\begin{tabular}[c]{@{}c@{}}News\\ Articles\end{tabular}} & GPT-2-large & Accuracy & 68 \\ \cmidrule(lr){3-8} 
& & {\begin{tabular}[c]{@{}c@{}}Prompt\\ Boosting\end{tabular}}  & AG’s News \cite{zhang2015character} & {\begin{tabular}[c]{@{}c@{}}News\\ Articles\end{tabular}} & RoBERTa-large & Accuracy & 85.2, 84.2 \\ \cmidrule(lr){4-8} 
& &  & TREC \cite{voorhees2000building} & Question Types & RoBERTa-large & Accuracy & 81.6, 84.5 \\ \cmidrule(lr){3-8} 
& & EASE & AG's News \cite{zhang2015character} ** & {\begin{tabular}[c]{@{}c@{}}News\\ Articles\end{tabular}} & GPT-3.5-turbo-1106 & Accuracy & 65 \\ \cmidrule(lr){3-8} 
& & {Random-prompt} & AG’s News \cite{zhang2015character} & {\begin{tabular}[c]{@{}c@{}}News\\ Articles\end{tabular}} & GPT-3.5 & Accuracy & 83.6 (Random with Context) \\ \cmidrule(lr){4-8} 
& &  & TREC \cite{voorhees2000building} & Question Types & GPT-3.5 & Accuracy & 77.7 (Random Vocabulary ) \\ \cmidrule(lr){4-8} 
& &  & DBPedia  \cite{lehmann2015dbpedia} & Wikipedia Ontologies & GPT-3.5 & Accuracy & 91.6(Random w/o Context) \\ \hline
& Sentiment Analysis  & {RLPrompt} & SST-2  \cite{socher2013recursive} * & Movie Reviews & RoBERTa-large & Accuracy & 92.5 (5 tokens) \\ \cmidrule(lr){4-8} 
& &  & Yelp P  \cite{zhang2015character} * & {\begin{tabular}[c]{@{}c@{}}Yelp\\ Reviews\end{tabular}} & RoBERTa-large & Accuracy & 95.1 (5 tokens) \\ \cmidrule(lr){4-8} 
& &  & MR \cite{pang2005seeing} * & Movie Reviews & RoBERTa-large & Accuracy & 87.1 (5  tokens) \\ \cmidrule(lr){4-8} 
& &  & CR \cite{hu2004mining} * & Product Reviews & RoBERTa-large & Accuracy & 89.5 (5 tokens) \\ \cmidrule(lr){4-8} 
& & & SST-5 & Movie Reviews & RoBERTa-large & Accuracy & 41.4 (5 tokens) \\ \cmidrule(lr){4-8} 
& &  & Yelp \cite{zhang2015character} & {\begin{tabular}[c]{@{}c@{}}Yelp\\ Reviews\end{tabular}} & RoBERTa-large & Accuracy & 45.6 (2 tokens) \\ \cmidrule(lr){3-8} 
& & BDPL & SST-2  \cite{socher2013recursive} & Movie Reviews & GPT-3 Davinci & Accuracy & 89.3 \\ \cmidrule(lr){3-8} 
\rotatebox{90}{\begin{tabular}[c]{@{}l@{}}Binary and \\ Multiclass \\ Classification\end{tabular}} & & {TEMPERA} & SST-2  \cite{socher2013recursive} * & Movie Reviews & RoBERTa-large & Accuracy & 91.9 \\ \cmidrule(lr){4-8} 
& &  & Yelp P \cite{zhang2015character} * & {\begin{tabular}[c]{@{}c@{}}Yelp\\ Reviews\end{tabular}} & RoBERTa-large & Accuracy & 92.6 \\ \cmidrule(lr){4-8} 
& & & MR \cite{pang2005seeing} * & Movie Reviews & RoBERTa-large & Accuracy & 88 \\ \cmidrule(lr){4-8} 
& &  & CR \cite{hu2004mining} * & Product Reviews & RoBERTa-large & Accuracy & 91.1 \\ \cmidrule(lr){3-8} 
& & {\begin{tabular}[c]{@{}c@{}}Automate\\-CoT\end{tabular}} & SST-2 \cite{socher2013recursive} & Movie Reviews & Code-Davinci-002 & Exact Match Accuracy & 92.8 \\ \cmidrule(lr){3-8} 
& & {DEPT} & SST-2 \cite{socher2013recursive} & Movie Reviews & LLAMA-2-13B & Accuracy & 96.01 \\ \cmidrule(lr){4-8} 
& &  & Yelp P \cite{zhang2015character} & {\begin{tabular}[c]{@{}c@{}}Yelp\\ Reviews\end{tabular}} & T5-base & Accuracy & 96.8 \\ \cmidrule(lr){3-8} 
& & {Random-prompt} & SST-2 \cite{socher2013recursive} & Movie Reviews & GPT-3.5 & Accuracy & 94.7 (Random Vocabulary) \\ \cmidrule(lr){4-8} 
& &  & SST-5 & Movie Reviews & LLAMA-2-7B & Accuracy & 49.5 (Random with Context) \\ \cmidrule(lr){4-8} 
& &  & MR \cite{pang2005seeing} & Movie Reviews & Mistral-7B, LLAMA-2-7B & Accuracy & 93.1 (random vocabulary), 93.1 (Random with Context) \\ \cmidrule(lr){4-8} 
& &  & CR \cite{hu2004mining} & Product Reviews & GPT-3.5 & Accuracy & 91.6 (Random Vocabulary) \\ \cmidrule(lr){3-8} 
& & {Active Examples} & SST-2 \cite{socher2013recursive} & Movie Reviews & CURIE & Accuracy & 938 \\ \cmidrule(lr){4-8} 
& &  & Amazon P & Product Reviews & CURIE (Calibration) & Accuracy & 948 \\ \cmidrule(lr){3-8} 
& & {BBT} & SST-2 \cite{socher2013recursive} *** & Movie Reviews & RoBERTa-large & Accuracy & 89.56 \\ \cmidrule(lr){4-8} 
& &  & Yelp P \cite{zhang2015character} * & {\begin{tabular}[c]{@{}c@{}}Yelp\\ Reviews\end{tabular}} & RoBERTa-large & Accuracy & 91.5 \\ \cmidrule(lr){3-8} 
& & CLAPS & SST-2 \cite{socher2013recursive} *** & Movie Reviews & Flan-T5large & Accuracy & 94.27 ( Greedy) \\ \cmidrule(lr){3-8} 
& & {FEDBPT} & SST-2 \cite{socher2013recursive} & Movie Reviews & RoBERTa-large & Accuracy & 87.16 (IID) \\ \cmidrule(lr){4-8} 
& &  & Yelp P \cite{zhang2015character} & {\begin{tabular}[c]{@{}c@{}}Yelp\\ Reviews\end{tabular}} & RoBERTa-large & Accuracy & 91.12 (IID) \\ \cmidrule(lr){3-8} 
& & {BBTv2} & SST-2 \cite{socher2013recursive} *** & Movie Reviews & RoBERTa-large & Accuracy & 90.33 \\ \cmidrule(lr){4-8} 
& &  & Yelp \cite{zhang2015character} & {\begin{tabular}[c]{@{}c@{}}Yelp\\ Reviews\end{tabular}}& RoBERTa-large & Accuracy & 92.86 \\ \cmidrule(lr){3-8} 
& & {EvoPrompt} & SST-2 \cite{socher2013recursive} & Movie Reviews & Alpaca-7b & Accuracy & 95.13 \\ \cmidrule(lr){4-8} 
& &  & MR \cite{pang2005seeing} & Movie Reviews & Alpaca-7b & Accuracy & 91.4 \\ \cmidrule(lr){4-8} 
& & & CR \cite{hu2004mining} & Product Reviews & Alpaca-7b & Accuracy & 90.22 \\ \cmidrule(lr){4-8} 
& &  & SST-5 \cite{socher2013recursive} & Movie Reviews & Alpaca-7b & Accuracy & 49.91 \\ \cmidrule(lr){3-8} 
& & {\begin{tabular}[c]{@{}c@{}}Prompt-\\ BO\end{tabular}} & SST-2 \cite{socher2013recursive} & Movie Reviews & RoBERTa-large & Accuracy & 86.2 \\ \cmidrule(lr){3-8} 
& & LoPT & SST-2 \cite{socher2013recursive} & Movie Reviews & T5-base & Accuracy & 92.9(LoPT-1) \\ \cmidrule(lr){3-8} 
& & PRewriter & SST-2 \cite{socher2013recursive} & Movie Reviews & PaLM2-S & Accuracy/F1-Score & 96.6 \\ \cmidrule(lr){3-8} 
& & {Discrete-v} & SST-2 \cite{socher2013recursive} & Movie Reviews & BERT-base-cased & Accuracy & 74.9 \\ \cmidrule(lr){4-8} 
& &  & IMDB & Movie Reviews & BERT-base-cased & Accuracy & 73.45 \\ \cmidrule(lr){4-8} 
& & & Amazon Polarity \cite{mcauley2013hidden} & Product Reviews & BERT-base-cased & Accuracy & 80.76 \\ \cmidrule(lr){3-8} 
& & {Waywardness} & SST-2 \cite{socher2013recursive} ** & Movie Reviews & GPT-2 & Prompt F1 & 98.5 \\ \cmidrule(lr){4-8} 
& &  & SST-5 \cite{socher2013recursive} * & Movie Reviews & GPT-2 & Prompt F1 & 96.1 \\ \cmidrule(lr){3-8} 
& & {\begin{tabular}[c]{@{}c@{}}AUTO\\ PROMPT\end{tabular}} & SST-2 \cite{socher2013recursive} ** & Movie Reviews & RoBERTa-large & Accuracy & 91.4 \\ \cmidrule(lr){3-8} 
& & {\begin{tabular}[c]{@{}c@{}}Fluent \\Prompt \end{tabular}}  & SST-2 \cite{socher2013recursive} ** & Movie Reviews & GPT-2 large & Accuracy & 88.2 \\ \cmidrule(lr){4-8} 
& &  & Amazon Polarity \cite{mcauley2013hidden} & Product Reviews & GPT-2 large & Accuracy & 85.3 \\ \cmidrule(lr){3-8} 
& & {\begin{tabular}[c]{@{}c@{}}Stable \\PRompt\end{tabular}}  & SST-2 \cite{socher2013recursive} * & Movie Reviews & RoBERTa-large & Accuracy & 92.8 \\ \cmidrule(lr){4-8} 
& &  & MR \cite{pang2005seeing} * & Movie Reviews & RoBERTa-large & Accuracy & 87.4 \\ \cmidrule(lr){3-8} 
& & {MAPO} & SST-2 \cite{socher2013recursive} * & Movie Reviews & RoBERTa-large & Accuracy & 96.1 \\ \cmidrule(lr){4-8} 
& &  & Yelp P \cite{zhang2015character} * & {\begin{tabular}[c]{@{}c@{}}Yelp\\ Reviews\end{tabular}} & RoBERTa-large & Accuracy & 93.5 \\ \cmidrule(lr){4-8} 
& & & MR \cite{pang2005seeing} * & Movie Reviews & RoBERTa-large & Accuracy & 90.2 \\ \cmidrule(lr){4-8} 
& &  & CR \cite{hu2004mining} * & Product Reviews & RoBERTa-large & Accuracy & 88.9 \\ \cmidrule(lr){3-8} 
& & {\begin{tabular}[c]{@{}c@{}}Prompt\\ Boosting\end{tabular}}  & SST-2 \cite{socher2013recursive} & Movie Reviews & RoBERTa-large & Accuracy & 87.6, 87.6 \\ \cmidrule(lr){4-8} 
&  & & MR \cite{pang2005seeing} * & Movie Reviews & RoBERTa-large & Accuracy & 84.6, 84.7 \\ \cmidrule(lr){4-8} 
& &  & SST-5 \cite{socher2013recursive} & Movie Reviews & RoBERTa-large & Accuracy & 42.3, 44.0 \\ \cmidrule(lr){4-8} 
& & & CR \cite{hu2004mining} * & Product Reviews & RoBERTa-large & Accuracy & 86.8, 88.1 \\ \cmidrule(lr){4-8} 
& &  & MPQA & News & RoBERTa-large & Accuracy & 72.7, 75.4 \\ \cmidrule(lr){3-8} 
& & Adv-ICL & Yelp \cite{zhang2015character} & {\begin{tabular}[c]{@{}c@{}}Yelp\\ Reviews\end{tabular}} & text-davinci-002 & Accuracy & 74.4 \\ \cmidrule(lr){3-8} 
& & {EASE} & SST-5 \cite{socher2013recursive} * & Movie Reviews & GPT-3.5-turbo-1106 & Accuracy & 53.3 \\ \cmidrule(lr){4-8} 
& &  & MR \cite{pang2005seeing} * & Movie Reviews & GPT-3.5-turbo-1106 & Accuracy & 95 \\ \cmidrule(lr){4-8} 
& & & Yelp P \cite{zhang2015character} * & {\begin{tabular}[c]{@{}c@{}}Yelp\\ Reviews\end{tabular}} & GPT-3.5-turbo-1106 & Accuracy & 95 \\ \cmidrule(lr){4-8} 
& &  & CR \cite{hu2004mining} * & Product Reviews & GPT-3.5-turbo-1106 & Accuracy & 100 \\ \cmidrule(lr){4-8} 
& & & SST-2 (IIT) * & Movie Reviews & GPT-3.5-turbo-1106 & Accuracy & 100 \\ \cmidrule(lr){3-8} 
& & {\begin{tabular}[c]{@{}c@{}}Prompt\\ BREEDER\end{tabular}}  & SST-2 (IIT)  \cite{honovich2022instruction} * & Movie Reviews & PaLM2-L & Execution Accuracy & 93(fewshot) \\ \cmidrule(lr){3-8} 
& & APE & SST-2 (IIT)  \cite{honovich2022instruction} * & Movie Reviews & text-davinci-002 & Execution Accuracy & 94(zeroshot) \\ \cmidrule(lr){3-8} 
& & MoP & SST-2 (IIT)  \cite{honovich2022instruction} & Movie Reviews & \_ & \_ & \_ \\ \cmidrule(lr){3-8} 
& & PE2 & SST-2 (IIT) \cite{honovich2022instruction} & Movie Reviews & GPT-3.5-turbo-instruct  as the task model and GPT-4 as the prompt proposal model. & mean & 88.8 \\ \cmidrule(lr){3-8} 
& & InstructZero & SST-2 (IIT) \cite{honovich2022instruction} & Movie Reviews & Vicuna Model & Exact Match & 0.93 \\ \hline
% \end{tabular}
\end{longtable}

\normalsize
%%%%%%%%%%%%%%%%%%%%%%%%%%%%%%%%%%%%%%%%%%%%%%%%%%%%%%%%%%%%%%%%%%%%%%%%%%%%%%%%%%%%%%%%%%%%%%%%%%%%%
%Trends

Table \ref{Classification} shows that 3 distinct tasks, namely abstract sentence roles/sequential sentence classification, citation intent and topic classification can be categorized in multiclass classification. BDPL along with GPT-3 pre-trained model is employed for sequential sentence classification as well as for citation intent task. It is evident from Table \ref{Classification}, BDPL shows superior performance of about 9\% for sequential sentence classification task. Moreover, Table \ref{Classification} shows performance evaluation of 17 distinct approaches on 4 datasets. AG's News is a widely used dataset for this task, comprising four predefined classes namely, world, sports, business and technology. Waywardness, LoPT, FluentPrompt and EASE utilized a large amount of training data, due to which LoPT along with GPT-2 large pre-trained model, demonstrate a higher accuracy score of 90\%. However, FluentPrompt and EASE underperform as with reported accuracy below 70\%. Even though approaches including RLPrompt, TEMPERA, BBT, BBTv2 and CLAPS utilized least amount of training data still perform well and report their accuracy above 80\%. TREC data is also utilized by various approaches. PromptBoosting with RoBERTa-large utilized a larger amount of training data but still unable to outperform PROMPTAGENT along with GPT-3.5 pre-trained model. However, Random prompt paired with GPT-3.5 model utilized least amount of training data still outperform RLPrompt with RoBERTa-large and Active Examples approaches. Moreover, for DBPedia data, BBTv2 along with RoBERTa-large pre-trained model outperform all other approaches. However, Random prompt approach with GPT-3.5 model, utilized least amount of training data still outperform other approaches including RLPrompt and BBT paired with RoBERTa-large model.\\

Sentiment analysis is a fundamental task in natural language understanding (NLU), with a vast amount of available data supporting its study. This task can be formulated as both binary (positive vs. negative) and multiclass classification (very positive/positive/neutral/negative/very negative). To account for this diversity, a specialized category has been introduced that integrates both binary and multiclass sentiment analysis. Table \ref{Classification} demonstrates performance of 29 distinct approaches on 10 unique datasets. Stanford Sentiment Treebank (SST-2) is a binary-class sentiment analysis dataset, used to assess the performance of 22 distinct approaches. Waywardness, AUTOPROMPT and FluentPrompt utilized a larger amount of training data thus exhibit remarkable accuracy of nearly 90\%. Even though BBT, BBTv2 and CLAPS utilized a smaller training data size, still manages to outperform other approaches, indicates their generalization capability. Notably, RLPrompt, TEMPERA, StablePRompt and MAPO utilizes same amount of training data (least) but higher testing data, shows an efficient accuracy of above 90\%. Another subset of data SST-2 of instruction induction task (IIT) data is also employed to assess performance of 6 distinct approaches. PROMPTBREEDER along with PaLM2-L and APE paired with text-davinci-002 pre-trained model conducted experiments on few-shot settings. APE along with text-davinci-002 model exhibit 1\% higher execution accuracy than PROMPTBREEDER with PaLM2-L model. Moreover, another dataset of SST, SST-5, is utilized, featuring 5 classes. However, due to difference in data split, performance of various approaches can not be compared.

Moreover, for MR data, EASE along with GPT-3.5 model outperform all other approaches followed by MAPO with RoBERTa-large model. However, PromptBoosting accompanied with RoBERTa-large model underperform with accuracy below 85\%. While other approaches including RLPrompt, TEMPERA and StablePRompt along with RoBERTa-large model demonstrate accuracy ranges from above 85\% to below 90\%. Additionally, EASE also exhibit an exceptional accuracy of 100\% on CR data while all other approaches perform nearly 90\% exept PromptBossting. Notably, PromptBossting approach along with RoBERTa-large pre-trained model underperform but with the increase of training data size its accuracy increase nearly 1\%.  Table \ref{Classification} exhibit performance of various approaches on Yelp review polarity, Yelp P, data. However, Yelp P fails to differentiate between approaches on performance basis, as nearly all methods achieve an accuracy above 90\%. A multiclass data of Yelp reviews, Yelp, is shown in Table \ref{Classification}. Although various approaches utilized Yelp data, their direct comparison is not possible since all approaches utilized different splits of data.\\
%%%%%%%%%%%%%%%%%%%%%%%%%%%%%%%%%%%%%%%%%%%%%%%%%%%%%%%%%%%%%%%%%%%%%%%%%%%%%%%%%%%%%%%%%%%%%%%%%%%%%
\subsubsection{Benchmark Datasets of Classification}

\begin{figure}{htbp}
\centering
\includegraphics[width=1\linewidth]{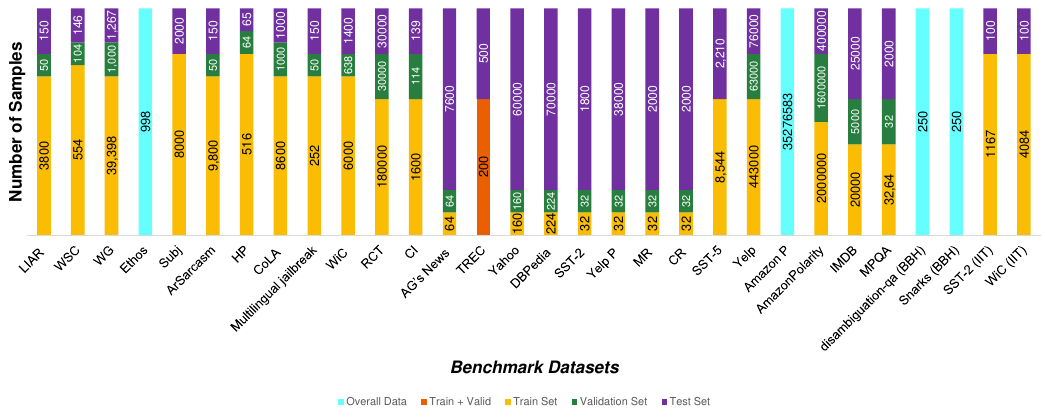}
\caption{Benchmark datasets of classification tasks}
\label{fig:1}
\end{figure}

Figure \ref{fig:1} represents a sample distribution of 30 different benchmark datasets used to evaluate classification tasks across different prompt optimization strategies. These datasets reflect a diverse range of classification tasks, including sentiment analysis, hate speech detection,  topic classification, and fake news detection. However, a critical analysis reveals several challenges that impact the fairness, consistency, and generalizability of evaluations. A critical concern is the considerable disproportion in dataset sizes, which ranges from small-scale benchmarks like ETHOS (998 samples) and TREC (700 samples) to massive datasets like Amazon Polarity (2M samples) and Yelp (443K samples). This imbalance raises concerns about whether prompt-optimization techniques are being fairly assessed across different training regimes  few-shot learning on small datasets versus large-scale fine-tuning on massive corpora. Furthermore, domain diversity remains a challenge, as general-purpose sentiment datasets (e.g., IMDB, Yelp, SST-2) dominate, while domain-specific datasets, such as RCT (biomedical classification) and CoLA (linguistic acceptability), are underrepresented. This hinders the ability to measure prompt adaptability in specialized domains. Hence, a more structured and standardized approach to dataset selection is essential. This includes ensuring balanced dataset sizes, complete data splits, and broader domain representation to facilitate a more reliable and interpretable evaluation of prompt optimization strategies.
\subsection{Question Answering Task}
Question answering tasks can be broadly categorized into two types based on how the answer is generated. These types include extractive question answering and abstractive question answering. Extractive question answering identifies and extracts a specific segment from the provided context to the question. While abstractive question answering generates an answer based on context and answer is not a direct copy of the text.\\ 

Table \ref{QA-Table} provides a comprehensive overview of 10 different prompt optimization strategies that have been evaluated for extractive or abstractive QA. Specifically, 4 different prompt optimization strategies have been evaluated for extractive QA, 9 different strategies are utilized for abstractive QA and only 3 distinct strategies have been employed for both extractive and abstractive QA. Furthermore, as shown in Table \ref{QA-Table}, 7 unique datasets have been used for extractive QA task evaluation, while 11 unique datasets have been utilized for abstractive QA task evaluation. Based on answer type these datasets fall into 11 distinct groups including entity completion, Text Segments, long and short text, text, True False (Binary), Multiple-choice, Date, Number, Factoid Questions, List Questions and Summary Questions. Lastly, the pretrained model stands out as the top-performing model, accompanied by its evaluation metric and performance, among all the LLMs used to assess a particular method, as detailed in Table \ref{QA-Table}.

A critical analysis of Table \ref{QA-Table} for extractive QA task reveals that larger models (e.g., T5-xxl vs. T5-base, DeBERTa-xlarge) generally achieve higher F1-scores across various methods and datasets which highlights the significance of model scaleability for QA. The methods such as DEPT and P-tuning v2 have been evaluated on numerous datasets and provide a more comprehensive assessment of model generalization and robustness. In contrast, approaches evaluated on fewer datasets may offer limited insights into a model's performance across diverse question types and domains. Furthermore, several prompt optimization strategies have utilized ReCoRD, NQ and SQuAD 1.1 for evaluation. However, direct performance comparisons across these datasets is challenging due to methodological differences. For instance, Prompt tuning and P-tuning v2 have been evaluated on SQuAD 1.1 dataset but using distinct evaluation measures. On the other hand, PRewriter and DEPT methods have used NQ dataset but with varying sample sizes. These inconsistencies in evaluation approaches and dataset splitting hinder direct comparisons of different methods. 
\scriptsize
% \usepackage{multirow}
\renewcommand{\arraystretch}{1}
\begin{longtable}{|p{0.57cm}|p{1.5cm}|p{1.35cm}|p{1.62cm}|p{1.6cm}|p{1.35cm}|p{1.5cm}|p{1.85cm}|}
\caption{Overview of performance evaluation of various QA approaches. ``*" symbol following a dataset name signifies the identical train/dev/test split was utilized for the reported results.}
\label{QA-Table}\\
\hline
\textbf{Task} & \textbf{Method} & \textbf{Data} & \textbf{Domain} & \textbf{\begin{tabular}[c]{@{}l@{}}Answer\\ Type\end{tabular}} & \textbf{Pre-Trained Model} & \textbf{Evaluation Metric} & \textbf{Performance} \\ \hline
\multirow{10}{*}{\rotatebox[origin=c]{90}{Extractive QA}} & \multirow{2}{*}{\begin{tabular}[c]{@{}l@{}}Prompt-\\ Tuning\end{tabular}} & \begin{tabular}[c]{@{}l@{}}ReCoRD\\ \cite{zhang2018record} *\end{tabular}   & \begin{tabular}[c]{@{}l@{}}News\\ CNN, \\Daily\\Mail\end{tabular} & \begin{tabular}[c]{@{}l@{}}Entity \\ Completion\end{tabular} & T5-xxl & F1 & 93.5 \\ \cmidrule(lr){3-8} 
&  &\begin{tabular}[c]{@{}l@{}}SQuAD\\ 1.1 dev\\ \cite{rajpurkar2016squad} *\end{tabular}   & Wikipedia Articles & \begin{tabular}[c]{@{}l@{}}Text\\ Segments\end{tabular} & T5-xxl & F1 mean & 94.8 \\ \cmidrule(lr){2-8} 
& \multirow{3}{*}{\begin{tabular}[c]{@{}c@{}}P-tuning  \\ V2 \end{tabular}} & \begin{tabular}[c]{@{}l@{}}SQuAD\\ 1.1 dev\\ \cite{rajpurkar2016squad} *\end{tabular} & Wikipedia Articles & \begin{tabular}[c]{@{}l@{}}Text\\ Segments\end{tabular} & DeBERTa-xlarge & F1 & 95.7 \\ \cmidrule(lr){3-8} 
&  & \begin{tabular}[c]{@{}l@{}}SQuAD\\ 2.0 dev\\ \cite{rajpurkar2016squad} \end{tabular} & Wikipedia Articles & \begin{tabular}[c]{@{}l@{}}Text\\ Segments\end{tabular} & DeBERTa-xlarge & F1 & 91.1 \\ \cmidrule(lr){3-8} 
&  & \begin{tabular}[c]{@{}l@{}}ReCoRD\\ \cite{zhang2018record} *\end{tabular} &\begin{tabular}[c]{@{}l@{}}News\\ CNN, \\Daily\\Mail\end{tabular} & \begin{tabular}[c]{@{}l@{}}Entity\\ Completion\end{tabular} & GLM-xxlarge & F1 & 92.5 \\ \cmidrule(lr){2-8} 
& PRewriter & NQ \cite{kwiatkowski2019natural} & Wikipedia & \begin{tabular}[c]{@{}l@{}}Long/Short\\ Text\end{tabular} & PaLM2-s & \begin{tabular}[c]{@{}l@{}}Exact\\ Match \\ Accuracy\end{tabular} & 30.2  \\ \cmidrule(lr){2-8} 
& \multirow{4}{*}{DEPT} & NQ \cite{kwiatkowski2019natural} & Wikipedia & \begin{tabular}[c]{@{}l@{}}Long/Short\\ Text\end{tabular} & T5-base & F1 & 73.2 \\ \cmidrule(lr){3-8} 
&  & HotpotQA \cite{yang2018hotpotqa} & Wikipedia & Long Text & T5-base & F1 & 76.8 \\ \cmidrule(lr){3-8} 
&  & SearchQA \cite{dunn2017searchqa}  & Web Snippets & Text & T5-base & F1 & 77.6 \\ \cmidrule(lr){3-8} 
&  & \begin{tabular}[c]{@{}l@{}}NewsQA \\ \cite{trischler2016newsqa}\end{tabular}  & \begin{tabular}[c]{@{}c@{}}News \\ Articles\end{tabular} & \begin{tabular}[c]{@{}l@{}} Text\\ Segments\end{tabular} & T5-base & F1 & 64.4 \\ \hline
\multirow{21}{*}{\rotatebox[origin=c]{90}{Abstractive QA}} & \multirow{8}{*}{\begin{tabular}[c]{@{}c@{}}Prompt- \\Tuning \end{tabular}} & \begin{tabular}[c]{@{}c@{}}BoolQ  \\ \cite{clark2019boolq} * \end{tabular} & Google Queries, Wikipedia & True/False & T5-xxl & Accuracy & 91.3 \\ \cmidrule(lr){3-8} 
&  & \begin{tabular}[c]{@{}l@{}}COPA \\ \cite{roemmele2011choice} *\end{tabular}   & \begin{tabular}[c]{@{}l@{}}Blogs,\\ Photography \\ Encyclopedia\end{tabular} & \begin{tabular}[c]{@{}l@{}}MCQs \\ (1/2)\end{tabular} & T5-xxl & Accuracy & 100 \\ \cmidrule(lr){3-8} 
&  & \begin{tabular}[c]{@{}l@{}}MultiRC \\ \cite{khashabi2018looking} \end{tabular}   & Various & MCQs & T5-xxl & F1 & 89 \\ \cmidrule(lr){3-8} 
&  & \begin{tabular}[c]{@{}l@{}} DuoRC \\ \cite{saha2018duorc}\end{tabular}   & \begin{tabular}[c]{@{}l@{}} Wikipedia \\and IMDb\\ Movie Plots\end{tabular} & Text & T5-xxl & F1 mean & 67.7 \\ \cmidrule(lr){3-8} 
&  & \begin{tabular}[c]{@{}l@{}} DROP  \\ \cite{dua2019drop}\end{tabular}  & \begin{tabular}[c]{@{}l@{}}Reasoning \\based \\Wikipedia \\ Passage \end{tabular} & \begin{tabular}[c]{@{}l@{}}Text, Date, \\ Number, \\ Others\end{tabular} & T5-xxl & F1 mean & 67.1 \\ \cmidrule(lr){3-8} 
&  & \begin{tabular}[c]{@{}l@{}}Textbook  \\ QA \cite{kembhavi2017you} \end{tabular}  & Science Topics & \begin{tabular}[c]{@{}l@{}}True/False, \\ MCQs\end{tabular} & T5-xxl & F1 mean & 66.8 \\ \cmidrule(lr){3-8} 
&  & \begin{tabular}[c]{@{}l@{}}BioASQ \\ \cite{BioASQ}\end{tabular}   & Biomedical & \begin{tabular}[c]{@{}l@{}}Questions: \\Yes/No, \\ Factoid \\ List  \\ Summary \end{tabular} & T5-xxl & F1 mean & 79.1 \\ \cmidrule(lr){3-8} 
&  & \begin{tabular}[c]{@{}l@{}} RACE \\ \cite{lai2017race}\end{tabular}   & English Exams & MCQs & T5-xxl & F1 mean & 60.7 \\ \cmidrule(lr){2-8} 
& \multirow{3}{*}{\begin{tabular}[c]{@{}c@{}}P-tuning  \\ V2 \end{tabular}} & \begin{tabular}[c]{@{}c@{}}BoolQ  \\ \cite{clark2019boolq} * \end{tabular} & Google Queries, Wikipedia & True/False  & GLM-xxlarge & Accuracy & 88.8 \\ \cmidrule(lr){3-8} 
&  & \begin{tabular}[c]{@{}l@{}}MultiRC \\ \cite{khashabi2018looking} *\end{tabular}  & Various & MCQs  & GLM-xxlarge & F1 & 88.1 \\ \cmidrule(lr){3-8} 
&  & \begin{tabular}[c]{@{}l@{}}COPA \\ \cite{roemmele2011choice} *\end{tabular} & \begin{tabular}[c]{@{}l@{}}Blogs,\\ Photography \\ Encyclopedia\end{tabular} & \begin{tabular}[c]{@{}l@{}}MCQs \\ (1/2)\end{tabular} & GLM-xxlarge & Accuracy & 98 \\ \cmidrule(lr){2-8} 
& Adv-ICL & \begin{tabular}[c]{@{}l@{}}COPA \\ \cite{roemmele2011choice} \end{tabular} & \begin{tabular}[c]{@{}l@{}}Blogs,\\ Photography \\ Encyclopedia\end{tabular} & \begin{tabular}[c]{@{}l@{}}MCQs \\ (1/2)\end{tabular} & GPT-3.5-turbo-0613 & Accuracy & 95.8 \\ \cmidrule(lr){2-8} 
& \multirow{2}{*}{DEPT} & \begin{tabular}[c]{@{}l@{}}MultiRC \\ \cite{khashabi2018looking} \end{tabular}  & Various & MCQs  & T5-base & F1 & 74.3 \\ \cmidrule(lr){3-8} 
&  & \begin{tabular}[c]{@{}c@{}}BoolQ  \\ \cite{clark2019boolq} \end{tabular} & Google Queries, Wikipedia & True/False  & T5-base & Accuracy & 79.3 \\ \cmidrule(lr){2-8} 
& \multirow{3}{*}{P-tuning} & \begin{tabular}[c]{@{}c@{}}BoolQ  \\ \cite{clark2019boolq} * \end{tabular} & Google Queries, Wikipedia & True/False  & GPT-2-Med & Accuracy & 78.9 \\ \cmidrule(lr){3-8} 
&  & \begin{tabular}[c]{@{}l@{}}MultiRC \\ \cite{khashabi2018looking} *\end{tabular} & Various & MCQs  & GPT-2-Med & Exact Match, F1 & \begin{tabular}[c]{@{}l@{}}29.3(EM), \\ 74.2 (F1)\end{tabular} \\ \cmidrule(lr){3-8} 
&  & \begin{tabular}[c]{@{}l@{}}COPA \\ \cite{roemmele2011choice} \end{tabular} & \begin{tabular}[c]{@{}l@{}}Blogs,\\ Photography \\ Encyclopedia\end{tabular} & \begin{tabular}[c]{@{}l@{}}MCQs \\ (1/2)\end{tabular} & GPT-2-Med & Accuracy & 81 \\ \cmidrule(lr){2-8} 
& Automate-CoT & \begin{tabular}[c]{@{}l@{}}OpenBook \\ QA \cite{mihaylov2018can}\end{tabular}    & Science Concept & \begin{tabular}[c]{@{}l@{}}MCQs \\ (1-4)\end{tabular} & GPT-3.5-turbo & Exact Match & 83 \\ \cmidrule(lr){2-8} 
& LoPT & \begin{tabular}[c]{@{}c@{}}BoolQ  \\ \cite{clark2019boolq} * \end{tabular}& Google Queries, Wikipedia & True/False  & T5-base & Accuracy & 76.5 (LoPT-1) \\ \cmidrule(lr){2-8} 
& APE & \begin{tabular}[c]{@{}l@{}}Truthful \\QA \cite{lin2021truthfulqa}\end{tabular}   & \begin{tabular}[c]{@{}l@{}}38 \\Categories\end{tabular} & True/False & InstructGPT & (\% True + \% Info) & 40\% \\ \cmidrule(lr){2-8} 
& \begin{tabular}[c]{@{}l@{}}Prompt- \\ AGENT\end{tabular}& \begin{tabular}[c]{@{}l@{}}MedQA \\ \cite{jin2021disease}\end{tabular}  & \begin{tabular}[c]{@{}l@{}}Medical \\ Board\\ Exams\end{tabular} & \begin{tabular}[c]{@{}l@{}}MCQs \\ (1-4)\end{tabular} & GPT-4 & Accuracy & 0.774 \\ \hline
\end{longtable}

\normalsize

The analysis of abstractive QA task reveals that in general methods employing larger models tend to dominate but dataset complexity significantly impacts performance. Tasks requiring causal reasoning, such as COPA, achieve near-perfect accuracy, indicating that models excel at structured decision-making. However, multi-step reasoning tasks like DROP and TextbookQA remain challenging. Biomedical and domain-specific datasets (BioASQ, MedQA) present difficulties which highlight that general-purpose LLMs might struggle with specialized knowledge even with optimized prompt. Contrarily, smaller models (GPT-2, T5-Base) consistently underperform on diverse dataset. Additionally, BoolQ, COPA and MultiRC datasets have been widely utilized across various prompt optimization strategies. Moreover, evaluation inconsistencies of different methods hinder direct comparisons and performance benchmarking.
%%%%%%%%%%%%%%%%%%%%%%%%%%%%%%%%%%%%%%%%%%%%%%%%%%%%%%%%%%%%%%%%%%%%%%%

\subsubsection{Benchmark Datasets of Question Answering Task}
\begin{figure}
\centering
\includegraphics[width=0.6\linewidth]{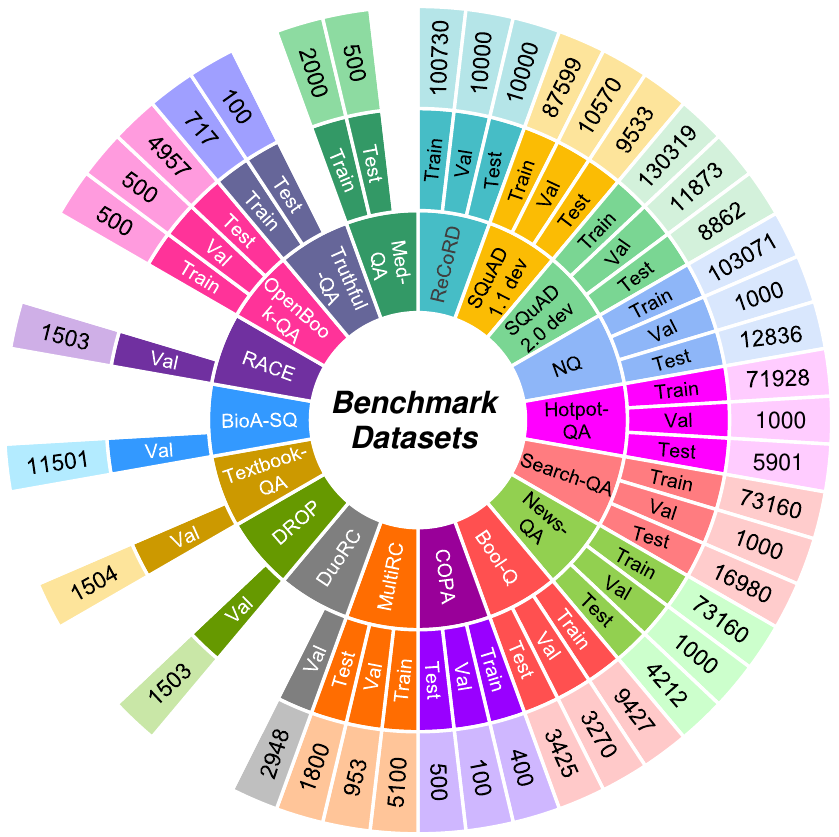}
\caption{Benchmark datasets of Question Answering tasks}
\label{QA-dataset-graph}
\end{figure}

Figure \ref{QA-dataset-graph} illustrates the sample distribution across eighteen benchmarking datasets used for extractive and abstractive Question Answering (QA) tasks. These datasets cover a diverse range of domains, including news, Wikipedia, biomedical, Google queries, and science. A notable observation is that the majority of datasets follow a structured train-validation-test split, ensuring comprehensive evaluation. Large-scale datasets such as ReCoRD (100,730 train, 10,000 val, 10,000 test), NQ (103,071 train, 1,000 val, 12,836 test), and SearchQA (73,160 train, 1,000 val, 16,980 test) provide ample training data, allowing models to learn robust patterns. In contrast, smaller datasets, such as COPA (400 train, 100 val, 500 test) and OpenBookQA (500 train, 500 val, 4,957 test), offer significantly fewer samples, which may limit model generalization and necessitate the use of few-shot or zero-shot learning approaches. A key inconsistency in the sample distribution is the size unbalance set in certain benchmarks. Datasets such as RACE (1,503 samples), BioA-SQ (11,501 samples), TextbookQA (1,504 samples), DROP (1,503 samples), and DuoRC (2,948 samples) only provide validation samples. 
Another interesting trend is the variation in test set sizes across datasets. While datasets like NQ (12,836 test samples) and SearchQA (16,980 test samples) have extensive evaluation sets, others such as TruthfulQA (717 train, 100 test samples) and MED-QA (2,000 train, 500 test samples) have significantly smaller test sets. This raises concerns about whether prompt optimization strategies are being fairly assessed across datasets of different scales.\\

\subsection{Natural Language Inference}
Table \ref{NLI-Table} presents a comprehensive performance evaluation of prompt optimization strategies used for Natural Language Inference (NLI) tasks, predominantly focusing on English language datasets, but also include a multilingual evaluation. It is evident from Table \ref{NLI-Table} that 18 distinct prompt optimization strategies have been evaluated against 8 unique dataset namely MNLI, QNLI, RTE, e-SNLI, CB, SciTail, SICK-E and XNLI. 
Table \ref{NLI-Table} shows that, on the RTE dataset, P-tuning v2 paired with the GLM-xxlarge pre-trained model achieves excellent results, surpassing 90\%. In contrast, BDPL combined with GPT-3 Davinci demonstrates the lowest performance at 57\%. P-tuning and LoPT perform moderately, with reported accuracy above 70\%. Moreover, various approaches, including BBT, BBTv2, CLAPS, StablePRompt, MAPO, and TEMPERA used different data split of RTE with reduced training and validation data. However, none achieve an accuracy above 90\% which highlights the importance of the size of the training and validation data. \\

MNLI is the second most widely used dataset across various approaches, with performance heavily influenced by training data size, similar to RTE. BDPL and DEPT utilize the largest training datasets, yet BDPL with GPT-3 Davinci struggles, whereas DEPT with T5-base achieves strong accuracy. Other approaches, including BBT, BBTv2, StablePRompt, MAPO, and TEMPERA, use smaller training and validation sets and test data. Despite significant variations in data sizes, EASE with GPT-3.5-turbo-1106 and CLAPS with Flan-T5large both achieve strong accuracy above 80\%. On CB data, almost all approaches exhibit significant performance above 90\%. Prompt Tuning paired with T5-xxl pre-trained model and P-tuning v2 accompanied by RoBERTa-large pre-trained model shows exceptional performance of accuracy 100\%. 

\scriptsize
% \usepackage{multirow}
\renewcommand{\arraystretch}{1}
\begin{longtable}{|p{0.52cm}|p{1.5cm}|p{1.9cm}|p{1.8cm}|p{1.48cm}|p{1.5cm}|p{1.85cm}|}
\caption{Overview of performance evaluation of various NLI approaches. ``*", ``**" and ``***" symbols following a dataset name signify that the identical train/dev/test split was utilized for the reported results.}
\label{NLI-Table}\\
\hline
\textbf{Task} & \textbf{Methods} & \textbf{Data} & \textbf{Domain} & \textbf{Pre-trained Model} & \textbf{Evaluation Metric} & \textbf{Performance} \\ \hline
\multirow{47}{*}{\rotatebox[origin=c]{90}{NLI (English)}} & \multirow{3}{*}{BDPL} & MNLI \cite{williams2017broad} * & Misc & GPT-3 Davinci & Accuracy & 54.6 \\ \cmidrule(lr){3-7} 
&  & QNLI \cite{wang2018glue} & Wikipedia & GPT-3 Davinci & Accuracy & 56.2 \\ \cmidrule(lr){3-7} 
&  & RTE \cite{giampiccolo2007third} * & News, Wikipedia & GPT-3 Davinci & Accuracy & 57.2 \\ \cmidrule(lr){2-7} 
& EASE & MNLI \cite{williams2017broad} ** & Misc & GPT-3.5-turbo-1106 & Accuracy & 81.7 \\ \cmidrule(lr){2-7} 
& Automate-CoT & e-SNLI \cite{blunsom2018snli} & Commonsense Knowledge & text-davinci-002 & Exact match & 86.4 \\ \cmidrule(lr){2-7} 
& \multirow{5}{*}{DEPT} & MNLI \cite{williams2017broad} * & Misc & T5-base & Accuracy & 85 \\ \cmidrule(lr){3-7} 
&  & QNLI \cite{wang2018glue} & Wikipedia & T5-base & Accuracy & 93.2 \\ \cmidrule(lr){3-7} 
&  & RTE \cite{giampiccolo2007third} & News, Wikipedia & T5-base & Accuracy & 79.1 \\ \cmidrule(lr){3-7} 
&  & CB \cite{de2019commitmentbank} & Various & T5-base & Accuracy & 92.9 \\ \cmidrule(lr){3-7} 
&  & SciTail \cite{khot2018scitail} & Science Exams & T5-base & Accuracy & 95.6 \\ \cmidrule(lr){2-7} 
& \multirow{2}{*}{\begin{tabular}[c]{@{}l@{}}Prompt- \\Tuning\end{tabular}} & CB \cite{de2019commitmentbank} * & Various & T5-xxl & Accuracy & 100 \\ \cmidrule(lr){3-7} 
&  & RTE \cite{giampiccolo2007third} & News, Wikipedia & T5-xxl & Accuracy & 93.5 \\ \cmidrule(lr){2-7} 
& \multirow{2}{*}{BBT} & RTE \cite{giampiccolo2007third} ** & News, Wikipedia & RoBERTa-large & Accuracy & 52.59 \\ \cmidrule(lr){3-7} 
&  & SNLI \cite{bowman2015large} * & Various & RoBERTa-large & Accuracy & 46.58 \\ \cmidrule(lr){2-7} 
& \multirow{4}{*}{CLAPS} & MNLI \cite{williams2017broad} & Misc & Flan-T5large & Accuracy & 81.81 (Genetics) \\ \cmidrule(lr){3-7} 
&  & QNLI \cite{wang2018glue} & Wikipedia & Flan-T5large & Accuracy & 86.47(Greedy) \\ \cmidrule(lr){3-7} 
&  & RTE \cite{giampiccolo2007third} ** & News, Wikipedia & Flan-T5large & Accuracy & 86.28 (Greedy) \\ \cmidrule(lr){3-7} 
&  & SNLI \cite{bowman2015large} * & Various & Flan-T5large & Accuracy & 84.75 (Greedy) \\ \cmidrule(lr){2-7} 
& \multirow{2}{*}{BBTv2} & RTE \cite{giampiccolo2007third} ** & News, Wikipedia & RoBERTa-large & Accuracy & 56.68 \\ \cmidrule(lr){3-7} 
&  & SNLI \cite{bowman2015large} * & Various & RoBERTa-large & Accuracy & 57.27 \\ \cmidrule(lr){2-7} 
& \begin{tabular}[c]{@{}l@{}}Prompt \\AGENT\end{tabular} & CB \cite{de2019commitmentbank} & Various & GPT-3.5 & Accuracy & 0.914 \\ \cmidrule(lr){2-7} 
& {\begin{tabular}[c]{@{}c@{}}AUTO\\PROMPT\end{tabular}} & SICK-E \cite{marelli2014semeval} & \begin{tabular}[c]{@{}l@{}}ImageFlickr,\\ STS \\ MSRVideo \\Description\end{tabular} & RoBERTa-large & Accuracy & 87.3, 69.3 \\ \cmidrule(lr){2-7} 
& \multirow{3}{*}{{\begin{tabular}[c]{@{}c@{}}Prompt\\ -BO\end{tabular}}} & MNLI \cite{williams2017broad} & Misc & RoBERTa-large & Accuracy & 29.6 \\ \cmidrule(lr){3-7} 
&  & QNLI \cite{wang2018glue} & Wikipedia & RoBERTa-large & Accuracy & 52.9 \\ \cmidrule(lr){3-7} 
&  & RTE \cite{giampiccolo2007third}  & News, Wikipedia & RoBERTa-large & Accuracy & 51 \\ \cmidrule(lr){2-7} 
& \multirow{2}{*}{LoPT} & CB \cite{de2019commitmentbank} * & Various & T5-base & Accuracy & 90.4 (LoPT-1), 74.0 (LoPT-2) \\ \cmidrule(lr){3-7} 
&  & RTE \cite{giampiccolo2007third}  * & News, Wikipedia & T5-base & Accuracy & 73.8 (LoPT-1), 74.3 (LoPT-2) \\ \cmidrule(lr){2-7} 
& \multirow{4}{*}{{\begin{tabular}[c]{@{}c@{}}Stable\\PRompt\end{tabular}}} & MNLI \cite{williams2017broad} *** & Misc & RoBERTa-large & Accuracy & 49.1 \\ \cmidrule(lr){3-7} 
&  & QNLI \cite{wang2018glue} * & Wikipedia & RoBERTa-large & Accuracy & 59.1 \\ \cmidrule(lr){3-7} 
&  & RTE \cite{giampiccolo2007third}  ** & News, Wikipedia & RoBERTa-large & Accuracy & 62.9 \\ \cmidrule(lr){3-7} 
&  & SNLI \cite{bowman2015large} * & Various & RoBERTa-large & Accuracy & 55.3 \\ \cmidrule(lr){2-7} 
& \multirow{4}{*}{MAPO} & MNLI \cite{williams2017broad} *** & Misc & RoBERTa-large & Accuracy & 55.7 \\ \cmidrule(lr){3-7} 
&  & QNLI \cite{wang2018glue} * & Wikipedia & RoBERTa-large & Accuracy & 63.1 \\ \cmidrule(lr){3-7} 
&  & RTE \cite{giampiccolo2007third}  ** & News, Wikipedia & RoBERTa-large & Accuracy & 75.3 \\ \cmidrule(lr){3-7} 
&  & SNLI \cite{bowman2015large} * & Various & RoBERTa-large & Accuracy & 60 \\ \cmidrule(lr){2-7} 
& \multirow{4}{*}{{\begin{tabular}[c]{@{}c@{}}Prompt\\ Boosting\end{tabular}} } & SNLI \cite{bowman2015large}* & Various & RoBERTa-large & Accuracy & 61.3, 62.0 \\ \cmidrule(lr){3-7} 
&  & MNLI \cite{williams2017broad} & Misc & RoBERTa-large & Accuracy & 52.5, 53.8 \\ \cmidrule(lr){3-7} 
&  & QNLI \cite{wang2018glue} & Wikipedia & RoBERTa-large & Accuracy & 58.0, 58.3 \\ \cmidrule(lr){3-7} 
&  & RTE \cite{giampiccolo2007third} & News, Wikipedia & RoBERTa-large & Accuracy & 60.0, 60.3 \\ \cmidrule(lr){2-7} 
& \multirow{2}{*}{\begin{tabular}[c]{@{}c@{}}P-tuning  \\ V2 \end{tabular}} & CB \cite{de2019commitmentbank} * & Various & RoBERTa-large & Accuracy & 100 \\ \cmidrule(lr){3-7} 
&  & RTE \cite{giampiccolo2007third} * & News, Wikipedia & GLM-xxlarge & Accuracy & 93.1 \\ \cmidrule(lr){2-7} 
& \multirow{2}{*}{P-tuning} & CB \cite{de2019commitmentbank} * & Various & GPT-2-Med & Accuracy & 98.2 \\ \cmidrule(lr){3-7} 
&  & RTE \cite{giampiccolo2007third} * & News, Wikipedia & BERT-large, GPT-2-Med. & Accuracy & 75.5 \\ \cmidrule(lr){2-7} 
& \multirow{4}{*}{TEMPERA} & RTE \cite{giampiccolo2007third} ** & News, Wikipedia & RoBERTa-large & Accuracy & 60.3 \\ \cmidrule(lr){3-7} 
&  & SNLI \cite{bowman2015large} & Various & RoBERTa-large & Accuracy & 56.4 \\ \cmidrule(lr){3-7} 
&  & QNLI \cite{wang2018glue} & Wikipedia & RoBERTa-large & Accuracy & 57.4 \\ \cmidrule(lr){3-7} 
&  & MNLI \cite{williams2017broad} ** & Misc & RoBERTa-large & Accuracy & 45.2 \\ \hline
\rotatebox{90}{NLI (Multilingual)} & CLAPS & \begin{tabular}[c]{@{}l@{}}XNLI \cite{conneau2018xnli} (  \\ German,\\ English, \\ Spanish, \\Bulgarian,\\ French,\\ Hindi, \\  Russian, \\Swahili, \\ Turkish\end{tabular} & Multiple Genre & Flan-T5large & Average Accuracy & 53.49 (Genetics) \\ \hline
\end{longtable}

%%%%%%%%%%%%%%%%%%%%%%%%%%%%%%%%%%%%%%%%%%%%%%%%%%%%%%%%%%%%%%%%%%%%%%%%%%%%%%%%%%%%%%%%%%%%%%%
%NLI
\normalsize

CLAPS accompanied by Flan-T5large model is evaluated on multilingual data. It ultilized multilingual data of 9 distinct languages including English. It exhibit a performance of nearly 53\%. While on particularly English data, it demonstrate accuracy above than 80\%. Although its performance can not be directly compared but it gives us a slight idea that it performs better on a single language data than multilingual.\\
DEPT, PromptBoosting and P-Tuning approaches are also evaluated on few-shot settings. DEPT paired with T5-base is evaluated on various few-shot settings including 4-shot, 16-shot and 32-shot. A positive correlation is evident between the number of shots and overall performance, as performance increases with an increase in the number of shots. DEPT with T5-base is evaluated on 4 distinct data namely, MNLI, QNLI, RTE and CB data. It demonstrate high performance on QNLI data followed by CB data. However, it exhibits lowest performance on MNLI data. P-tuning accompanied by ALBERT-xxlarge is also evaluated on RTE and CB data on 32-shot setting. DEPT illustrates better performance on RTE data than P-tuning. However, for CB data, P-tuning exhibits improvement in performance and performs better than DEPT. Performance of PromptBoosting is also assessed on RTE data on an entirely different setting. 
%%%%%%%%%%%%%%%%%%%%%%%%%%%%%%%%%%%%%%%%%%%%%%%%%%%%%%%%%%%%%%%%%%%%%%%%%%%%%%%%%%%%%%%%%%%%%%%
\begin{figure}
\centering
\includegraphics[width=1\linewidth]{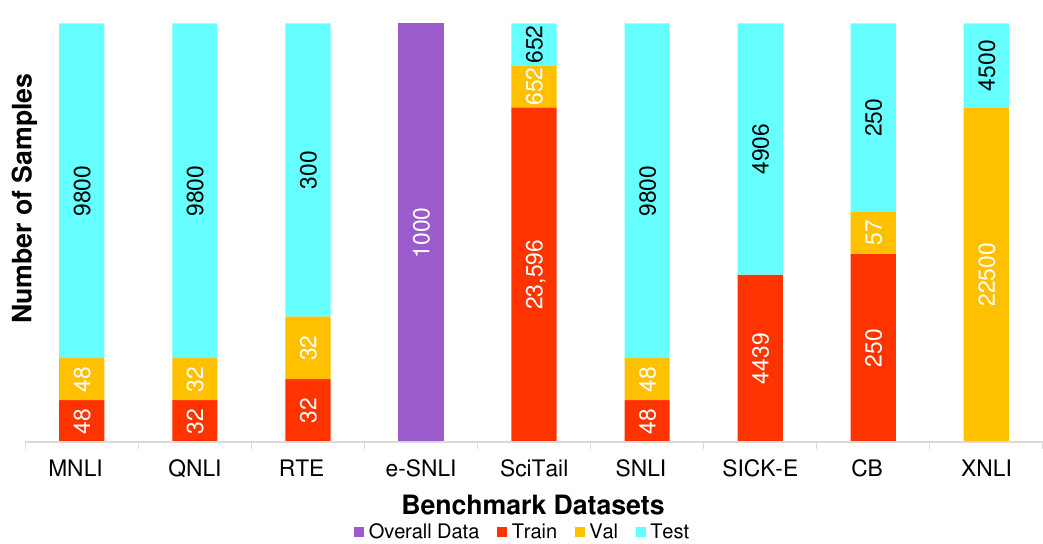}
\caption{Benchmark datasets of NLI tasks}
\label{NLI-Dataset}
\end{figure}

\subsubsection{Benchmark Datasets of NLI Tasks}
Figure \ref{NLI-Dataset} illustrates the sample distribution across nine benchmarking datasets used for Natural Language Inference (NLI) tasks. These datasets belong to diverse types of domains namely, commonsense knowledge, news, Wikipedia, etc.
A notable observation in dataset split variations is the disproportionate test set sizes in MNLI, SNLI, and QNLI, where the test sets contain 9,800 samples, significantly larger than the training and validation splits, which include only 48, 48, and 32 samples, respectively. A similar trend is observed in RTE, where the test set consists of 300 samples, while the training and validation sets collectively contain only 32 samples. These differences in samples size highlight that these datasets have been evaluated under a few-shot experimental setting across different prompt optimization strategies. In contrast, SciTail follows a different distribution, with a large training set of 23,596 samples, a validation set, and a test set, each containing 652 samples. SICK-E follows yet another distinct trend of almost an equal split between the training and test sets. The CB dataset maintains 250 samples each for training and testing, with a small validation set of 57 samples. XNLI dataset has been evaluated under zero-shot settings with a large validation set of 22,500 samples and a test set of 4,500 samples. The e-SNLI is the only dataset of overall 1000 samples. These varying dataset distributions highlight inconsistencies in data partitioning, which may impact the fairness and comparability of prompt optimization strategies across different NLI tasks.

\subsection{Natural Language Generation}
The Natural Language Generation (NLG) task is broadly categorized into 6 distinct tasks based on the type of text generation. These types include summarization, table-to text generation, text style transfer, machine translation, grammatical and structural reasoning or language transformation and common concepts. Furthermore, these tasks are divided into sub-tasks on the basis of text. Table \ref{NLG} provides a high-level overview of 12 unique optimization approaches and 17 distinct datasets for NLG task. An in-depth analysis of Table \ref{NLG} reveals that 3 distinct approaches EvoPrompt, Prefix-Tuning and Adv-ICL, with their best pre-trained model evaluated for 3 unique datasets including SAMSum, XSUM and CNN\textbackslash Daily Mail. Adv-ICL exhibits a higher ROUGE score on XSUM than CNN\textbackslash Daily Mail data, indicates that XSUM is comparatively more structured data.

For table-to-text generation task, prefix-tuning with GPT-2-large establishes a strong baseline, demonstrating high performance on the E2E dataset. However, for DART dataset, despite utilizing the same base model, prefix-tuning performs significantly worse, which highlights the importance of the specific data characteristics. While Adv-ICL leverages larger language models like GPT-3.5-turbo and text-davinci-002, its performance is inconsistent, which demonstrates that increasing model size doesn't guarantee improved results and may be sensitive to the dataset. Furthermore, the substantial performance differences observed between datasets like E2E and WebNLG underscore the influence of table and text characteristics on generation quality.
\scriptsize
% \usepackage{multirow}
\renewcommand{\arraystretch}{1}
\begin{longtable}{|p{0.68cm}|p{2cm}|p{1.8cm}|p{1.5cm}|p{1.5cm}|p{1.45cm}|p{1.5cm}|p{1.85cm}|}
\caption{Overview of performance evaluation of various NLG approaches. ``*" symbol following a dataset name signifies that the identical train/dev/test split was utilized for the reported results.}\label{NLG}\\
\hline
\textbf{Task} & \textbf{Sub-Task} & \textbf{Methods} & \textbf{Data} & \textbf{Domain} & \textbf{Pre-trained Model} & \textbf{Evaluation Metric} & \textbf{Performance} \\ \hline
\rotatebox{90}{Summarization} &  {\begin{tabular}[c]{@{}c@{}}Dialogue \\ Summarization \end{tabular}} & EvoPrompt & SAMSum \cite{gliwa2019samsum} & Various  & GPT-3.5 & ROUGE-1, ROUGE-2, ROUGE-L & 46.49, 19.49, 41.96 \\ \cmidrule(lr){2-8} 
& & Prefix-Tuning & XSUM \cite{narayan2018don} & {\begin{tabular}[c]{@{}c@{}}News \\ Articles\end{tabular}} & BART-large & ROUGE-1, ROUGE-2 ROUGE-L & news-to-sports (39.23, 16.74, 31.51) within-news (39.41, 16.87, 31.47) \\ \cmidrule(lr){3-8} 
& {\begin{tabular}[c]{@{}c@{}}Abstractive \\Summarization\end{tabular}} & \multirow{2}{*}{Adv-ICL} & CNN/Daily Mail \cite{nallapati2016abstractive} & {\begin{tabular}[c]{@{}c@{}}CNN, \\Daily\\ Mail\\Websites\end{tabular}} & text-davinci-002 & ROUGE-L & 23.4 \\ \cmidrule(lr){4-8} 
& &  & XSUM \cite{narayan2018don} & {\begin{tabular}[c]{@{}c@{}}BBC\\News\\ Articles\end{tabular}} & text-davinci-002 & ROUGE-L & 30.9 \\ \hline
& {\begin{tabular}[c]{@{}c@{}}Table-to-text \\Generation\end{tabular}} & \multirow{3}{*}{Prefix-Tuning} & E2E \cite{novikova2017e2e} & Restaurant Reviews & GPT-2-large & BLEU, NIST, MET, R-L, CIDEr & 70.3, 8.85, 46.2, 71.7, 2.47 \\ \cmidrule(lr){4-8} 
&  & & WebNLG \cite{gardent2017webnlg} & Various & GPT-2-large & BLEU, MET, TER & 64.11, 0.46, 0.33 \\ \cmidrule(lr){4-8} 
{\rotatebox{90}{\begin{tabular}[c]{@{}c@{}}Table-to-text \\Generation\end{tabular}}} & & & DART \cite{nan2020dart} & Wikipedia & GPT-2-large & BLEU,MET, TER, Mover, BERT, BLEURT & 50.84, 0.41, 0.43, 0.52, 0.95, 0.42 \\ \cmidrule(lr){3-8} 
& & {Adv-ICL} & E2E \cite{novikova2017e2e} & Restaurant Reviews & GPT 3.5-turbo-0613 & ROUGE-L & 51.1 \\ \cmidrule(lr){4-8} 
& &  & WebNLG \cite{gardent2017webnlg} & Various & text-davinci-002 & ROUGE-L & 65.4 \\ \hline
\rotatebox{90}{Text Style Transfer (TST)} & {\begin{tabular}[c]{@{}c@{}}Text \\Simplification\end{tabular}} & EvoPrompt & ASSET \cite{alva2020asset} & \_ & GPT-3.5 & SARI & 47.4 \\ \cmidrule(lr){2-8} 
& {\begin{tabular}[c]{@{}c@{}}Sentiment \\Style \\ Transfer\end{tabular}}& RLPrompt & Yelp \cite{shen2017style} & {\begin{tabular}[c]{@{}c@{}}Yelp \\Reviews\end{tabular}} & GPT-2-xl & Content, Style, Fluency, J(C, S, F), GM(C, S, F), BLEU, BERTScore, PPL & 72.1, 96.5(distilGPT-2), 91.6 (GPT-2-large),61.4, 84.7, 24.2, 59.0, 34.3 \\ \cmidrule(lr){2-8}
& {\begin{tabular}[c]{@{}c@{}}Author \\Style Transfer\end{tabular}} & RLPrompt & Shakespeare Plays \cite{xu2012paraphrasing} &  {\begin{tabular}[c]{@{}c@{}}Sparknotes\\ and Enotes \end{tabular}} & GPT-2-xl & Content, Style, Fluency, J(C, S, F), GM(C, S, F),BLEU, BERTScore, PPL & 51.8, 65.1,  85.2, 26.7,  66.0, 13.1, 39.0, 63.2 \\ \cmidrule(lr){2-8} 
& {\begin{tabular}[c]{@{}c@{}}Formality \\Style Transfer\end{tabular}} & {\begin{tabular}[c]{@{}c@{}}Prompt\\ BREEDER\end{tabular}}  & IIT \cite{honovich2022instruction} * & \_ & PaLM2-L & Execution Accuracy & 7 \\ \cmidrule(lr){3-8} 
& & APE & IIT \cite{honovich2022instruction} * & \_ & text-davinci-002 & Execution Accuracy & 70 (fewshot) \\ \cmidrule(lr){3-8} 
& & MoP & IIT \cite{honovich2022instruction} & \_ & \_ & \_ & \_ \\ \cmidrule(lr){3-8} 
& & PE2 & IIT \cite{honovich2022instruction} & \_ & GPT-4  (prompt proposal model) and text-davinci-003 (task model) & Accuracy & 61.26 \\ \cmidrule(lr){3-8} 
& & InstructZero & IIT \cite{honovich2022instruction} * & \_ & Vicuna Model & F1-Score & 0.63 \\ \cmidrule(lr){3-8} 
& & StablePRompt & IIT \cite{honovich2022instruction} * & \_ & Gemma-7B, InstructGPT3.5 & F1-Score Accuracy & 0.46, 58 \\ \hline 
\rotatebox{90}{Machine Translation} & {\begin{tabular}[c]{@{}c@{}}Single Word \\Translations\end{tabular}} & {\begin{tabular}[c]{@{}c@{}}Prompt\\ BREEDER\end{tabular}}  & Translation English-German, Translation English-Spanish, Translation English-French * & \_ & PaLM2-L & Execution Accuracy & 87, 91,91 \\ \cmidrule(lr){3-8} 
& & APE & Translation English-German, Translation English-Spanish, Translation English-French * & \_ & text-davinci-002 & Execution Accuracy & 86,91,90 (fewshot) \\ \cmidrule(lr){3-8} 
& & MoP & Translation English-German, Translation English-Spanish, Translation English-French & \_ & \_ & \_ & \_ \\ \cmidrule(lr){3-8} 
& & PE2 & Translation English-German, Translation English-Spanish, Translation English-French * & \_ & GPT-4 as prompt proposal model and text-davinci-003 as task model & Accuracy & 84.40, 85.40, 80.00 \\ \cmidrule(lr){3-8} 
& & EASE & Translation English-German, Translation English-Spanish, Translation English-French & \_ & GPT-3.5-turbo-1106 & Accuracy & 84.7, 89.3, 88.0 (with instructions) \\ \cmidrule(lr){3-8} 
& & InstructZero & Translation English-German, Translation English-Spanish, Translation English-French & \_ & Vicuna Model & Exact Match & 0.84,  0.87, 0.89 \\ \cmidrule(lr){3-8} 
& & StablePRompt & Translation English-German, Translation English-Spanish, Translation English-French * & \_ & InstructGPT-3.5 & Accuracy & 83, 89,  90 \\ \cmidrule(lr){2-8} 
& {\begin{tabular}[c]{@{}c@{}}Sentence \\Translations\end{tabular}} & {Adv-ICL} & LIRO (WMT-16-ro-en) \cite{dumitrescu2021liro} & News & text-davinci-002 & ROUGE-L & 81.2 \\ \cmidrule(lr){4-8} 
& &  & TED Talks \cite{qi2018and} & talks data & ChatGPT & ROUGE-L & 43.2 \\ \hline
\rotatebox{90}{\begin{tabular}[c]{@{}l@{}}Grammatical and \\ Structural Reasoning\\ \textbackslash{}Linguistic Transformation\end{tabular}} & Passivization & {\begin{tabular}[c]{@{}c@{}}Prompt\\ BREEDER\end{tabular}}  & IIT \cite{honovich2022instruction} * & \_ & PaLM2-L & Execution Accuracy & 100 \\ \cmidrule(lr){3-8} 
& & APE & IIT \cite{honovich2022instruction} * & \_ & text-davinci-002 & Execution Accuracy & 100 \\ \cmidrule(lr){3-8} 
& & MoP & IIT \cite{honovich2022instruction} & \_ & \_ & \_ & \_ \\ \cmidrule(lr){3-8} 
& & EASE & IIT \cite{honovich2022instruction} * & \_ & GPT-3.5-turbo-1106 & Accuracy & 100 \\ \cmidrule(lr){3-8} 
& & InstructZero & IIT \cite{honovich2022instruction} & \_ & Vicuna Model & Exact Match & 1 \\ \cmidrule(lr){3-8} 
& & StablePRompt & IIT \cite{honovich2022instruction} * & \_ & Gemma-7B, InstructGPT-3.5 & Exact Match, Accuracy & 1, 100 \\ \cmidrule(lr){2-8} 
& Negation & {\begin{tabular}[c]{@{}c@{}}Prompt\\ BREEDER\end{tabular}} & IIT \cite{honovich2022instruction} * & \_ & PaLM2-L & Execution Accuracy & 90 \\ \cmidrule(lr){3-8} 
& & APE & IIT \cite{honovich2022instruction} * & \_ & text-davinci-002 & Execution Accuracy & 90 (fewshot) \\ \cmidrule(lr){3-8} 
& & MoP & IIT \cite{honovich2022instruction} & \_ & \_ & \_ & \_ \\ \cmidrule(lr){3-8} 
&  & PE2 & IIT \cite{honovich2022instruction} & \_ & GPT-4 (prompt proposal model),text-davinci-003 (task model)  & Accuracy & 76 \\ \cmidrule(lr){3-8} 
& & EASE & IIT \cite{honovich2022instruction} * & \_ & GPT-3.5-turbo-1106 & Accuracy & 100 with instructions \\ \cmidrule(lr){3-8} 
& & InstructZero & IIT \cite{honovich2022instruction} & \_ & Vicuna Model & Exact Match & 0.8 \\ \cmidrule(lr){3-8} 
& & StablePRompt & IIT \cite{honovich2022instruction} * & \_ & Gemma-7B, InstructGPT-3.5 & Exact Match, Accuracy & 0.75, 84 \\ \cmidrule(lr){2-8} 
& Word Sorting & PE2 & BBH \cite{suzgun2022bigbenchhard} & \_ & GPT-3.5-turbo-instruct  (task model), GPT-4  turbo (prompt proposal model) & Accuracy & 66 \\ \cmidrule(lr){3-8} 
& & MoP & BBH \cite{suzgun2022bigbenchhard} & \_ & \_ & \_ & \_ \\ \cmidrule(lr){3-8} 
& & StablePRompt & BBH \cite{suzgun2022bigbenchhard} * & \_ & Gemma-7B & Exact Match & 100 \\ \cmidrule(lr){3-8} 
& & Adv-ICL & BBH \cite{suzgun2022bigbenchhard} * & \_ & GPT-3.5-turbo-061 & Exact Match &  55 \\ \cmidrule(lr){3-8} 
& & OPRO & BBH \cite{suzgun2022bigbenchhard} & \_ & PaLM2-L Scorer and PaLM2-L-IToptimizer & Accuracy & 54.5 \\ \cmidrule(lr){3-8} 
& & EvoPrompt & BBH \cite{suzgun2022bigbenchhard} * & \_ & GPT3.5 & Accuracy & 56.4 \\ \cmidrule(lr){3-8} 
& & APE & BBH \cite{suzgun2022bigbenchhard} * & \_ & text-davinci-002 & Normalized score & 30 \\ \hline
{\rotatebox{90}{\begin{tabular}[c]{@{}c@{}}Common \\Concept\end{tabular}}} & Common Concept & {\begin{tabular}[c]{@{}c@{}}Prompt\\ BREEDER\end{tabular}} & IIT \cite{honovich2022instruction} * & \_ & PaLM2-L & Execution Accuracy & 0 \\ \cmidrule(lr){3-8} 
& & APE & IIT \cite{honovich2022instruction} * & \_ & text-davinci-002 & Execution Accuracy & 32 (fewshot) \\ \cmidrule(lr){3-8} 
& & MoP & IIT \cite{honovich2022instruction} * & \_ & \_ & \_ & \_ \\ \cmidrule(lr){3-8} 
& & InstructZero & IIT \cite{honovich2022instruction} * & \_ & Vicuna Model & F1-Score & 0.15 \\ \cmidrule(lr){3-8} 
& & StablePRompt & IIT \cite{honovich2022instruction} * & \_ & Gemma-7B & F1-Score & 0.75 \\ \hline
\end{longtable}

\normalsize
%%%%%%%%%%%%%%%%%%%%%%%%%%%%%%%%%%%%%%%%%%%%%%%%%%%%%%%%%%%%%%%%%%%%%%%%%%%%%%%%%
%Trends

Moreover, text style transfer falls into 4 categories based on text style. It is evident from Table \ref{NLG} that 8 different optimization methods have been evaluated on 4 distinct datasets. RLPrompt along with GPT-2-xl model exhibit higher performance for sentiment style transfer task than author style. Specifically, IIT dataset has been frequently used by 6 approaches to evaluate their effectiveness across formality style transfer task. Although 4 among 6 prompt optimization strategies employed same dataset size but 2 methods including PROMPTBREEDER and APE have reported results on execution accuracy, 2 (InstructZero and StablePRompt) on F1-score, and 1 (StablePRompt) on accuracy. Hence, based on execution accuracy, APE outperform other methods using few shot approach. Similarly, for single word translations, APE emerges as top-performing model in terms of execution accuracy on few shot setting.

However, for sentence translation task, Adv-ICL demonstrate higher ROUGE score on domain specific dataset. Table \ref{NLG} shows that 9 distinct methods employed for 2 unique datasets for grammatical and structural reasoning task. Among all six different optimization approaches, all approaches perform well on passivization compared to negation. On the other hand, 7 distincts methods have used for word sorting task but with varying sample sizes. These inconsistencies in evaluation approaches and dataset splitting hinder direct comparisons of different methods. Moreover, for common concept task, IIT data has been frequently used by 5 approaches to evaluate their effectiveness. Although, dataset size for experiments in each method is same but 2 different evaluation metrics are reported including execution accuracy and F1-score. Hence, based on F1-score StablePRompt outperforms other approaches.
%%%%%%%%%%%%%%%%%%%%%%%%%%%%%%%%%%%%%%%%%%%%%%%%%%%%%%%%%%%%%%%%%%%%%%%%%%%%%%%%%

\subsubsection{Benchmark Datasets of NLG Tasks}

\begin{figure}
\centering
\includegraphics[width=0.5\linewidth]{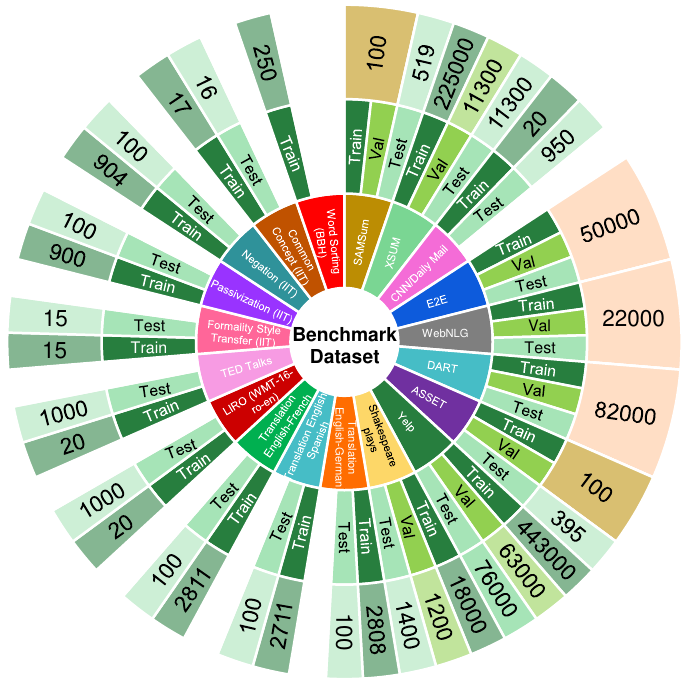}
\caption{Benchmark datasets of NLG tasks}
\label{NLG-datasets}
\end{figure}

Figure \ref{NLG-datasets} illustrates the sample distribution across nineteen benchmarking datasets used for Natural Language Generation (NLG) tasks. All of these tasks have been evaluated on a diverse array of datasets with varying sizes. A notable observation is the comprehensive evaluation of generation tasks across datasets of varying sizes. Small-scale benchmarks, such as Formality Style Transfer (IIT) with 30 samples and Common Concept with 33 samples, contrast sharply with large-scale datasets like XSUM (247,600 samples) and Yelp (582,000 samples). The machine translation task has been evaluated across multiple language pairs, including English-German (2,908 samples), English-French (2,911 samples), and English-Spanish (28,111 samples). Several datasets from the IIT benchmark, such as Negation, Common Concept, Formality Style Transfer, and Passivization have been used. Additionally, datasets like E2E, WebNLG, and DART, with overall sample sizes of 5,000, 22,000, and 82,000, respectively. These trends suggest that NLG tasks have been evaluated using large-scale datasets. The varying dataset sizes and domains ensure a comprehensive assessment of prompt optimization strategies. This approach enhances the robustness and generalizability of the evaluations, which provide insights into how models perform under different data constraints.
\subsection{Semantic Parsing}
\normalsize
Table \ref{Semantic-Table} represents a comprehensive overview of the prompt optimization strategies for two types of semantic parsing tasks: 1) Question Decomposition Meaning Representation 2) Task-Oriented Dialogue. Question decomposition meaning representation task involves breaking down complex natural language questions into structured logical forms. While task-oriented dialogue focuses on understanding user intents and mapping them to actions within predefined domains. ERP approach is evaluated using two different LLMs 1) inference model 2) scoring model. The inference model generates predictions from the input data while scoring model evaluates the quality or confidence of the inference LLM’s output by assigning a score. For all 3 datasets, the EPR approach employed the GPT-NEO model for scoring which ensures consistency in evaluation. However, GPT-J emerged as the top-performing inference LLM for the BREAK and SMCALFLOW datasets, which indicates its superior ability to handle logical form generation and structured dialogue tasks. On the other hand, CODEX outperformed other models as the leading inference LLM for the MTOP dataset, which suggests its effectiveness in task-oriented dialogue parsing. This analysis highlights that different LLMs excel in specific domains, which demonstrates the importance of selecting the right model based on task complexity and dataset characteristics. The performance metrics highlight the potential of the EPR approach in handling various semantic parsing challenges, from breaking down complex questions to understanding user intents in conversational contexts.\newline
\newline
\newline
\scriptsize
% \usepackage{multirow}
% \renewcommand{\arraystretch}{1}
\begin{longtable}{|p{1.85cm}|p{1.1cm}|p{2cm}|p{1.68cm}|p{1.8cm}|p{1.5cm}|p{1.85cm}|}
\caption{Overview of performance evaluation of various Semantic Parsing Tasks.}
\label{Semantic-Table}\\
\hline
\textbf{Sub-Task} & \textbf{Method} & \textbf{Data} & \textbf{Domain} & \textbf{Pre-trained Model} & \textbf{Evaluation Metric} & \textbf{Performance} \\ \hline
\begin{tabular}[c]{@{}l@{}}Question\\ Decomposition \\ Meaning\\ Representation\end{tabular} & \multirow{3}{*}{EPR} & BREAK \cite{wolfson2020break} & \begin{tabular}[c]{@{}l@{}}Multi-\\Domain\end{tabular} & \multirow{3}{*}{\begin{tabular}[c]{@{}l@{}}GPT-NEO \\as  inference \\and  scoring\\ LLM\end{tabular}} & \begin{tabular}[c]{@{}l@{}}LF-EM \\ (Logical\\ Form \\ Exact \\Match)\end{tabular} & 31.9 \\ \cmidrule(lr){1-1} \cmidrule(lr){3-4} \cmidrule(lr){6-7} 
\multirow{2}{*}{{\begin{tabular}[c]{@{}l@{}}Task \\ Oriented \\ Diaglouge\end{tabular}}} &  & MTOP \cite{li2020mtop} & \begin{tabular}[c]{@{}l@{}}Alarm, \\Calling,\\ Event, \\ Messaging, \\Music,News, \\ People,\\Recipes,\\ Reminder, \\ Timer,\\ Weather\end{tabular} &  & Exact Match (EM) & 64.2 \\ \cmidrule(lr){3-4} \cmidrule(lr){6-7} 
&  & SMCALFLOW \cite{andreas2020task} & \begin{tabular}[c]{@{}l@{}}Calendar,\\ Weather, \\ Places, \\People\end{tabular} &  & Exact Match (EM) & 54.3 \\ \hline
\end{longtable}

%%%%%%%%%%%%%%%%%%%%%%%%%%%%%%%%%%%%%%%%%%%%%%%%%%%%%%%%%%%%%%%%%%%%%%%%%%%%%%%%%%%%%%%%%%%%%%
%Trends
\normalsize

%%%%%%%%%%%%%%%%%%%%%%%%%%%%%%%%%%%%%%%%%%%%%%%%%%%%%%%%%%%%%%%%%%%%%%%%%%%%%%%%%%%%%%%%%%%%%%
\subsubsection{Benchmark Datasets of Semantic Parsing Tasks}

\begin{figure}
\centering
\includegraphics[width=1\linewidth]{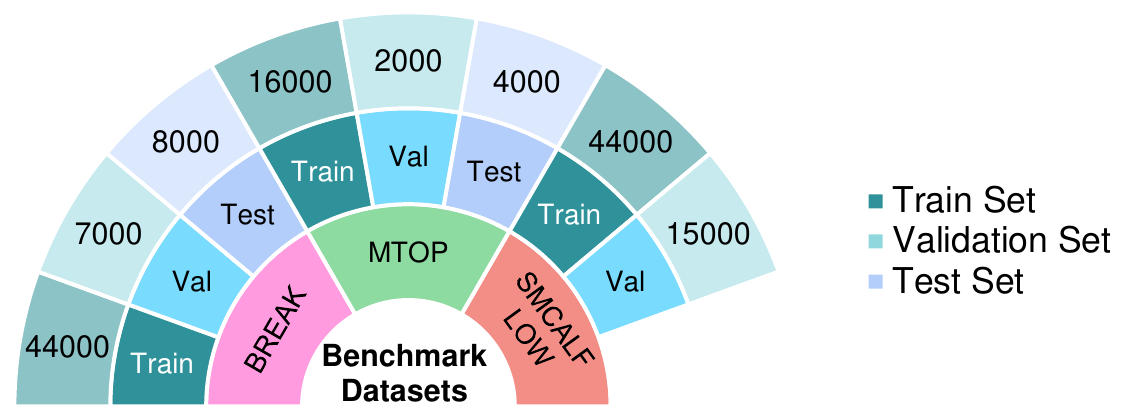}
\caption{Benchmark Datasets of Semantic Parsing Tasks}
\label{semantic-parsing-datasets}
\end{figure}

Figure \ref{semantic-parsing-datasets} represents the sample distribution of three benchmark datasets used to evaluate semantic parsing based tasks across different prompt optimization strategies. BREAK and MTOP datasets focus on Question Decomposition and Meaning Representation, which aids models break down complex queries into simpler sub-questions. Moreover, MTOP is a multilingual task-oriented semantic parsing dataset, which makes it valuable for understanding structured queries across multiple languages. On the other hand, the SMCALFLOW dataset involves  Task-Oriented Dialogue and primarily focuses on conversations related to scheduling, weather, and personal assistants. The varying dataset sizes indicate different levels of complexity, generalization challenges, and optimization needs, which are crucial for assessing the effectiveness of prompt-based semantic parsing methods across diverse NLP tasks. A key observation is the limited availability of benchmark datasets for semantic understanding, with only three existing in this domain. From the available benchmarks, the data split trends in the BREAK and MTOP datasets follow a similar pattern, though MTOP has a lower number of samples. Notably, the test split is absent in the SMCALFLOW benchmark, which may impact its use for comprehensive evaluation.
\subsection{Semantic Similarity based Tasks}
Table \ref{Similarity-Table} provides a comprehensive overview of 8 different prompt optimization strategies employed across 3 different types of similarity tasks including 1) sentence similarity, 2) paraphrase similarity and 3) question matching task. Sentence similarity computes the degree of semantic equivalence between two sentences, while paraphrase similarity determines whether two sentences convey the same meaning despite differences in words. In contrast, the question matching task evaluates whether two questions are semantically equivalent or not. It is evident from Table \ref{Similarity-Table} that 8 different prompt optimization strategies have been evaluated against 3 distinct datasets for sentence similarity task. In contrast, 8 different methods have been applied to 3 different datasets for paraphrase similarity. Specifically, STS-B dataset of IIT has been frequently used by 7 approaches to evaluate their effectiveness across sentence similarity task. Although, dataset size for experiments in each method is same but 4 methods have reported results on execution accuracy, 1 on mean accuracy, 1 on pearson co-relation and 1 on accuracy. Hence, based on execution accuracy PROMPTBREEDER outperforms other approaches using few shot approach. Moreover, DEPT method accompanied by T5-base pre-trained model, utilized a larger training data as compared to other approaches and demonstrated a high pearson correlation of above 90\% on STS-B dataset.

For the paraphrase similarity task, MRPC (Microsoft Research Paraphrase Corpus) is the most frequently used dataset followed by QQP (Quora Question Pairs). Contrarily, (PAWS) Paraphrase Adversaries from Word Scrambling is utilized exclusively by DEPT. It is evident from Table \ref{Similarity-Table} that all three approaches report their F1-Score ranges from 75\% to 78\%. Across MRPC dataset, DEPT (T5-base) achieves the highest accuracy (90.7\%) and outperforms other methods like EASE (81.7\%) and CLAPS (77.11\%). Similarly, for the QQP dataset, DEPT also achieves a high accuracy of 90.4\%, significantly surpassing models like GPT-3 Davinci (57.8\% F1-score) and RoBERTa-large (53.8\% F1-score). While BDPL performs well on MRPC (83.4\% F1-score), it struggles on QQP (57.8\% F1-score), suggesting dataset-specific strengths. Overall, DEPT emerges as the most effective approach for paraphrase similarity detection. For question matching task, BBTv2 method is employed in conjunction with CPM-2 for Chinese language. However, BBTv2 appears more effective for paraphrase similarity than for this question matching task, suggesting potential limitations in complexities of question semantics.
\scriptsize
\begin{longtable}{|p{0.57cm}|p{2cm}|p{2cm}|p{2.cm}|p{2.cm}|p{1.5cm}|p{1.85cm}|}
\caption{Overview of performance evaluation of various Similarity approaches. "*" symbol following a dataset name signifies that the identical train/dev/test split was utilized for the reported results.}
\label{Similarity-Table}\\
\hline
\textbf{Task} & \textbf{Method} & \textbf{Data} & \textbf{Domain} & \textbf{Pre-trained Model} & \textbf{Evaluation Metric} & \textbf{Performance} \\ \hline
\multirow{8}{*}{\rotatebox[origin=c]{90}{Sentence Similarly}} & PROMPTAGENT & BIOSSES \cite{souganciouglu2017biosses} & Biomedical & GPT-4 & Accuracy & 0.8 \\ \cmidrule(lr){2-7} 
& DEPT & STS-B \cite{cer2017semeval} & Misc. & T5-base & Pearson correlation & 90.8 \\ \cmidrule(lr){2-7} 
& {\begin{tabular}[c]{@{}c@{}}Prompt\\ BREEDER\end{tabular}}  & STS-B (IIT) \cite{honovich2022instruction} * & Misc. & PaLM2-L & Execution Accuracy & 56 (fewshot) \\ \cmidrule(lr){2-7} 
& APE & STS-B (IIT) \cite{honovich2022instruction} * & Misc. & text-davinci-002 & Execution Accuracy & 43 (fewshot) \\ \cmidrule(lr){2-7} 
& MoP & STS-B (IIT) \cite{honovich2022instruction} * & Misc. & GPT-3.5-turbo-Instruct model & Execution Accuracy &  \\ \cmidrule(lr){2-7} 
& PE2 & STS-B (IIT) \cite{honovich2022instruction} * & Misc. & text-davinci-003 & Mean score & 20 \\ \cmidrule(lr){2-7} 
& EASE & STS-B (IIT) \cite{honovich2022instruction} * & Misc. & GPT-3.5-turbo-1106 & Accuracy & 58.3 \\ \cmidrule(lr){2-7} 
& InstructZero & STS-B (IIT) \cite{honovich2022instruction} * & Misc. & Smaller Vicuna & Execution Accuracy & 0.19 \\ \hline
\multirow{13}{*}{\rotatebox[origin=c]{90}{Parahrase Similarity}} & \multirow{2}{*}{BDPL} & MRPC \cite{dolan2005automatically} & News & GPT-3 Davinci & F1-score & 83.4 \\ \cmidrule(lr){3-7} 
&  & QQP \cite{quora2017questionpairs} & Quora & GPT-3 Davinci & F1-score & 57.8 \\ \cmidrule(lr){2-7} 
& BBT & MRPC \cite{dolan2005automatically} * & News & RoBERTa-large & F1-score & 75.51 \\ \cmidrule(lr){2-7} 
& \multirow{2}{*}{CLAPS} & MRPC \cite{dolan2005automatically} * & News & Flan-T5large & Accuracy & 77.11 \\ \cmidrule(lr){3-7} 
&  & QQP \cite{quora2017questionpairs} & Quora & Flan-T5base & Accuracy & 81.51 \\ \cmidrule(lr){2-7} 
& BBTv2 & MRPC \cite{dolan2005automatically} * & News & RoBERTa-large & F1-score & 77.01 \\ \cmidrule(lr){2-7} 
& \multirow{2}{*}{{\begin{tabular}[c]{@{}c@{}}Prompt-\\ BO\end{tabular}}} & MRPC \cite{dolan2005automatically} * & News & RoBERTa-large & F1-score & 78.1 \\ \cmidrule(lr){3-7} 
&  & QQP \cite{quora2017questionpairs} & Quora & RoBERTa-large & F1-score & 53.8 \\ \cmidrule(lr){2-7} 
& StablePRompt & MRPC \cite{dolan2005automatically} & News & RoBERTa-large & Accuracy & 71.3 \\ \cmidrule(lr){2-7} 
& EASE & MRPC \cite{dolan2005automatically} * & News & GPT-3.5-turbo-1106 & Accuracy & 81.7 \\ \cmidrule(lr){2-7} 
& \multirow{3}{*}{DEPT} & MRPC \cite{dolan2005automatically} & News & T5-base & Accuracy & 90.7 \\ \cmidrule(lr){3-7} 
&  & QQP \cite{quora2017questionpairs} & Quora & T5-base & Accuracy & 90.4 \\ \cmidrule(lr){3-7} 
&  & PAWS  \cite{zhang2019paws} & Quora, Wikipedia & T5-base & Accuracy & 93.7 \\ \hline
\rotatebox{90}{Question Matching Task (Chinese)} & BBTv2 & LCQMC \cite{liu2018lcqmc} & \begin{tabular}[c]{@{}l@{}}Various \\ domains: \\ Daily Life, \\ Education, \\ Entertainment,\\ Computer\\ Games,\\ Social,\\ Natural\\ Science, \\ Sports \end{tabular} & CPM-2 & Accuracy & 59.1 \\ \hline
\end{longtable}

\normalsize

\begin{figure}
\centering
\includegraphics[width=1\linewidth]{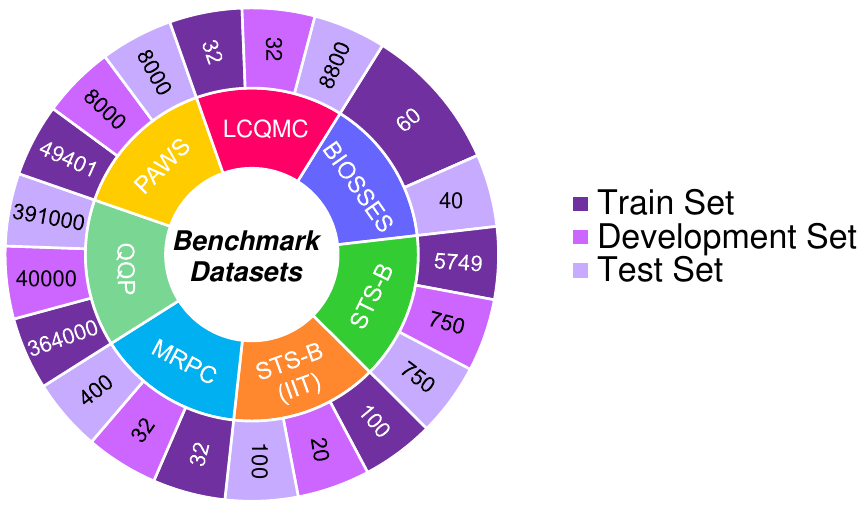}
\caption{Benchmark datasets of Similarity-based tasks}
\label{similariy-datasets}
\end{figure}

\subsubsection{Benchmark Datasets of Similarity-based Tasks}

Figure \ref{similariy-datasets} illustrates the sample distribution across seven benchmarking datasets used to assess prompt optimization strategies in similarity based tasks. Among these datasets, BIOSSES, STS-B, and STS-B (IIT) are used for sentence similarity tasks, while MRPC, QQP, and PAWS focus on paraphrase similarity. Noteably, only one dataset (LCQMC)  focuses on the Question Matching Task of Chinese data. A key insight from the dataset distribution is a considerable imbalance in dataset sizes, which ranges from small-scale benchmarks like BIOSSES (100 samples), STS-B (IIT) (220 samples), and MRPC (464 samples) to significantly larger datasets such as QQP (795,000 samples). This disparity in dataset size can introduce biases in prompt optimization evaluations, as models trained on smaller datasets may not generalize as effectively as those trained on larger ones. Consequently, this imbalance may impact the fairness and reliability of comparative assessments which potentially skews performance metrics towards datasets with more extensive training samples. Hence, it is crucial to ensure robust and equitable evaluations in semantic parsing based tasks.

Another notable trend is the substantial variation in test set sizes, with some datasets allocating unusually large number of samples for testing. For instance, LCQMC and QQP exhibit extreme discrepancies, with test sets of 8,800 and 391,000 samples, respectively, compared to training sets of only 32 and 3,564,000 samples and development sets of 32 and 40,000 samples. Such inconsistencies in data splits may influence model evaluation and generalization, potentially skewing performance comparisons across datasets.
\subsection{Information Extraction}
Information Extraction (IE)  focuses on automatically extracting structured information from unstructured textual data. Table \ref{Info-Table} presents a comparative performance analysis of various Information Extraction (IE) approaches across different tasks, sub-tasks, datasets, and domains.  The analysis focuses on understanding the strengths and weaknesses of different methods and the impact of factors like pre-trained models and data characteristics. For NER, P-tuning v2 illustrates competitive results across all 3 datasets. However, it demonstrates superior performance on CoNLL03 data followed by OntoNotes 5.0. It shows comparatively less micro F1-score on CoNLL04 data. One of the primary reasons behind performance difference is due to disparity in training data size. It is observed that P-tuning v2 performs better with a large amount of training data and vice versa. Notably, P-tuning v2 approach achieves its best performance using the DeBERTa-xlarge model on both the CoNLL03 and OntoNotes 5.0 datasets. However, for the CoNLL04 dataset, RoBERTa-large model emerges as the top-performing pre-trained model with the same approach.\\

To conduct a performance evaluation of relation extraction, 3 distinct approaches and 5 unique datasets are utilized. T-REx Extended data is employed for 2 different approaches namely AUTOPROMPT and Soft Prompt. It is evident from Table \ref{Info-Table}, AUTOPROMPT along with BERT perform better as compared to Soft Prompt with BERT-large pre-trained model. Moreover, soft prompt paired with BERT-large model is evaluated on 4 distinct datasets. Among all datasets, it demonstrates highest precision on T-REx Extended followed by T-REx Original. However, it exhibit worst precision on Google-RE and ConceptNet data, below 15\%. Moreover, for fact retrieval task, LAMA-TREx is utilized to assess the performance of P-tuning and AUTOPROMPT approach.  Table \ref{Info-Table} illustrates that P-tuning demonstrates high precision on 29k data size than 34k. However, AUTOPROMPT with BERT on LAMA-TREx (34k data) achieves lower precision compared to the P-tuning approach. Notably, AUTOPROMPT paired with BERT pre-trained model on T-REx Extendede data, achieves higher precision above 90\%  for relation extraction than fact retrieval task (less than 50\%). It indicates that AUTOPROMPT is more efficient to capture relations between text than extracting factual information.
\scriptsize
\begin{longtable}{|p{0.6cm}|p{2cm}|p{1.52cm}|p{2.5cm}|p{1.35cm}|p{1.5cm}|p{1.5cm}|p{1.85cm}|}
\caption{Overview of performance evaluation of various Information Extraction approaches. "*" symbol following a dataset name signifies the identical train/dev/test split was utilized for the reported results.}
\label{Info-Table}\\
\hline
\textbf{Task} & \textbf{Sub-Task} & \textbf{Method} & \textbf{Data} & \textbf{Domain} & \textbf{Pre-trained Model} & \textbf{Evaluation Metric} & \textbf{Performance} \\ \hline
\multirow{7}{*}{\rotatebox[origin=c]{90}{\begin{tabular}[c]{@{}c@{}}Token Level \\ Semantic Annotation\end{tabular}}}  & \multirow{4}{*}{{\begin{tabular}[c]{@{}c@{}}Named \\ Entity\\Recognition \end{tabular}} } & \multirow{3}{*}{\begin{tabular}[c]{@{}c@{}}P-tuning  \\ V2 \end{tabular}} & CoNLL03 \cite{sang2003introduction} & {\begin{tabular}[c]{@{}c@{}}Reuters\\ News\\ Stories \end{tabular}} & DeBERTa-xlarge & micro F1 & 93.1 \\ \cmidrule(lr){4-8} 
& &  & CoNLL04 \cite{carreras-marquez-2004-introduction} & News & RoBERTa-large & micro F1 & 88.4 \\ \cmidrule(lr){4-8} 
& &  & OntoNotes 5.0 \cite{weischedel2013ontonotes} & \begin{tabular}[c]{@{}l@{}}Newswire,\\ Broadcast \\ News and\\ Talks,\\ Telephone \\Talks, \\ Web Text\end{tabular} & DeBERTa-xlarge & micro F1 & 90.4 \\ \cmidrule(lr){3-8} 
&  & {\begin{tabular}[c]{@{}c@{}} PROMPT\\AGENT \end{tabular}} & NCBI \cite{dougan2014ncbi} & Medical & GPT-4 & F1-Score & 0.697 \\ \cmidrule(lr){2-8}
& \multirow{3}{*}{{\begin{tabular}[c]{@{}c@{}}Semantic \\ Role\\Labeling \end{tabular}} } & \multirow{3}{*}{\begin{tabular}[c]{@{}c@{}}P-tuning  \\ V2 \end{tabular}} & CoNLL12 \cite{pradhan2012conll} & \_  & DeBERTa-xlarge & micro F1 & 85.7 \\ \cmidrule(lr){4-8}
& &  & CoNLL05 WSJ \cite{carreras2005introduction} & News & DeBERTa-xlarge & micro F1 & 90.6 \\ \cmidrule(lr){4-8} 
& &  & CoNLL05 Brown \cite{carreras2005introduction} & News & DeBERTa-xlarge & micro F1 & 86.3 \\ \hline
\multirow{7}{*}{\rotatebox[origin=c]{90}{Relation Extraction}} & \multirow{6}{*}{{\begin{tabular}[c]{@{}c@{}}Relation \\ Extraction \end{tabular}} } & {\begin{tabular}[c]{@{}c@{}}AUTO \\ PROMPT \end{tabular}} & T-REx Extended \cite{shin2020autoprompt} * & \begin{tabular}[c]{@{}l@{}}Wikipedia \\Abstract, \\ Wikidata\end{tabular} & BERT & P@1 & 90.73 \\ \cmidrule(lr){3-8} 
& & \multirow{4}{*}{Soft Prompt} & T-REx Original \cite{elsahar2018t} & \begin{tabular}[c]{@{}l@{}}DBpedia \\Abstract, \\ Wikipedia \\Sentence\end{tabular} & BERT-large & P@1 & 51.9 \\ \cmidrule(lr){4-8} 
& &  & T-REx Extended \cite{shin2020autoprompt} * & \begin{tabular}[c]{@{}l@{}}Wikipedia\\ Abstract, \\ Wikidata\end{tabular} & BERT-large & P@1 & 52.5 \\ \cmidrule(lr){4-8} 
& &  & Google-RE \cite{google_relation_extraction} & Wikipedia & BERT-large-cased model & (P@1),  (P@10) ,(MRR) & 12.9, 34.7, 20.3 \\ \cmidrule(lr){4-8} 
& &  & ConceptNet \cite{speer2017conceptnet} & Wikipedia & BERT-large & (P@1), (P@10) ,(MRR) & 14.5, 38.6, 22.1 \\ \cmidrule(lr){3-8} 
& & BDPL & SciERC \cite{luan2018multi} & Computer Science & GPT-3 Davinci & F1-Score & 6.6 \\ \cmidrule(lr){2-8} 
& \begin{tabular}[c]{@{}l@{}}Relation\\ Extraction \\from \\ Reading\\ Comprehension\end{tabular}& Prompt-Tuning & RE \cite{levy2017zero} & Wikipedia & T5-xxl & mean F1 & 88.8 \\ \hline
\multirow{3}{*}{\rotatebox[origin=c]{90}{Fact Retrieval}} & \multirow{3}{*}{Fact Retrieval} & P-tuning & LAMA-TREx \cite{petroni2019language} & Wikidata & \begin{tabular}[c]{@{}l@{}}BERT\\-large, \\ MegatronLM\\ (11B)\end{tabular} & P@1 & \begin{tabular}[c]{@{}l@{}}50.6\\ (LAMA\\-34k), \\ 64.2 \\(LAMA\\-29k)\end{tabular} \\ \cmidrule(lr){3-8} 
& & \multirow{2}{*}{{\begin{tabular}[c]{@{}c@{}}AUTO \\ PROMPT \end{tabular}}} & LAMA-TREx \cite{petroni2019language} & Wikidata & BERT & (P@1),  (P@10) ,(MRR) & \begin{tabular}[c]{@{}l@{}}7 tokens\\ (43.34,\\  73.93,\\ 53.89)\end{tabular} \\ \cmidrule(lr){4-8} 
& &  & T-REx Extended \cite{shin2020autoprompt} * & \begin{tabular}[c]{@{}l@{}}Wikipedia\\ Abstract, \\ Wikidata\end{tabular} & BERT & (P@1),  (P@10) ,(MRR) & \begin{tabular}[c]{@{}l@{}}7 tokens\\ (45.57, \\72.02, \\ 54.89)\end{tabular} \\ \hline
\end{longtable}
\normalsize

%%%%%%%%%%%%%%%%%%%%%%%%%%%%%%%%%%%%%%%%%%%%%%%%%%%%%%%%%%%%%%%%%%%%%%%%%%%%%%%%%%%%%%%%%%%%%%%%%%%%%%%%%%%%%%%%%

\subsection{Linguistic and Semantic Understanding based Tasks}
Linguistic and semantic understanding task can be broadly categorized into Sentence and Word-Level Understanding based on the scope of the language unit being analyzed and the type of linguistic knowledge required for the task. Sentence-Level Understanding focus on comprehending the meaning of entire sentences or larger chunks of text. While word-Level Understanding focus on understanding the meaning and relationships of individual words. Table \ref{LSU} provides a comprehensive overview of 11 different prompt optimization strategies that have been used for sentence-level-understanding-based tasks. Table \ref{LSU} demonstrates that methods evaluated on the IIT dataset (PROMPTBREEDER, APE, PE2, EASE, InstructZero, StablePRompt) show lower performance compared to those on BBH. This suggests that the IIT dataset presents a greater challenge for sentence-level understanding, potentially due to differences in data characteristics or task complexity. The variety of metrics used on the IIT dataset (Exact Match, Exact Set, Accuracy) also makes direct comparison difficult. Performance on the Salient Translation Error Detection task (BBH) is considerably lower across all methods compared to the Hyperbaton task. This suggests that identifying translation errors is a more challenging problem. The significant drop in performance for PE2 and Adv-ICL on this sub-task is noteworthy. However, in most cases, direct performance comparisons across dataset is challenging due to varying sample sizes.

\scriptsize
% \usepackage{multirow}
% \renewcommand{\arraystretch}{1}
\begin{longtable}{|p{0.58cm}|p{1.57cm}|p{2cm}|p{1.6cm}|p{2.4cm}|p{1.5cm}|p{1.85cm}|}
\caption{Overview of performance evaluation of various Linguistic and Semantic Understanding approaches. ``*" symbol following a dataset name signifies the identical train/dev/test split was utilized for the reported results.}\label{LSU}\\
\hline
\textbf{Task} & \textbf{Sub-Task} & \textbf{Method} & \textbf{Data} & \textbf{Pre-trained Model} & \textbf{Evaluation Metric} & \textbf{Performance} \\ \hline
{\rotatebox{90}{\begin{tabular}[c]{@{}l@{}}Sentence-Level\\ Understanding\end{tabular}}} & Hyperbaton & OPRO & \begin{tabular}[c]{@{}l@{}}BBH \cite{suzgun2022bigbenchhard} \\**\end{tabular} & \begin{tabular}[c]{@{}l@{}}PaLM 2-L-IT\\Optimizer \\,PaLM 2-L \\scorer\end{tabular} & Accuracy & 96 \\ \cmidrule(lr){3-7}  
& & EvoPrompt & BBH \cite{suzgun2022bigbenchhard} * & GPT-3.5 & Accuracy & 81.2 \\ \cmidrule(lr){3-7} 
& & MoP & BBH \cite{suzgun2022bigbenchhard} & \_ & \_ & \_ \\ \cmidrule(lr){3-7} 
& & AEO & \begin{tabular}[c]{@{}l@{}}BBH \cite{suzgun2022bigbenchhard} \\***\end{tabular}& \begin{tabular}[c]{@{}l@{}}text-bison\\ task model \\ ,PaLM 2-L  \\ LLM mutator.\end{tabular} & Accuracy & 85.5 \\ \cmidrule(lr){3-7} 
& & StablePRompt & BBH \cite{suzgun2022bigbenchhard} * & Gemma-7B & Accuracy & 75.6 \\ \cmidrule(lr){3-7} 
& & PE2 & \begin{tabular}[c]{@{}l@{}}BBH \cite{suzgun2022bigbenchhard} \\****\end{tabular}& \begin{tabular}[c]{@{}l@{}}GPT-3.5-\\turbo-instruct \\ -task model,\\ GPT-4-turbo\\ -optimizer\end{tabular} & Accuracy & 90 \\ \cmidrule(lr){3-7} 
& & Adv-ICL & BBH \cite{suzgun2022bigbenchhard} * & GPT-3.5-turbo-0613 & Accuracy & 84 \\ \cmidrule(lr){2-7} 
& {Starting With} &{\begin{tabular}[c]{@{}c@{}}Prompt\\ BREEDER\end{tabular}}  & IIT \cite{honovich2022instruction} * & PaLM2-L & Exact Match & 71 \\ \cmidrule(lr){3-7} 
& & APE & IIT \cite{honovich2022instruction} * & text-davinci-002 & Exact Match & 69 (fewshot) \\ \cmidrule(lr){3-7} 
& & MoP & IIT \cite{honovich2022instruction}  & \_ & \_ & \_ \\ \cmidrule(lr){3-7} 
& & PE2 & IIT \cite{honovich2022instruction} & \begin{tabular}[c]{@{}l@{}}text-davinci-003 as \\ task model and \\ GPT-4 as optimizer\end{tabular} & mean & 67.6 \\ \cmidrule(lr){3-7} 
& & EASE & IIT \cite{honovich2022instruction} * & GPT-3.5-turbo-1106 & Accuracy & 81.7 \\ \cmidrule(lr){3-7} 
& & InstructZero & IIT \cite{honovich2022instruction} & Vicuna-13B & Exact Set & 0.51 \\ \cmidrule(lr){3-7} 
& & StablePRompt & IIT \cite{honovich2022instruction} * & \begin{tabular}[c]{@{}l@{}}Gemma-7B, \\ InstructGPT-3.5\end{tabular} & \begin{tabular}[c]{@{}l@{}}Exact Set, \\ Accuracy\end{tabular} & 0.375, 66 \\ \cmidrule(lr){2-7}
& {\begin{tabular}[c]{@{}c@{}}Salient\\ Translation \\ Error\\ Detection\end{tabular}} & OPRO & \begin{tabular}[c]{@{}l@{}}BBH \cite{suzgun2022bigbenchhard} \\**\end{tabular} & \begin{tabular}[c]{@{}l@{}}GPT-3.5-turbo \\ optimizer and \\ PaLM 2-L scorer\end{tabular} & Accuracy & 67.5 \\ \cmidrule(lr){3-7} 
& & EvoPrompt & BBH \cite{suzgun2022bigbenchhard} *& GPT-3.5 & Accuracy & 62.8 \\ \cmidrule(lr){3-7} 
& & MoP & BBH \cite{suzgun2022bigbenchhard} & \_ & \_ & \_ \\ \cmidrule(lr){3-7} 
& & AEO & \begin{tabular}[c]{@{}l@{}}BBH \cite{suzgun2022bigbenchhard} \\***\end{tabular} & \begin{tabular}[c]{@{}l@{}}text-bison task \\ model and  PaLM 2-L \\ LLM mutator.\end{tabular} & Accuracy & 58.7 \\ \cmidrule(lr){3-7} 
& & PE2 & \begin{tabular}[c]{@{}l@{}}BBH \cite{suzgun2022bigbenchhard} \\****\end{tabular}   & \begin{tabular}[c]{@{}l@{}}GPT-3.5-\\turbo-instruct \\ -task model, \\ GPT-4-turbo-\\ optimizer\end{tabular} & Accuracy & 36 \\ \cmidrule(lr){3-7} 
& & Adv-ICL & BBH \cite{suzgun2022bigbenchhard} * & GPT-3.5-turbo-061 & Accuracy & Less than 55 \\ \hline
\rotatebox{90}{\begin{tabular}[c]{@{}l@{}}Word-Level \\ Understanding\end{tabular}} & Antonyms & {\begin{tabular}[c]{@{}c@{}}Prompt\\ BREEDER\end{tabular}}  & IIT \cite{honovich2022instruction} * & PaLM2-L & Exact Match & 87 \\ \cmidrule(lr){3-7} 
& & APE & IIT \cite{honovich2022instruction} * & text-davinci-002 & Exact Match & 86 (fewshot) \\ \cmidrule(lr){3-7} 
& & MoP & IIT \cite{honovich2022instruction} & \_ & \_ & \_ \\ \cmidrule(lr){3-7} 
& & PE2 &\begin{tabular}[c]{@{}l@{}}IIT \cite{honovich2022instruction}\\ ******\end{tabular}  & \begin{tabular}[c]{@{}l@{}}text-davinci-003 as \\ task model and \\ GPT-4 as optimizer\end{tabular} & mean & 78.8 \\ \cmidrule(lr){3-7} 
& & EASE & IIT \cite{honovich2022instruction} * & GPT-3.5-turbo-1106 & Accuracy & 90 \\ \cmidrule(lr){3-7} 
& & InstructZero & \begin{tabular}[c]{@{}l@{}}IIT \cite{honovich2022instruction}\\ *******\end{tabular}  & Vicuna-13B & Exact Match & 0.89 \\ \cmidrule(lr){3-7} 
& & StablePRompt & IIT \cite{honovich2022instruction} * & \begin{tabular}[c]{@{}l@{}}Gemma-7B, \\ InstructGPT3.5\end{tabular} & \begin{tabular}[c]{@{}l@{}}Exact Match, \\ Accuracy\end{tabular} & 0.75,85 \\ \cmidrule(lr){2-7} 
& Pluralization & {\begin{tabular}[c]{@{}c@{}}Prompt\\ BREEDER\end{tabular}} & \begin{tabular}[c]{@{}l@{}}IIT \cite{honovich2022instruction}\\ **\end{tabular}  & PaLM2-L & Exact Match & 100 \\ \cmidrule(lr){3-7} 
& & APE & \begin{tabular}[c]{@{}l@{}}IIT \cite{honovich2022instruction}\\ **\end{tabular}  & text-davinci-002 & Exact Match & 100 \\ \cmidrule(lr){3-7} 
& & MoP & IIT \cite{honovich2022instruction} & \_ & \_ & \_ \\ \cmidrule(lr){3-7} 
& & EASE & IIT \cite{honovich2022instruction} & GPT-3.5-turbo-1106 & Accuracy & 100 \\ \cmidrule(lr){3-7} 
& & InstructZero & \begin{tabular}[c]{@{}l@{}}IIT \cite{honovich2022instruction}\\ *******\end{tabular}  & Vicuna-13B & Exact Match & 1 \\ \cmidrule(lr){3-7} 
& & StablePRompt & \begin{tabular}[c]{@{}l@{}}IIT \cite{honovich2022instruction}\\ **\end{tabular}  & \begin{tabular}[c]{@{}l@{}}Gemma-7B, \\ InstructGPT3.5\end{tabular} & \begin{tabular}[c]{@{}l@{}}Exact Match, \\ Accuracy\end{tabular} & 1, 99 \\ \cmidrule(lr){2-7}
& First Letter  & {\begin{tabular}[c]{@{}c@{}}Prompt\\ BREEDER\end{tabular}} & \begin{tabular}[c]{@{}l@{}}IIT \cite{honovich2022instruction}\\ ***\end{tabular}  & PaLM2-L & Exact Match & 100 \\ \cmidrule(lr){3-7} 
& & APE & \begin{tabular}[c]{@{}l@{}}IIT \cite{honovich2022instruction}\\ ***\end{tabular}  & text-davinci-002 & Exact Match & 100 \\ \cmidrule(lr){3-7} 
& & MoP & IIT \cite{honovich2022instruction} & \_ & \_ & \_ \\ \cmidrule(lr){3-7} 
& & EASE & \begin{tabular}[c]{@{}l@{}}IIT \cite{honovich2022instruction}\\ ***\end{tabular}  & GPT-3.5-turbo-1106 & Accuracy & 100 \\ \cmidrule(lr){3-7} 
& & InstructZero & \begin{tabular}[c]{@{}l@{}}IIT \cite{honovich2022instruction}\\ *******\end{tabular}  & Vicuna-13B & Exact Match & 1 \\ \cmidrule(lr){3-7} 
& & StablePRompt & \begin{tabular}[c]{@{}l@{}}IIT \cite{honovich2022instruction}\\ ***\end{tabular}  & \begin{tabular}[c]{@{}l@{}}Gemma-7B, \\ InstructGPT-3.5\end{tabular} & \begin{tabular}[c]{@{}l@{}}Exact Match, \\ Accuracy\end{tabular} & 0.937,100 \\ \cmidrule(lr){2-7}
& Second Letter & {\begin{tabular}[c]{@{}c@{}}Prompt\\ BREEDER\end{tabular}}  & \begin{tabular}[c]{@{}l@{}}IIT \cite{honovich2022instruction}\\ ***\end{tabular}  & PaLM2-L & Exact Match & 95 \\ \cmidrule(lr){3-7} 
& & APE & \begin{tabular}[c]{@{}l@{}}IIT \cite{honovich2022instruction}\\ ***\end{tabular}  & text-davinci-002 & Exact Match & 87 (zeroshot) \\ \cmidrule(lr){3-7} 
& & MoP & IIT \cite{honovich2022instruction} & \_ & \_ & \_ \\ \cmidrule(lr){3-7} 
& & PE2 & \begin{tabular}[c]{@{}l@{}}IIT \cite{honovich2022instruction}\\ ******\end{tabular}  & \begin{tabular}[c]{@{}l@{}}text-davinci-003 as \\ task model and \\ GPT-4 as optimizer\end{tabular} & mean & 94.2 \\ \cmidrule(lr){3-7} 
& & EASE & \begin{tabular}[c]{@{}l@{}}IIT \cite{honovich2022instruction}\\ ***\end{tabular}  & GPT-3.5-turbo-1106 & Accuracy & 100 \\ \cmidrule(lr){3-7} 
& & InstructZero &  \begin{tabular}[c]{@{}l@{}}IIT \cite{honovich2022instruction}\\ *******\end{tabular}   & Vicuna-13B & Exact Match & 0.62 \\ \cmidrule(lr){3-7} 
& & StablePRompt & \begin{tabular}[c]{@{}l@{}}IIT \cite{honovich2022instruction}\\ ***\end{tabular}  & \begin{tabular}[c]{@{}l@{}}Gemma-7B, \\ InstructGPT-3.5\end{tabular} & \begin{tabular}[c]{@{}l@{}}Exact Match, \\ Accuracy\end{tabular} & 0.187, 100 \\ \cmidrule(lr){2-7}
& List Letter & {\begin{tabular}[c]{@{}c@{}}Prompt\\ BREEDER\end{tabular}}  & \begin{tabular}[c]{@{}l@{}}IIT \cite{honovich2022instruction}\\ ***\end{tabular}  & PaLM2-L & Exact Match & 99 \\ \cmidrule(lr){3-7} 
& & APE & \begin{tabular}[c]{@{}l@{}}IIT \cite{honovich2022instruction}\\ ***\end{tabular}  & text-davinci-002 & Exact Match & 100 (fewshot) \\ \cmidrule(lr){3-7} 
& & MoP & IIT \cite{honovich2022instruction} & \_ & \_ & \_ \\ \cmidrule(lr){3-7} 
& & InstructZero & \begin{tabular}[c]{@{}l@{}}IIT \cite{honovich2022instruction}\\ *******\end{tabular}  & Vicuna-13B & Exact Match & 1 \\ \cmidrule(lr){3-7} 
& & StablePRompt & \begin{tabular}[c]{@{}l@{}}IIT \cite{honovich2022instruction}\\ ***\end{tabular}  & \begin{tabular}[c]{@{}l@{}}Gemma-7B, \\ InstructGPT-3.5\end{tabular} & \begin{tabular}[c]{@{}l@{}}ExactMatch, \\ Accuracy\end{tabular} & 0.875, 100 \\ \cmidrule(lr){2-7}
& Synonyms & {\begin{tabular}[c]{@{}c@{}}Prompt\\ BREEDER\end{tabular}}  & \begin{tabular}[c]{@{}l@{}}IIT \cite{honovich2022instruction}\\ ****\end{tabular}  & PaLM2-L & Exact Match & 43 \\ \cmidrule(lr){3-7} 
& & APE & \begin{tabular}[c]{@{}l@{}}IIT \cite{honovich2022instruction}\\ ****\end{tabular}  & text-davinci-002 & Exact Match & 22 (zeroshot) \\ \cmidrule(lr){3-7} 
& & MoP & IIT \cite{honovich2022instruction} & \_ & \_ & \_ \\ \cmidrule(lr){3-7} 
& & PE2 & \begin{tabular}[c]{@{}l@{}}IIT \cite{honovich2022instruction}\\ ******\end{tabular}  & \begin{tabular}[c]{@{}l@{}}text-davinci-003 as\\ task model and \\ GPT-4 as optimizer\end{tabular} & mean & 27.8 \\ \cmidrule(lr){3-7} 
& & EASE & \begin{tabular}[c]{@{}l@{}}IIT \cite{honovich2022instruction}\\ ****\end{tabular}  & GPT-3.5-turbo-1106 & Accuracy & 31.7 \\ \cmidrule(lr){3-7} 
& & InstructZero & \begin{tabular}[c]{@{}l@{}}IIT \cite{honovich2022instruction}\\ *******\end{tabular}  & Vicuna-13B & Contain & 0.38 \\ \cmidrule(lr){3-7} 
& & StablePRompt & \begin{tabular}[c]{@{}l@{}}IIT \cite{honovich2022instruction}\\ ****\end{tabular}  & \begin{tabular}[c]{@{}l@{}}Gemma-7B, \\ InstructGPT-3.5\end{tabular} & \begin{tabular}[c]{@{}l@{}}Contain, \\ Accuracy\end{tabular} & 0.125,43 \\ \cmidrule(lr){2-7}
& Rhymes & {\begin{tabular}[c]{@{}c@{}}Prompt\\ BREEDER\end{tabular}}  & \begin{tabular}[c]{@{}l@{}}IIT \cite{honovich2022instruction}\\ *****\end{tabular}  & PaLM2-L & Exact Match & 100 \\ \cmidrule(lr){3-7} 
& & APE & \begin{tabular}[c]{@{}l@{}}IIT \cite{honovich2022instruction}\\ *****\end{tabular}  & text-davinci-002 & Exact Match & 100 (zeroshot) \\ \cmidrule(lr){3-7} 
& & MoP & IIT \cite{honovich2022instruction} & \_ & \_ & \_ \\ \cmidrule(lr){3-7} 
& & PE2 & \begin{tabular}[c]{@{}l@{}}IIT \cite{honovich2022instruction}\\ ******\end{tabular}  & \begin{tabular}[c]{@{}l@{}}text-davinci-003 as \\ task model and \\ GPT-4 as optimizer\end{tabular} & mean & 65 \\ \cmidrule(lr){3-7} 
& & EASE & \begin{tabular}[c]{@{}l@{}}IIT \cite{honovich2022instruction}\\ *****\end{tabular} & GPT-3.5-turbo-1106 & Accuracy & 100 \\ \cmidrule(lr){3-7} 
& & InstructZero & \begin{tabular}[c]{@{}l@{}}IIT \cite{honovich2022instruction}\\ *******\end{tabular}  & Vicuna-13B & Exact Match & 0.46 \\ \cmidrule(lr){3-7} 
& & StablePRompt & \begin{tabular}[c]{@{}l@{}}IIT \cite{honovich2022instruction}\\ *****\end{tabular}  & \begin{tabular}[c]{@{}l@{}}Gemma-7B, \\ InstructGPT-3.5\end{tabular} & \begin{tabular}[c]{@{}l@{}}ExactMatch, \\ Accuracy\end{tabular} & \begin{tabular}[c]{@{}l@{}}0.0625, \\ 95 \end{tabular} \\ \hline
% \end{tabular}}
\end{longtable}

\normalsize
%%%%%%%%%%%%%%%%%%%%%%%%%%%%%%%%%%%%%%%%%%%%%%%%%%%%%%%%%%%%%%%%%%%%%%%%%%%%%%%%
%Trends

Furthermore, the use of different evaluation metrics (Accuracy, Exact Match, Exact Set) across datasets and even within the same task (IIT) makes it difficult to directly compare the effectiveness of the various methods. Standardized evaluation metrics are crucial for future research. Word-level understanding further categorized into 7 tasks on the basis of individual words and their attributes. The analysis of word-level understanding tasks reveals that morphological tasks like Pluralization, First Letter, Second Letter, and Rhymes are generally well-handled, with many methods (PROMPTBREEDER, APE, EASE, StablePRompt, InstructZero) achieving near-perfect or perfect scores which demonstrates the relative ease of these tasks for LLMs. PaLM2-L and GPT-3.5-turbo-1106 consistently emerge as strong performers across most tasks, particularly the morphological ones which highlights their effectiveness in word-level understanding. APE also shows strong performance, though generally slightly below PROMPTBREEDER and EASE. However, semantic tasks like Antonyms and Synonyms pose a significantly greater challenge, with much lower performance scores which shows the difficulty of capturing semantic relationships. PE2 exhibits varied performance, while InstructZero with Vicuna-13B shows mixed results, excelling on some morphological tasks but struggling on others. StablePRompt demonstrates consistent, though not top-tier, performance. The use of ``Contain" as a metric for Synonyms, instead of ``Exact Match," acknowledges the difficulty of perfect synonym identification. Furthermore, most of the approaches underperform for synonyms task, suggesting potential limitations and complexities.
%%%%%%%%%%%%%%%%%%%%%%%%%%%%%%%%%%%%%%%%%%%%%%%%%%%%%%%%%%%%%%%%%%%%%%%%%%%%%%%%
\subsubsection{Benchmark Datasets of Semantic and Linguistic Understanding based Tasks}

\begin{figure}
\centering
\includegraphics[width=1\linewidth]{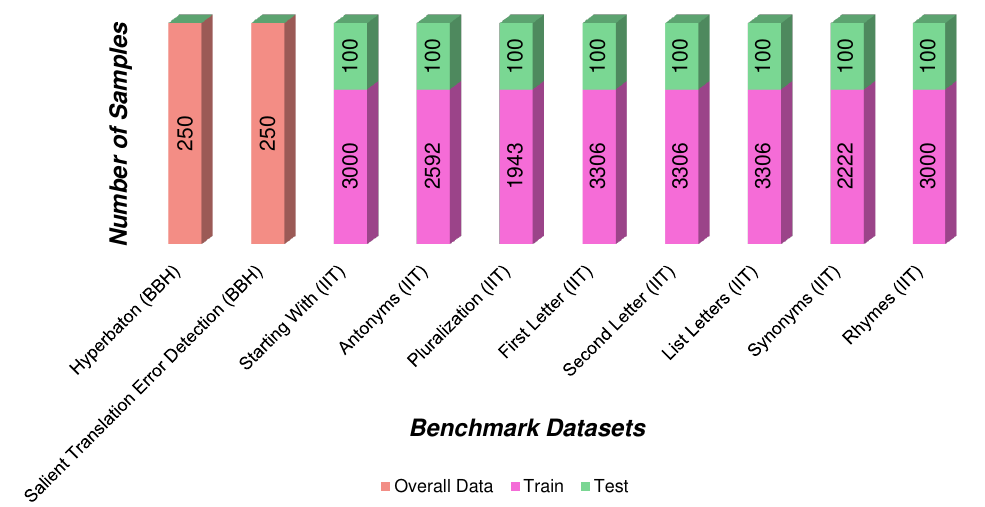}
\caption{Benchmark datasets of linguistic and semantic understanding based tasks}
\label{linguistic-semantic-dataset}
\end{figure}

Figure \ref{linguistic-semantic-dataset} illustrates the sample distribution across 10 benchmarking datasets used in linguistic and semantic understanding based tasks. There are primarily two benchmarks available for these tasks, BBH and IIT, each containing multiple datasets domains. From the BBH benchmark, the Hyperbaton and Salient Translation Error Detection datasets are included, both with an overall sample size of 250. In contrast, the remaining eight datasets belong to IIT, with sample sizes ranging from 2,043 for Pluralization to 3,406 for Second Letter, List Letters, and First Letter datasets. Among these datasets, Hyperbaton (BBH), Starting With (IIT), and Salient Translation Error Detection (BBH) focus on sentence-level understanding. In contrast, Antonyms (IIT), Pluralization (IIT), First Letter (IIT), Second Letter (IIT), List Letters (IIT), Synonyms (IIT), and Rhymes (IIT) involve word-level understanding. Moreover, Hyperbaton (BBH) has two options for selection, whereas Salient Translation Error Detection (BBH) offers six labels. A notable observation is the consistent test set size of 100 samples across all IIT datasets, including Antonyms, Synonyms, and Rhymes, ensuring uniformity in evaluation, and supporting fair performance comparison across different tasks.
\subsection{Knowledge and Contextual Understanding}
Table \ref{KCU} provides a high-level overview of 12 distinct prompt optimization strategies employed for knowledge-based task and contextual understanding task. Particularly, BBH dataset is employed to evaluate 3 distinct sub-tasks of knowledge-based including sports understanding, penguins in a table and ruin names. Table \ref{KCU} shows that EvoPrompt combined with GPT-3.5 pre-trained model achieves excellent results, by scoring around 97\% accuracy for sports understanding. However, for ruin names, despite utilizing the same base model, EvoPrompt performs significantly worse which highlights its task-specific adaptability. On the other hand, STablePRompt paired with Gemma-7B exhibits worst performance despite using a large amount of training data. Although there exist significant variations in data sizes, still PROMPTAGENT with GPT-4 achieves an excellent accuracy above 95\%. IIT is the second most widely used dataset utilized to evaluate 2 sub-tasks of knowledge-based namely taxonomy and larger animals. Despite prominent variations in data sizes, EASE along with GPT-3.5-turbo-1106 pre-trained model demonstrates an excellent accuracy score of 100\% on both tasks. It can be seen from Table \ref{KCU} that performance of StablePRompt varied significantly across tasks. Even though InstructZero along with Vicuna model employed with smallest training data size, it still delivered comparable results, demonstrating its effectiveness in knowledge-based tasks. \\

Table \ref{KCU} provides a comprehensive overview of 8 different prompt optimization strategies that have been evaluated for contextual understanding task, on BBH dataset. It is evident from Table \ref{KCU} that EvoPrompt along with GPT-3.5 model exhibits higher accuracy score than Adv-ICL paired with GPT-3.5-turbo-0613 model. OPRO along with PaLM2-L pre-trained model demonstrates strong performance despite the reduced training and validation data. Specifically, for movie recommendation task, OPRO outperforms all other approaches with accuracy score above 90\%. However, performance of PE2 optimization approach varied significantly across tasks, which highlights its problem-specific flexibility.
\scriptsize
\begin{longtable}{|p{0.58cm}|p{2cm}|p{2cm}|p{2cm}|p{1.85cm}|p{1.5cm}|p{1.85cm}|}
\caption{Overview of performance evaluation of various Knowledge and Contextual Understanding approaches. "*" symbol following a dataset name signifies identical train/dev/test split was utilized for the reported results.}\label{KCU}\\
\hline
\textbf{Task} & \textbf{Sub-Task} & \textbf{Method} & \textbf{Data} & \textbf{Pre-trained Model} & \textbf{Evaluation Metric} & \textbf{Performance} \\ \hline
\rotatebox{90}{\begin{tabular}[c]{@{}l@{}}Knowledge-Based\\  Tasks\end{tabular}} & {\begin{tabular}[c]{@{}c@{}}Sports \\ Understanding\end{tabular}} & OPRO & BBH \cite{suzgun2022bigbenchhard} *** & \begin{tabular}[c]{@{}l@{}}PaLM 2-L\\  Scorer \\ and PaLM\\ 2-L-IT \\ Optimizer\end{tabular} & Accuracy & 88 \\ \cmidrule(lr){3-7} 
& & EvoPrompt & BBH \cite{suzgun2022bigbenchhard} * & GPT-3.5 & Accuracy & 96.8 \\ \cmidrule(lr){3-7} 
& & APE & BBH \cite{suzgun2022bigbenchhard} * & \begin{tabular}[c]{@{}l@{}}text-\\davinci-\\002\end{tabular}  & Normalized Score & 36 \\ \cmidrule(lr){3-7} 
& & MoP & BBH \cite{suzgun2022bigbenchhard} & \_ & \_ & \_ \\ \cmidrule(lr){3-7} 
& & StablePRompt & BBH \cite{suzgun2022bigbenchhard} * & Gemma-7B & Accuracy & 60.12 \\ \cmidrule(lr){3-7} 
& & PE2 & BBH \cite{suzgun2022bigbenchhard} **** & \begin{tabular}[c]{@{}l@{}}GPT-3.5\\-turbo\\-instruct \\ task model\\ and \\ GPT-4-\\turbo \\optimizer\end{tabular} & Accuracy & 64 \\ \cmidrule(lr){3-7} 
& & Adv-ICL & BBH \cite{suzgun2022bigbenchhard} * & \begin{tabular}[c]{@{}l@{}}GPT-3.5\\-turbo-\\0613\end{tabular} & Accuracy & less than 88 \\ \cmidrule(lr){2-7}
& {\begin{tabular}[c]{@{}c@{}}Penguins in \\ a Table\end{tabular}} & OPRO & BBH \cite{suzgun2022bigbenchhard} *** & \begin{tabular}[c]{@{}l@{}}PaLM\\ 2-L\\ Scorer\\ and \\ GPT-3.5-\\turbo \\Optimizer\end{tabular} & Accuracy & 72.6 \\ \cmidrule(lr){3-7} 
& & EvoPrompt & BBH \cite{suzgun2022bigbenchhard} * & GPT-3.5 & Accuracy & 84.25 \\ \cmidrule(lr){3-7} 
& & MoP & BBH \cite{suzgun2022bigbenchhard} & \_ & \_ & \_ \\ \cmidrule(lr){3-7} 
& & PROMPTAGENT & BBH \cite{suzgun2022bigbenchhard} & GPT-4 & Accuracy & 0.962 \\ \cmidrule(lr){3-7} 
& & PE2 & BBH \cite{suzgun2022bigbenchhard} **** & \begin{tabular}[c]{@{}l@{}}GPT-3.5\\-turbo\\-instruct \\ Task \\Model and \\ GPT-4-\\turbo \\Optimizer\end{tabular} & Accuracy & 73.91 \\ \cmidrule(lr){3-7} 
& & Adv-ICL & BBH \cite{suzgun2022bigbenchhard} * & \begin{tabular}[c]{@{}l@{}}GPT-3.5\\-turbo-\\0613\end{tabular}  & Accuracy &  76 \\ \cmidrule(lr){2-7}
& Taxonomy & {\begin{tabular}[c]{@{}c@{}}Prompt\\ BREEDER\end{tabular}}  & IIT \cite{honovich2022instruction} * & PaLM2-L & Execution Accuracy & 100 \\ \cmidrule(lr){3-7} 
& & APE & IIT \cite{honovich2022instruction} * & \begin{tabular}[c]{@{}l@{}}text-\\davinci-\\002\end{tabular} & Execution Accuracy & 79 (fewshot) \\ \cmidrule(lr){3-7} 
& & MoP & IIT \cite{honovich2022instruction} & \_ & \_ & \_ \\ \cmidrule(lr){3-7} 
& & PE2 & IIT \cite{honovich2022instruction} & \begin{tabular}[c]{@{}l@{}}text-davinci\\-003 as\\  Task \\Model and \\ GPT-4 as\\ Optimizer\end{tabular} & Mean & 89 \\ \cmidrule(lr){3-7} 
& & EASE & IIT \cite{honovich2022instruction} * & \begin{tabular}[c]{@{}l@{}}GPT-3.5\\-turbo-\\1106\end{tabular} & Accuracy & 100 \\ \cmidrule(lr){3-7} 
& & InstructZero & IIT \cite{honovich2022instruction} ** & Vicuna Model & Exact Set & 0.82 \\ \cmidrule(lr){3-7} 
& & StablePRompt & IIT \cite{honovich2022instruction} * & \begin{tabular}[c]{@{}l@{}}Gemma-7B, \\ Instruct\\GPT-3.5\end{tabular} & \begin{tabular}[c]{@{}l@{}}Exact Set, \\ Accuracy\end{tabular} & 0.5,  75 \\ \cmidrule(lr){2-7}
& {\begin{tabular}[c]{@{}c@{}}Larger \\ Animal\end{tabular}} & {\begin{tabular}[c]{@{}c@{}}Prompt\\ BREEDER\end{tabular}}  & IIT \cite{honovich2022instruction} ****** & PaLM2-L & Execution Accuracy & 97 \\ \cmidrule(lr){3-7} 
& & APE & IIT \cite{honovich2022instruction} ****** & \begin{tabular}[c]{@{}l@{}}text-\\davinci-\\002\end{tabular} & Execution Accuracy & 97 \\ \cmidrule(lr){3-7} 
& & MoP & IIT \cite{honovich2022instruction} & \_ & \_ & \_ \\ \cmidrule(lr){3-7} 
& & EASE & IIT \cite{honovich2022instruction} ****** & \begin{tabular}[c]{@{}l@{}}GPT-3.5\\-turbo-\\1106\end{tabular}  & Accuracy & 100 \\ \cmidrule(lr){3-7} 
& & InstructZero & IIT \cite{honovich2022instruction} ** & Vicuna Model & \begin{tabular}[c]{@{}l@{}}Exact \\Match \\ Accuracy\end{tabular} & 0.91 \\ \cmidrule(lr){3-7} 
& & StablePRompt & IIT \cite{honovich2022instruction} ****** & \begin{tabular}[c]{@{}l@{}}Gemma-7B, \\ Instruct\\GPT-3.5\end{tabular} & \begin{tabular}[c]{@{}l@{}}Exact \\Match, \\ Accuracy\end{tabular} & 0.937,93 \\ \cmidrule(lr){2-7}
& {\begin{tabular}[c]{@{}c@{}}Ruin \\ Names\end{tabular}} & OPRO & BBH \cite{suzgun2022bigbenchhard} *** & \begin{tabular}[c]{@{}l@{}}PaLM 2-L \\Scorer  and \\PaLM 2-L-IT \\ Optimizer\end{tabular} & Accuracy & 88 \\ \cmidrule(lr){3-7} 
& & EvoPrompt & BBH \cite{suzgun2022bigbenchhard} * & GPT-3.5 & Accuracy & 69.6 \\ \cmidrule(lr){3-7} 
& & APE & BBH  \cite{suzgun2022bigbenchhard} * & \begin{tabular}[c]{@{}l@{}}text-\\davinci-\\002\end{tabular} & Normalized Score & -14.7 \\ \cmidrule(lr){3-7} 
& & MoP & BBH \cite{suzgun2022bigbenchhard} & \_ & \_ & \_ \\ \cmidrule(lr){3-7} 
& & StablePRompt & BBH \cite{suzgun2022bigbenchhard} * & Gemma-7B & Accuracy & 37.08 \\ \cmidrule(lr){3-7} 
& & PE2 & BBH \cite{suzgun2022bigbenchhard} **** & \begin{tabular}[c]{@{}l@{}}GPT-3.5-\\turbo-instruct \\ Task Model \\and \\ GPT-4-turbo \\Optimizer\end{tabular} & Accuracy & 72 \\ \cmidrule(lr){3-7} 
& & Adv-ICL & BBH \cite{suzgun2022bigbenchhard} * & \begin{tabular}[c]{@{}l@{}}GPT-3.5\\-turbo-\\0613\end{tabular} & Accuracy &  57 \\ \hline
\rotatebox{90}{\begin{tabular}[c]{@{}l@{}}Contextual \\ Understanding \\ Tasks\end{tabular}}  & {\begin{tabular}[c]{@{}c@{}}Reasoning \\about\\ Colored\\ Objects \end{tabular}}  & OPRO & BBH \cite{suzgun2022bigbenchhard} ***** & \begin{tabular}[c]{@{}l@{}}PaLM 2-L\\ Scorer \\ and PaLM \\2-L-IT \\ Optimizer\end{tabular} & Accuracy & 85.5 \\ \cmidrule(lr){3-7} 
& & EvoPrompt & BBH \cite{suzgun2022bigbenchhard} * & GPT-3.5 & Accuracy & 88 \\ \cmidrule(lr){3-7} 
& & MoP & BBH \cite{suzgun2022bigbenchhard} & \_ & \_ & \_ \\ \cmidrule(lr){3-7} 
& & AEO & BBH \cite{suzgun2022bigbenchhard} & \begin{tabular}[c]{@{}l@{}}text-bison\\ task \\ model and \\ PaLM 2-L \\optimizer\end{tabular} & Accuracy & 83.5 \\ \cmidrule(lr){3-7} 
& & PE2 & BBH \cite{suzgun2022bigbenchhard} ****** & \begin{tabular}[c]{@{}l@{}}GPT-3.5-\\turbo-instruct\\ task model\\ and \\ GPT-4-turbo \\optimizer\end{tabular} & Accuracy & 58 \\ \cmidrule(lr){3-7} 
& & Adv-ICL & BBH \cite{suzgun2022bigbenchhard} * & \begin{tabular}[c]{@{}l@{}}GPT-3.5\\-turbo-\\0613\end{tabular} & \_ &  79 \\ \cmidrule(lr){2-7}
& {\begin{tabular}[c]{@{}c@{}}Movie\\ Recommendation \end{tabular}} & OPRO & BBH \cite{suzgun2022bigbenchhard} ***** & \begin{tabular}[c]{@{}l@{}}PaLM 2-L \\Scorer \\ and PaLM \\2-L-IT \\ Optimizer\end{tabular} & Accuracy & 90.5 \\ \cmidrule(lr){3-7} 
& & EvoPrompt & BBH \cite{suzgun2022bigbenchhard} * & GPT-3.5 & Accuracy & 86 \\ \cmidrule(lr){3-7} 
& & APE & BBH \cite{suzgun2022bigbenchhard} * & \begin{tabular}[c]{@{}l@{}}text-\\davinci-\\002\end{tabular} & Normalized Score & 12 \\ \cmidrule(lr){3-7} 
& & MoP & BBH \cite{suzgun2022bigbenchhard} & \_ & \_ & \_ \\ \cmidrule(lr){3-7} 
& & StablePRompt & BBH \cite{suzgun2022bigbenchhard} * & Gemma-7B & Accuracy & 55.3 \\ \cmidrule(lr){3-7} 
& & PE2 & BBH \cite{suzgun2022bigbenchhard} ****** & \begin{tabular}[c]{@{}l@{}}GPT-3.5-\\turbo-\\instruct \\ Task Model\\ and \\ GPT-4-turbo \\Optimizer\end{tabular} & Accuracy & 70 \\ \cmidrule(lr){3-7} 
& & Adv-ICL & BBH \cite{suzgun2022bigbenchhard} * & \begin{tabular}[c]{@{}l@{}}GPT-3.5\\-turbo-\\0613\end{tabular} & Accuracy &  73 \\ \hline
\end{longtable}

\normalsize
%%%%%%%%%%%%%%%%%%%%%%%%%%%%%%%%%%%%%%%%%%%%%%%%%%%%%%%%%%%%%%%%%%%%%%
%Trends
 
\subsubsection{Benchmark Datasets of knowledge and Contextual Understanding based Tasks}
\begin{figure}
\centering
\includegraphics[width=1\linewidth]{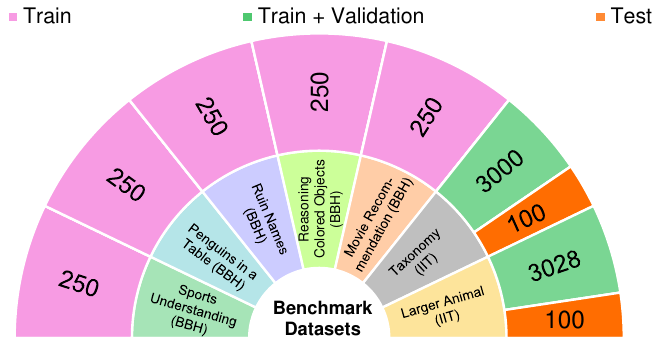}
\caption{Benchmark datasets of knowledge and contextual understanding based tasks}
\label{fig:5}
\end{figure}

Figure \ref{fig:5} illustrates the sample distribution across seven benchmarking datasets used for knowledge and contextual understanding based tasks. Among these datasets, Reasoning about Colored Objects (BBH), Movie Recommendation (BBH), and Penguins on a Table (BBH) are tasks based on multiple-choice questions (MCQs). Similar to the benchmark datasets used for other tasks, the two primary benchmarks for evaluating knowledge and contextual understanding across various prompt optimization strategies are IIT and BBH. The datasets have a considerable disproportion in dataset sizes, with the five BBH datasets each containing 250 training samples, while the two IIT datasets Taxonomy and Larger Animal have 3,000 and 3,028 samples for training and validation, respectively, with a consistent test set of 100 samples. This structured sample distribution ensures a standardized evaluation criterion to facilitate fair and reliable comparisons.

\subsection{Reasoning based Tasks}

Table \ref{reasoning} provides a comprehensive overview of 17 distinct prompt optimization strategies employed for various reasoning tasks, evaluated against 21 unique datasets. It is evident from Table \ref{reasoning} that BBH is one of the most widely used datasets for evaluating prompt optimization strategies, with model performance significantly influenced by the nature of the tasks. Moreover, approaches such as Adv-ICL, OPRO, and PE2 have been tested across a diverse range of tasks, which provides a more comprehensive assessment of a model’s generalization and robustness. In contrast, methods evaluated on a limited number of tasks may offer narrower insights into a model’s overall capabilities and performance.
\scriptsize
% \usepackage{multirow}
% \renewcommand{\arraystretch}{1}
\begin{longtable}{|p{0.65cm}|p{1.8cm}|p{1.58cm}|p{2cm}|p{1.5cm}|p{1.5cm}|p{1.5cm}|p{1.85cm}|}
\caption{Overview of existing reasoning task approaches. "*", "**" and "***" symbol following a dataset name signifies the identical train/dev/test split was utilized for the reported results.}\label{reasoning}\\
\hline
\textbf{Task} & \textbf{Sub-Task} & \textbf{Method} & \textbf{Data} & \textbf{Domain} & \textbf{Pre-trained Model} & \textbf{Evaluation Metric} & \textbf{Performance} \\ \hline
& {Commonsense reasoning} & \multirow{2}{*}{Auto-CoT} & CSQA \cite{talmor2018commonsenseqa} * & Multiple domains & \begin{tabular}[c]{@{}l@{}}text-\\davinci-\\002\end{tabular} & Accuracy & 74.4 \\ \cmidrule(lr){4-8} 
\rotatebox{90}{Commonsense reasoning} &  &  & StrategyQA \cite{geva2021did} & {\begin{tabular}[c]{@{}c@{}}Various\\ Topics\end{tabular}}  & \begin{tabular}[c]{@{}l@{}}text-\\davinci-\\002\end{tabular} & Accuracy & 65.4 \\ \cmidrule(lr){3-8} 
&  & \multirow{2}{*}{{\begin{tabular}[c]{@{}c@{}}Automate\\-CoT \end{tabular}}} & CSQA \cite{talmor2018commonsenseqa} * & {\begin{tabular}[c]{@{}c@{}}Multiple\\ Domains\end{tabular}}  & code-davinci-002 & Exact Match & 84 \\ \cmidrule(lr){4-8} 
&  &  & StrategyQA \cite{geva2021did} & {\begin{tabular}[c]{@{}c@{}}Various\\ Topics\end{tabular}}  & code-davinci-002 & Exact Match & 80.6 \\ \cmidrule(lr){3-8} 
&  & \multirow{2}{*}{{\begin{tabular}[c]{@{}c@{}}Prompt\\ BREEDER\end{tabular}} } & CSQA \cite{talmor2018commonsenseqa} & {\begin{tabular}[c]{@{}c@{}}Multiple\\ Domains\end{tabular}} & PaLM2-L & Accuracy & 85.9 (fewshot) \\ \cmidrule(lr){4-8} 
&  &  & StrategyQA \cite{geva2021did} & {\begin{tabular}[c]{@{}c@{}}Various\\ Topics\end{tabular}} & PaLM2-L & Accuracy & 80.2 (fewshot) \\ \hline
\rotatebox{90}{Symbolic Reasoning} & \multirow{3}{*}{{\begin{tabular}[c]{@{}c@{}}Symbolic\\ Reasoning\end{tabular}}} & \multirow{2}{*}{{\begin{tabular}[c]{@{}c@{}}Auto\\-CoT\end{tabular}}} & Last Letter Concatenation \cite{wei2022chain} * & \_ & \begin{tabular}[c]{@{}l@{}}text-\\davinci-\\002\end{tabular} & Accuracy & 59.7 \\ \cmidrule(lr){4-8} 
&  &  & Coin Flip \cite{wei2022chain} & \_ & text-davinci-002 & Accuracy & 99.9 \\ \cmidrule(lr){3-8} 
&  & Automate-CoT & Last Letter Concatenation \cite{wei2022chain} * & \_ & GPT-3.5-turbo & Exact Match & 76.2 \\ \cmidrule(lr){2-8} 
& \multirow{8}{*}{{\begin{tabular}[c]{@{}c@{}}Dyck\\ Languages\end{tabular}}} & OPRO & BBH \cite{suzgun2022bigbenchhard} ** & \_ & PaLM2-L Scorer and PaLM2-L-IT Optimizer & Accuracy & 100 \\ \cmidrule(lr){3-8} 
&  & {\begin{tabular}[c]{@{}c@{}}Evo\\Prompt \end{tabular}} & BBH \cite{suzgun2022bigbenchhard} * & \_ & GPT3.5 & Accuracy & 44.4 \\ \cmidrule(lr){3-8} 
&  & APE & BBH \cite{suzgun2022bigbenchhard} * & \_ & text-davinci-002 & Normalized Score & 18 \\ \cmidrule(lr){3-8} 
&  & MoP & BBH \cite{suzgun2022bigbenchhard} & \_ & \_ & \_ & \_ \\ \cmidrule(lr){3-8} 
&  & AEO & BBH \cite{suzgun2022bigbenchhard} **** & \_ & text-bison task model and PaLM 2-L Optimizer & Accuracy & 20.9 \\ \cmidrule(lr){3-8} 
&  & {\begin{tabular}[c]{@{}c@{}}Stable\\PRompt \end{tabular}} & BBH \cite{suzgun2022bigbenchhard} * & \_ & Gemma-7B & ExactMatch & 0 \\ \cmidrule(lr){3-8} 
&  & PE2 & BBH \cite{suzgun2022bigbenchhard} *** & \_ & GPT-3.5-turbo-instruct task model and  GPT-4-turbo optimizer & Accuracy & 8 \\ \cmidrule(lr){3-8} 
&  & Adv-ICL & BBH \cite{suzgun2022bigbenchhard} * & \_ & GPT-3.5-turbo-0613 & ExactMatch &  65 \\ \hline
\rotatebox{90}{Logical Reasoning} & {Boolean Expressions} & OPRO & BBH \cite{suzgun2022bigbenchhard} ** & \_ & PaLM2-L Scorer and the GPT-3.5-turbo optimizer & Accuracy & 89.5 \\ \cmidrule(lr){3-8} 
&  & {\begin{tabular}[c]{@{}c@{}}Evo\\Prompt\end{tabular}} & BBH \cite{suzgun2022bigbenchhard} * & \_ & GPT3.5 & Accuracy & 90.8 \\ \cmidrule(lr){3-8} 
&  & MoP & BBH \cite{suzgun2022bigbenchhard} & \_ & \_ & \_ & \_ \\ \cmidrule(lr){3-8} 
&  & PE2 & BBH \cite{suzgun2022bigbenchhard} *** & \_ & GPT-3.5-turbo-instruct task model and  GPT-4-turbo optimizer & Accuracy & 92 \\ \cmidrule(lr){3-8} 
&  & Adv-ICL & BBH \cite{suzgun2022bigbenchhard} * & \_ & GPT-3.5-turbo-0613 & Accurracy &  98 \\ \cmidrule(lr){2-8} 
& {{\begin{tabular}[c]{@{}c@{}}Formal\\ Fallacies \\Syllogisms\\ Negation\end{tabular}}} & OPRO & BBH \cite{suzgun2022bigbenchhard} ** & \_ & PaLM2-L Scorer and PaLM2-L-IT Optimizer & Accuracy & 64 \\ \cmidrule(lr){3-8} 
&  & {\begin{tabular}[c]{@{}c@{}}Evo\\Prompt\end{tabular}} & BBH \cite{suzgun2022bigbenchhard} * & \_ & GPT3.5 & Accuracy & 56 \\ \cmidrule(lr){3-8} 
&  & APE & BBH \cite{suzgun2022bigbenchhard} * & \_ & text-davinci-002 & Normalized Score & 12 \\ \cmidrule(lr){3-8} 
&  & MoP & BBH \cite{suzgun2022bigbenchhard} & \_ & \_ & \_ & \_ \\ \cmidrule(lr){3-8} 
&  & AEO & BBH \cite{suzgun2022bigbenchhard} **** & \_ & text-bison task model and PaLM 2-L optimizer & Accuracy & 73.1 \\ \cmidrule(lr){3-8} 
&  & {\begin{tabular}[c]{@{}c@{}}Stable\\PRompt\end{tabular}} & BBH \cite{suzgun2022bigbenchhard} * & \_ & Gemma-7B & Accuracy & 58.34 \\ \cmidrule(lr){3-8} 
&  & PE2 & BBH \cite{suzgun2022bigbenchhard} *** & \_ & GPT-3.5-turbo-instruct task model and  GPT-4-turbo optimizer & Accuracy & 52 \\ \cmidrule(lr){3-8} 
&  & Adv-ICL & BBH \cite{suzgun2022bigbenchhard}* & \_ & GPT-3.5-turbo-0613 & Accuracy & less than 54 \\ \cmidrule(lr){2-8} 
& {Logical Deduction} & OPRO & \begin{tabular}[c]{@{}l@{}}logical\_deduction\\ \_seven\_objects\end{tabular} & \_ & text-bison scorer and GPT-3.5-turbo optimizer & Accuracy & 57.5 \\ \cmidrule(lr){3-8} 
&  & {\begin{tabular}[c]{@{}c@{}}Evo\\Prompt\end{tabular}} & \begin{tabular}[c]{@{}l@{}}logical\_deduction\\ \_three\_objects *,\\ logical\_ deduction\_\\five\ \_objects * , \\ logical\_deduction\_\\ seven\_objects\end{tabular} & \_ & GPT-3.5 & Accuracy & 94.40, 65.20,  54.40 \\ \cmidrule(lr){3-8} 
&  & MoP & BBH \cite{suzgun2022bigbenchhard} * & \_ & \_ & \_ & \_ \\ \cmidrule(lr){3-8} 
&  & AEO & \begin{tabular}[c]{@{}l@{}}logical\_deduction\_\\ five\_objects *\end{tabular} * & \_ & text-bison task model and PaLM 2-L Optimizer & Accuracy & 54.7 \\ \cmidrule(lr){3-8} 
&  & PE2 & \begin{tabular}[c]{@{}l@{}}logical\_deduction\_\\ three\_objects *,  \\ logical\_deduction\_\\ five\_objects *,  \\ logical\_deduction\\ \_seven\_objects\end{tabular} & \_ & GPT-3.5-turbo-instruct task model and  GPT-4-turbo optimizer & Accuracy & 66, 60, 42 \\ \cmidrule(lr){3-8} 
&  & Adv-ICL & BBH \cite{suzgun2022bigbenchhard} * & \_ & GPT-3.5-turbo-0613 & Accuracy & 67 \\ \cmidrule(lr){2-8} 
& \begin{tabular}[c]{@{}l@{}}Tracking \\Shuffled\\ Objects\end{tabular} & OPRO & \begin{tabular}[c]{@{}l@{}}Tracking\_shuffled\\ \_objects\_seven\_objects\end{tabular} & \_ & PaLM2-L Scorer and the GPT-3.5-turbo optimizer & Accuracy & 63.5 \\ \cmidrule(lr){3-8} 
&  & {\begin{tabular}[c]{@{}c@{}}Evo\\Prompt\end{tabular}} & \begin{tabular}[c]{@{}l@{}}Tracking\_shuffled\_\\ objects\_three\_objects *,  \\ Tracking\_shuffled\_\\ objects\_five\_objects *, \\ Tracking\_shuffled\_\\ objects\_seven\_objects\end{tabular} & \_ & GPT-3.5 & Accuracy & 69.20, 81.20,  84.80 \\ \cmidrule(lr){3-8} 
&  & MoP & BBH \cite{suzgun2022bigbenchhard} * & \_ & \_ & \_ & \_ \\ \cmidrule(lr){3-8} 
&  & PE2 & \begin{tabular}[c]{@{}l@{}}Tracking\_shuffled\_\\ objects\_three\_objects *,  \\ Tracking\_shuffled\_\\ objects\_five\_objects *, \\ Tracking\_shuffled\_\\ objects\_seven\_objects\end{tabular} & \_ & GPT-3.5-turbo-instruct task model and  GPT-4-turbo optimizer & Accuracy & 64, 64, 66 \\ \cmidrule(lr){3-8} 
&  & Adv-ICL & BBH \cite{suzgun2022bigbenchhard} * & \_ & GPT-3.5-turbo-0613 & Accuracy &  62 \\ \cmidrule(lr){2-8} 
& {Web\_of\_lies} & OPRO & BBH \cite{suzgun2022bigbenchhard} ** & \_ & PaLM2-L Scorer and the GPT-3.5-turbo optimizer & Accuracy & 91 \\ \cmidrule(lr){3-8} 
&  & PE2 & BBH \cite{suzgun2022bigbenchhard} *** & \_ & GPT-3.5-turbo-instruct task model and  GPT-4-turbo optimizer & Accuracy & 46 \\ \cmidrule(lr){3-8} 
&  & Adv-ICL & BBH \cite{suzgun2022bigbenchhard} & \_ & GPT-3.5-turbo-0613 & Accuracy & 100 \\ \hline
\rotatebox{90}{Temporal Reasoning} & {{\begin{tabular}[c]{@{}c@{}}Date\\ Understanding\end{tabular}}} & OPRO & BBH \cite{suzgun2022bigbenchhard} ** & \_ & PaLM2-L Scorer and the PaLM2-L-IT Optimize & Accuracy & 84.5 \\ \cmidrule(lr){3-8} 
&  & {\begin{tabular}[c]{@{}c@{}}Evo\\Prompt\end{tabular}} & BBH \cite{suzgun2022bigbenchhard} * & \_ & GPT-3.5 & Accuracy & 85.6 \\ \cmidrule(lr){3-8} 
&  & MoP & BBH \cite{suzgun2022bigbenchhard} & \_ & \_ & \_ & \_ \\ \cmidrule(lr){3-8} 
&  & PE2 & BBH \cite{suzgun2022bigbenchhard} *** & \_ & GPT-3.5-turbo-instruct task model and  GPT-4-turbo optimizer & Accuracy & 76 \\ \cmidrule(lr){3-8} 
&  & Adv-ICL & BBH \cite{suzgun2022bigbenchhard} * & \_ & GPT-3.5-turbo-0613 & Accuracy &  81 \\ \cmidrule(lr){2-8} 
& \multirow{6}{*}{{\begin{tabular}[c]{@{}c@{}}Temporal\\ Sequences\end{tabular}}} & OPRO & BBH \cite{suzgun2022bigbenchhard} ** & \_ & PaLM2-L Scorer and PaLM2-L-IT Optimizer & Accuracy & 100 \\ \cmidrule(lr){3-8} 
&  & {\begin{tabular}[c]{@{}c@{}}Evo\\Prompt\end{tabular}} & BBH \cite{suzgun2022bigbenchhard} * & \_ & GPT-3.5 & Accuracy & 78.8 \\ \cmidrule(lr){3-8} 
&  & MoP & BBH \cite{suzgun2022bigbenchhard} & \_ & \_ & \_ & \_ \\ \cmidrule(lr){3-8} 
&  & {\begin{tabular}[c]{@{}c@{}}Prompt\\AGENT \end{tabular}} & BBH \cite{suzgun2022bigbenchhard} ***** & \_ & GPT-4 & Accuracy & 0.982 \\ \cmidrule(lr){3-8} 
&  & PE2 & BBH \cite{suzgun2022bigbenchhard} *** & \_ & GPT-3.5-turbo-instruct task model and  GPT-4-turbo optimizer & Accuracy & 82 \\ \cmidrule(lr){3-8} 
&  & Adv-ICL & BBH \cite{suzgun2022bigbenchhard} * & \_ & GPT-3.5-turbo-0613 & Accuracy &  89 \\ \hline
& {Geometric Shapes} & OPRO & BBH \cite{suzgun2022bigbenchhard} ** & \_ & PaLM2-L Scorer and PaLM2-L-IT Optimizer & Accuracy & 57 \\ \cmidrule(lr){3-8} 
&  & {\begin{tabular}[c]{@{}c@{}}Evo\\Prompt\end{tabular}} & BBH \cite{suzgun2022bigbenchhard} * & \_ & GPT-3.5 & Accuracy & 64 \\ \cmidrule(lr){3-8} 
\rotatebox{90}{\begin{tabular}[c]{@{}c@{}}Spatial and \\Geometric Reasoning \end{tabular}} &  & MoP & BBH \cite{suzgun2022bigbenchhard} & \_ & \_ & \_ & \_ \\ \cmidrule(lr){3-8} 
&  & {\begin{tabular}[c]{@{}c@{}}Prompt\\AGENT \end{tabular}} & BBH \cite{suzgun2022bigbenchhard} & \_ & GPT-4 & Accuracy & 0.68 \\ \cmidrule(lr){3-8} 
&  & PE2 & BBH \cite{suzgun2022bigbenchhard} *** & \_ & GPT-3.5-turbo-instruct task model and  GPT-4-turbo optimizer & Accuracy & 82 \\ \cmidrule(lr){3-8} 
&  & Adv-ICL & BBH \cite{suzgun2022bigbenchhard} * & \_ & GPT-3.5-turbo-0613 & Accuracy & 48 \\ \cmidrule(lr){2-8} 
& \multirow{8}{*}{Navigate} & OPRO & BBH \cite{suzgun2022bigbenchhard} ** & \_ & PaLM2-L Scorer and PaLM2-L-IT Optimizer & Accuracy & 75 \\ \cmidrule(lr){3-8} 
&  & {\begin{tabular}[c]{@{}c@{}}Evo\\Prompt\end{tabular}} & BBH \cite{suzgun2022bigbenchhard} * & \_ & GPT-3.5 & Accuracy & 94.2 \\ \cmidrule(lr){3-8} 
&  & APE & BBH \cite{suzgun2022bigbenchhard} * & \_ & text-davinci-002 & Normalized Score & 12 \\ \cmidrule(lr){3-8} 
&  & MoP & BBH \cite{suzgun2022bigbenchhard} & \_ & \_ & \_ & \_ \\ \cmidrule(lr){3-8} 
&  & ABO & BBH \cite{suzgun2022bigbenchhard} & \_ & GPT-3.5-Turbo & Accuracy & 0.985 \\ \cmidrule(lr){3-8} 
&  & {\begin{tabular}[c]{@{}c@{}}Stable\\PRompt\end{tabular}} & BBH \cite{suzgun2022bigbenchhard} * & \_ & Gemma-7B & Accuracy & 53.3 \\ \cmidrule(lr){3-8} 
&  & PE2 & BBH \cite{suzgun2022bigbenchhard} *** & \_ & GPT-3.5-turbo-instruct task model and  GPT-4-turbo optimizer & Accuracy & 84 \\ \cmidrule(lr){3-8} 
&  & Adv-ICL & BBH \cite{suzgun2022bigbenchhard} * & \_ & GPT-3.5-turbo-0613 & Accuracy &  95 \\ \hline
& {Sum} & {\begin{tabular}[c]{@{}c@{}}Prompt\\ BREEDER\end{tabular}}  & IIT \cite{honovich2022instruction} * & \_ & PaLM2-L & Execution Accuracy & 100 \\ \cmidrule(lr){3-8} 
&  & APE & IIT \cite{honovich2022instruction} * & \_ & text-davinci-002 & Execution Accuracy & 100 \\ \cmidrule(lr){3-8} 
&  & MoP & IIT \cite{honovich2022instruction} & \_ & \_ & \_ & \_ \\ \cmidrule(lr){3-8} 
\rotatebox{90}{\begin{tabular}[c]{@{}c@{}}Numerical \\ and Arithmetic \\ Reasoning\end{tabular}} &  & EASE & IIT \cite{honovich2022instruction} * & \_ & \begin{tabular}[c]{@{}l@{}}GPT-3.5\\-turbo-\\1106\end{tabular} & Accuracy & 100 \\ \cmidrule(lr){3-8} 
&  & {\begin{tabular}[c]{@{}c@{}}Instruct\\Zero\end{tabular}} & IIT \cite{honovich2022instruction} & \_ & Vicuna Model & Exact Match & 1 \\ \cmidrule(lr){3-8} 
&  & {\begin{tabular}[c]{@{}c@{}}Stable\\PRompt\end{tabular}} & IIT \cite{honovich2022instruction} * & \_ & Gemma-7B & Exact Match & 1 \\ \cmidrule(lr){2-8} 
& {Difference} & {\begin{tabular}[c]{@{}c@{}}Prompt\\ BREEDER\end{tabular}}  & IIT \cite{honovich2022instruction} * & \_ & PaLM2-L & Execution Accuracy & 100 \\ \cmidrule(lr){3-8} 
&  & APE & IIT \cite{honovich2022instruction} * & \_ & text-davinci-002 & Execution Accuracy & 100 \\ \cmidrule(lr){3-8} 
&  & MoP & IIT \cite{honovich2022instruction} & \_ & \_ & \_ & \_ \\ \cmidrule(lr){3-8} 
&  & EASE & IIT \cite{honovich2022instruction} ** & \_ & \begin{tabular}[c]{@{}l@{}}GPT-3.5\\-turbo-\\1106\end{tabular}& Accuracy & 100 \\ \cmidrule(lr){3-8} 
&  & {\begin{tabular}[c]{@{}c@{}}Instruct\\Zero\end{tabular}} & IIT \cite{honovich2022instruction} & \_ & Vicuna Model & Exact Match & 1 \\ \cmidrule(lr){3-8} 
&  & {\begin{tabular}[c]{@{}c@{}}Stable\\PRompt\end{tabular}} & IIT \cite{honovich2022instruction} ** & \_ & Gemma-7B & Exact Match & 1 \\ \cmidrule(lr){2-8} 
& {{\begin{tabular}[c]{@{}c@{}}Number\\ to Word\end{tabular}}} & {\begin{tabular}[c]{@{}c@{}}Prompt\\ BREEDER\end{tabular}}  & IIT \cite{honovich2022instruction} * & \_ & PaLM2-L & Execution Accuracy & 100 \\ \cmidrule(lr){3-8} 
&  & APE & IIT \cite{honovich2022instruction} * & \_ & text-davinci-002 & Execution Accuracy & 100 \\ \cmidrule(lr){3-8} 
&  & MoP & IIT \cite{honovich2022instruction} & \_ & \_ & \_ & \_ \\ \cmidrule(lr){3-8} 
&  & EASE & IIT \cite{honovich2022instruction} * & \_ & \begin{tabular}[c]{@{}l@{}}GPT-3.5\\-turbo-\\1106\end{tabular} & Accuracy & 100 \\ \cmidrule(lr){3-8} 
&  & {\begin{tabular}[c]{@{}c@{}}Instruct\\Zero\end{tabular}} & IIT \cite{honovich2022instruction} & \_ & Vicuna Model & Exact Match & 1 \\ \cmidrule(lr){3-8} 
&  & {\begin{tabular}[c]{@{}c@{}}Stable\\PRompt\end{tabular}} & IIT \cite{honovich2022instruction} * & \_ & Gemma-7B & Exact Match & 1 \\ \cmidrule(lr){2-8} 
& {Multi-Step Arithmetic} & OPRO & BBH \cite{suzgun2022bigbenchhard} ** & \_ & PaLM2-L Scorer and PaLM2-L-IT Optimizer & Accuracy & 55.5 \\ \cmidrule(lr){3-8} 
&  & {\begin{tabular}[c]{@{}c@{}}Evo\\Prompt\end{tabular}} & BBH \cite{suzgun2022bigbenchhard} * & \_ & GPT3.5 & Accuracy & 51.6 \\ \cmidrule(lr){3-8} 
&  & MoP & BBH \cite{suzgun2022bigbenchhard} & \_ & \_ & \_ & \_ \\ \cmidrule(lr){3-8} 
&  & PE2 & BBH \cite{suzgun2022bigbenchhard} *** & \_ & GPT-3.5-turbo-instruct task model and  GPT-4-turbo optimizer & Accuracy & 78 \\ \cmidrule(lr){3-8} 
&  & Adv-ICL & BBH \cite{suzgun2022bigbenchhard} * & \_ & GPT-3.5-turbo-0613 & Accuracy & 69 \\ \cmidrule(lr){2-8} 
& {Object Counting} & OPRO & BBH \cite{suzgun2022bigbenchhard} ** & \_ & PaLM2-L scorer and GPT-3.5-turboo ptimizer & Accuracy & 92.5 \\ \cmidrule(lr){3-8} 
&  & {\begin{tabular}[c]{@{}c@{}}Evo\\Prompt\end{tabular}} & BBH \cite{suzgun2022bigbenchhard} * & \_ & GPT3.5 & Accuracy & 87.6 \\ \cmidrule(lr){3-8} 
&  & APE & BBH \cite{suzgun2022bigbenchhard} * & \_ & text-davinci-002 & Normalized Score & 44 \\ \cmidrule(lr){3-8} 
&  & MoP & BBH \cite{suzgun2022bigbenchhard} & \_ & \_ & \_ & \_ \\ \cmidrule(lr){3-8} 
&  & {\begin{tabular}[c]{@{}c@{}}Prompt\\AGENT\end{tabular}}  & BBH \cite{suzgun2022bigbenchhard} ***** & \_ & GPT-4 & Accuracy & 0.88 \\ \cmidrule(lr){3-8} 
&  & ABO & BBH \cite{suzgun2022bigbenchhard} & \_ & GPT-3.5-Turbo & Accuracy & 0.975 \\ \cmidrule(lr){3-8} 
&  & {\begin{tabular}[c]{@{}c@{}}Stable\\PRompt\end{tabular}} & BBH \cite{suzgun2022bigbenchhard} * & \_ & Gemma-7B & ExactMatch & 15.71 \\ \cmidrule(lr){3-8} 
&  & PE2 & BBH \cite{suzgun2022bigbenchhard} *** & \_ & GPT-3.5-turbo-instruct task model and  GPT-4-turbo optimizer & Accuracy & 74 \\ \cmidrule(lr){3-8} 
&  & Adv-ICL & BBH \cite{suzgun2022bigbenchhard} * & \_ & GPT-3.5-turbo-0613 & ExactMatch &  83 \\ \cmidrule(lr){2-8} 
& {{\begin{tabular}[c]{@{}c@{}}Math\\ Reasoning\end{tabular}}} & \multirow{3}{*}{{\begin{tabular}[c]{@{}c@{}}Prompt\\-OIRL\end{tabular}}} & GSM8K \cite{cobbe2021training} * & {\begin{tabular}[c]{@{}c@{}}Math\\ Word\\ Problem\end{tabular}} & GPT3.5-turbo & Accuracy & 0.67 \\ \cmidrule(lr){4-8} 
&  &  & MultiArith \cite{roy2016solving} & {\begin{tabular}[c]{@{}c@{}}Arithmetic\\ Word\\ Problem\end{tabular}} & GPT3.5-turbo & Accuracy & 0.841 \\ \cmidrule(lr){4-8} 
&  &  & SVAMP \cite{patel2021nlp} &  Arithmetic Problems & GPT3.5-turbo & Accuracy & 0.736 \\ \cmidrule(lr){3-8} 
&  & {OPRO} & GSM8K \cite{cobbe2021training} * & {\begin{tabular}[c]{@{}c@{}}Math\\ Word\\ Problem\end{tabular}} & PaLM2-L Scorer and PaLM2-L-IT Optimizer & Accuracy & 80.2 \\ \cmidrule(lr){4-8} 
&  &  & MultiArith \cite{roy2016solving} * & {\begin{tabular}[c]{@{}c@{}}Arithmetic\\ Word\\ Problem\end{tabular}} & PaLM 2-L-IT & Accuracy & 96.8 \\ \cmidrule(lr){4-8} 
&  &  & AQuA-RAT \cite{ling2017program} &  {\begin{tabular}[c]{@{}c@{}}Algebric\\ Word\\ Problem\\and \\multi-step\\problems\end{tabular}} & PaLM 2-L-IT & Accuracy & 54.3 \\ \cmidrule(lr){3-8} 
&  &{\begin{tabular}[c]{@{}c@{}}Prompt\\ BREEDER\end{tabular}}  & GSM8K \cite{cobbe2021training} * & {\begin{tabular}[c]{@{}c@{}}Math\\ Word\\ Problem\end{tabular}} & PaLM 2-L & Accuracy & 83.9 \\ \cmidrule(lr){4-8} 
&  &  & SVAMP \cite{patel2021nlp} * & {\begin{tabular}[c]{@{}c@{}}Math\\  Problem\end{tabular}} & PaLM 2-L & Accuracy & 93.7 \\ \cmidrule(lr){4-8} 
&  &  & MultiArith \cite{roy2016solving} * & Arithmetic Word Problem & PaLM 2-L & Accuracy & 100 \\ \cmidrule(lr){4-8} 
&  &  & AddSub \cite{hosseini2014learning}* & {\begin{tabular}[c]{@{}c@{}}Math\\ Word\\ Problem\end{tabular}} & PaLM 2-L & Accuracy & 87.8 \\ \cmidrule(lr){4-8} 
&  &  & AQuA-RAT \cite{ling2017program} * & {\begin{tabular}[c]{@{}c@{}}Algebric\\ Word\\ Problem\\and \\multi-step\\problems\end{tabular}}  & PaLM 2-L & Accuracy & 64.6 \\ \cmidrule(lr){4-8} 
&  &  & SingleEq \cite{koncel2015parsing} * & Single-Equation Problems & PaLM 2-L & Accuracy & 98.9 \\ \cmidrule(lr){3-8} 
&  & {APE} & GSM8K \cite{cobbe2021training} * & {\begin{tabular}[c]{@{}c@{}}Math\\ Word\\ Problem\end{tabular}} & text-davinci-002 & Accuracy &  40 \\ \cmidrule(lr){4-8} 
&  &  & MultiArith \cite{roy2016solving} * & Arithmetic Word Problem & text-davinci-002 & Accuracy &  79 \\ \cmidrule(lr){4-8} 
&  &  & SVAMP \cite{patel2021nlp} * & Arithmetic Problems & text-davinci-002 & Accuracy &  74 \\ \cmidrule(lr){4-8} 
&  &  & AQuA-RAT \cite{ling2017program} * & {\begin{tabular}[c]{@{}c@{}}Algebric\\ Word\\ Problem\\and \\multi-step\\problems\end{tabular}}  & text-davinci-002 & Accuracy &  35 \\ \cmidrule(lr){4-8} 
&  &  & AddSub \cite{hosseini2014learning} * & {\begin{tabular}[c]{@{}c@{}}Math\\ Word\\ Problem\end{tabular}} & text-davinci-002 & Accuracy &  74 \\ \cmidrule(lr){4-8} 
&  &  & SingleEq \cite{koncel2015parsing} * & Single-Equation Problems & text-davinci-002 & Accuracy &  77 \\ \cmidrule(lr){3-8} 
&  & PRewriter & GSM8K \cite{cobbe2021training} & {\begin{tabular}[c]{@{}c@{}}Math\\ Word\\ Problem\end{tabular}} & PaLM2-L & Accuracy & \begin{tabular}[c]{@{}l@{}}83.8\\  (PRewrite-S)\end{tabular} \\ \cmidrule(lr){3-8} 
&  & \multirow{2}{*}{PE2} & MultiArith \cite{roy2016solving} * & Arithmetic Word  Problem & text-davinci-003 & Accuracy & 0.743 \\ \cmidrule(lr){4-8} 
&  &  & GSM8K \cite{cobbe2021training} & {\begin{tabular}[c]{@{}c@{}}Math\\ Word\\ Problem\end{tabular}} & text-davinci-004 & Accuracy & 0.505 \\ \cmidrule(lr){3-8} 
&  & \multirow{2}{*}{Adv-ICL} & GSM8K \cite{cobbe2021training} & {\begin{tabular}[c]{@{}c@{}}Math\\ Word\\ Problem\end{tabular}} & ChatGPT & Accuracy & 82.3 \\ \cmidrule(lr){4-8} 
&  &  & SVAMP \cite{patel2021nlp} & Arithmetic  Problems & ChatGPT & Accuracy & 81.1 \\ \cmidrule(lr){3-8} 
&  & \multirow{3}{*}{EASE} & GSM8K \cite{cobbe2021training} * & {\begin{tabular}[c]{@{}c@{}}Math\\ Word\\ Problem\end{tabular}} & \begin{tabular}[c]{@{}l@{}}GPT-3.5\\-turbo-\\1106\end{tabular} & Accuracy & 75 \\ \cmidrule(lr){4-8} 
&  &  & AQuA-RAT \cite{ling2017program} * & {\begin{tabular}[c]{@{}c@{}}Algebric\\ Word\\ Problem\\and \\multi-step\\problems\end{tabular}}  & \begin{tabular}[c]{@{}l@{}}GPT-3.5\\-turbo-\\1106\end{tabular} & Accuracy & 48 \\ \cmidrule(lr){4-8} 
&  &  & MATH \cite{hendrycks2021measuring} & {\begin{tabular}[c]{@{}c@{}}Math\\ Word\\ Problem\end{tabular}} & \begin{tabular}[c]{@{}l@{}}GPT-3.5\\-turbo-\\1106\end{tabular} & Accuracy & 65 \\ \cmidrule(lr){3-8} 
&  & \multirow{6}{*}{Auto-CoT} & MultiArith \cite{roy2016solving} * & Arithmetic Word Problem & text-davinci-002 & Accuracy & 92 \\ \cmidrule(lr){4-8} 
&  &  & GSM8K \cite{cobbe2021training} & {\begin{tabular}[c]{@{}c@{}}Math\\ Word\\ Problem\end{tabular}} & text-davinci-002 & Accuracy & 47.9 \\ \cmidrule(lr){4-8} 
&  &  & AddSub \cite{hosseini2014learning} * & {\begin{tabular}[c]{@{}c@{}}Math\\ Word\\ Problem\end{tabular}} & text-davinci-002 & Accuracy & 84.8 \\ \cmidrule(lr){4-8} 
&  &  & AQuA-RAT \cite{ling2017program} ** & {\begin{tabular}[c]{@{}c@{}}Algebric\\ Word\\ Problem\\and \\multi-step\\problems\end{tabular}}  & text-davinci-002 & Accuracy & 36.5 \\ \cmidrule(lr){4-8} 
&  &  & SingleEq \cite{koncel2015parsing} * & Single-Equation Problems & text-davinci-002 & Accuracy & 87 \\ \cmidrule(lr){4-8} 
&  &  & SVAMP \cite{patel2021nlp} * &  Arithmetic  Problems & text-davinci-002 & Accuracy & 69.5 \\ \cmidrule(lr){3-8} 
&  & \multirow{5}{*}{{\begin{tabular}[c]{@{}c@{}}Automate\\-CoT\end{tabular}}} & GSM8K \cite{cobbe2021training} & {\begin{tabular}[c]{@{}c@{}}Math\\ Word\\ Problem\end{tabular}} & code-davinci-002 & Exact Match & 82.4 \\ \cmidrule(lr){4-8} 
&  &  & SVAMP \cite{patel2021nlp} * &  Arithmetic  Problems & code-davinci-002 & Exact Match & 87.8 \\ \cmidrule(lr){4-8} 
&  &  & AQuA-RAT \cite{ling2017program} ** & {\begin{tabular}[c]{@{}c@{}}Algebric\\ Word\\ Problem\\and \\multi-step\\problems\end{tabular}}  & code-davinci-002 & Exact Match & 55.6 \\ \cmidrule(lr){4-8} 
&  &  & ASDiv & \_ & GPT-3.5-turbo & Exact Match & 91.5 \\ \cmidrule(lr){4-8} 
&  &  & SingleOp & Single-Equation Problems & code-davinci-002 & Exact Match & 94 \\ \hline
& {{\begin{tabular}[c]{@{}c@{}}Casual\\ Judgement\end{tabular}}} & OPRO & BBH \cite{suzgun2022bigbenchhard} ** & \_ & PaLM2-L scores and GPT-3.5-turbo optimizer & Accuracy & 58.7 \\ \cmidrule(lr){3-8} 
&  & {\begin{tabular}[c]{@{}c@{}}Evo\\Prompt\end{tabular}}& BBH \cite{suzgun2022bigbenchhard} * & \_ & GPT-3.5 & Accuracy & 65.78 \\ \cmidrule(lr){3-8} 
&  & APE & BBH \cite{suzgun2022bigbenchhard} * & \_ & text-davinci-002 & Normalized Score & 18 \\ \cmidrule(lr){3-8} 
\rotatebox{90}{\begin{tabular}[c]{@{}c@{}}Causal and \\ Counterfactual\\ Reasoning \end{tabular}} &  & MoP & BBH \cite{suzgun2022bigbenchhard} & \_ & \_ & \_ & \_ \\ \cmidrule(lr){3-8} 
&  & AEO & BBH \cite{suzgun2022bigbenchhard} **** & \_ & text-bison task model and PaLM 2-L optimizer & Accuracy & 63.7 \\ \cmidrule(lr){3-8} 
&  & {\begin{tabular}[c]{@{}c@{}}Prompt\\AGENT\end{tabular}} & BBH \cite{suzgun2022bigbenchhard} & \_ & GPT-4 & Accuracy & 0.77 \\ \cmidrule(lr){3-8} 
&  & {\begin{tabular}[c]{@{}c@{}}Stable\\PRompt\end{tabular}} & BBH \cite{suzgun2022bigbenchhard} * & \_ & Gemma-7B & Accuracy & 58.75 \\ \cmidrule(lr){3-8} 
&  & PE2 & BBH \cite{suzgun2022bigbenchhard} *** & \_ & GPT-3.5-turbo-instruct task model and  GPT-4-turbo optimizer & Accuracy & 56.32 \\ \cmidrule(lr){3-8} 
&  & Adv-ICL & BBH \cite{suzgun2022bigbenchhard} * & \_ & GPT-3.5-turbo-0613 & Accuracy &  62 \\ \cmidrule(lr){2-8} 
& {{\begin{tabular}[c]{@{}c@{}}Cause\\ Selection\end{tabular}}} & {\begin{tabular}[c]{@{}c@{}}Prompt\\ BREEDER\end{tabular}}  & IIT \cite{honovich2022instruction} * & \_ & PaLM2-L & Execution Accuracy & 100 \\ \cmidrule(lr){3-8} 
&  & APE & IIT \cite{honovich2022instruction} * & \_ & text-davinci-002 & Execution Accuracy & 100 (fewshot) \\ \cmidrule(lr){3-8} 
&  & MoP & IIT \cite{honovich2022instruction} * & \_ & \_ & \_ & \_ \\ \cmidrule(lr){3-8} 
&  & {\begin{tabular}[c]{@{}c@{}}Instruct\\Zero\end{tabular}} & IIT \cite{honovich2022instruction} * & \_ & Vicuna Model & F1-Score & 0.86 \\ \cmidrule(lr){3-8} 
&  & {\begin{tabular}[c]{@{}c@{}}Stable\\PRompt\end{tabular}} & IIT \cite{honovich2022instruction} * & \_ & Gemma-7B, InstructGPT3.5 & ExactMatch, Accuracy & 0.7, 92 \\ \hline
% \end{tabular}
\end{longtable}

\normalsize

Table \ref{reasoning} illustrates a high-level overview of 6 distinct optimization approaches and 7 unique datasets for tasks including logical deduction and tracking shuffle objects. However, inconsistencies in evaluation approaches and dataset splitting hinder direct comparisons. Moreover, for math reasoning task, 10 distinct approaches and 9 unique datasets are utilized. Table \ref{reasoning} shows that PROMPTBREEDER combined with PaLM 2-L pre-trained model achieves excellent results by scoring above 90\% accuracy for 3 distinct datasets namely MultiArith, SingleEq and GSM8K. However, PROMPTBREEDER shows significantly low performance of around 65\% on AuQA-RAT dataset. This suggests that the AQuA-RAT dataset presents a greater challenge, potentially due to differences in data characteristics or task complexity. While Adv-ICL demonstrates consistent, though not top-tier, performance on dataset including GSM8K and SVAMP. Moreover APE along with text-davinci-002 and EASE paired with GPT-3.5-turbo-1106 model consistently underperform. While other approaches (Prompt-OIRL, PRewriter, PE2, EASE, Auto-CoT and Automate-CoT) showed varied performance across various tasks,which highlights its task-specific adaptability.

Over symbolic reasoning task, Auto-CoT along with text-davinci-002 model exhibits near-perfect accuracy score of 99.9\%. While, for the last letter concatenation dataset, Auto-CoT performs significantly worse despite utilizing the same base model which highlights the importance of specific data characteristics. On the other hand, for data understanding task, almost all approaches exhibit significant performance above 75\%. While tasks including casual judgement and multi-step arithmetic remain challenging across most of the methods. Moreover, 3 distinct optimization approaches including Auto-CoT, Automate-CoT and PROMPTBREEDER are evaluated against CSQA and StrategyQA datasets for commonsense reasoning. However, despite using the same dataset direct performance comparisons of these prompt optimization strategies are difficult due to varying dataset sizes used for experimentation. Furthermore, the use of different evaluation metrics (Accuracy, Exact Match) makes it difficult to compare the effectiveness of these methods directly. Furthermore, OPRO optimization approach is evaluated using two different LLMs 1) optimizer model 2) scoring model. Optimizer model is responsible for refining prompt inputs, while scoring model quantifies output quality by assigning confidence scores. OPRO employed PaLM2-L model as scoring model while for optimizer model OPRO utilizes PaLM2-L IT and GPT-3.5-turbo. However, their impact on performance was minimal.

 \subsubsection{Benchmark Dataset of Reasoning Tasks}

\begin{figure}
\centering
\includegraphics[width=1\linewidth]{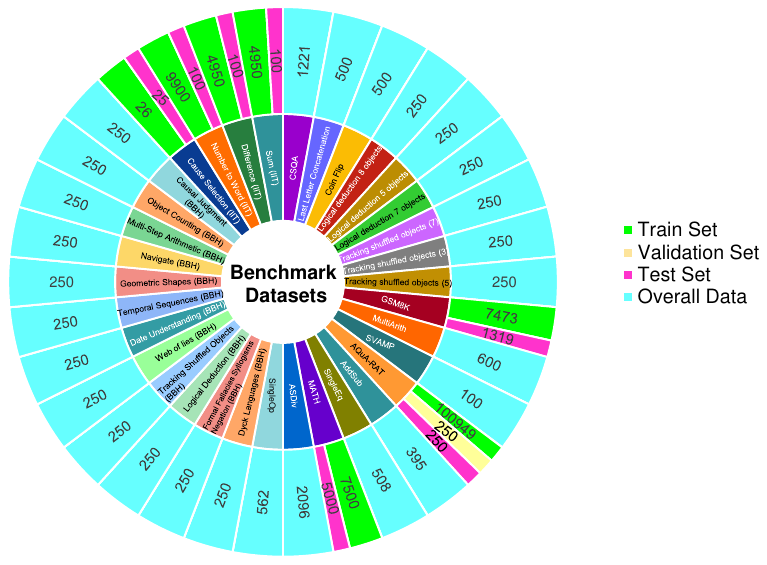}
\caption{Benchmark datasets of reasoning based tasks}
\label{reasoning-datasets}
\end{figure}

Figure \ref{reasoning-datasets} illustrates the sample distribution across thirty-four benchmarking datasets used to evaluate different prompt optimization strategies for reasoning-based tasks. These datasets cover a diverse range of reasoning tasks, including Commonsense reasoning, Symbolic Reasoning, Logical Reasoning, Temporal Reasoning,  Spatial and Geometric Reasoning, Numerical and Arithmetic Reasoning, and Causal and Counterfactual Reasoning. A critical analysis reveals that this domain has been extensively evaluated, encompassing varied reasoning domains. The majority of datasets originate from the BBH benchmark, including Causal Judgment, Object Counting, Web of Lies, Logical Deduction, and Geometric Shapes, each with an overall data size of 250 samples. Additionally, several datasets from the IIT benchmark have been incorporated, including SUM (4,950 training, 100 test), Difference (4,950 training, 100 test), Number of Words (9,900 training, 100 test), and Cause Selection (26 training, 25 test). This dataset diversity ensures a comprehensive evaluation of reasoning-based tasks across different prompt optimization strategies. Despite the large number of datasets, the variation in dataset sizes remains relatively limited. The majority of datasets contain 250 samples (Logical Deduction: 5, 7, 8 objects), while some (Last Letter Concatenation, Coin Flip) have 500 samples. A few datasets, such as GSM8K (8,792 samples) and MATH (12,500 samples), contain larger sample sizes, with AQuA-RAT standing out with 101,449 samples. This pattern suggests that while reasoning-based tasks have been widely explored, the distribution of dataset sizes remains skewed, with only a few large-scale datasets. The high number of benchmarks with relatively small and uniform sample sizes allows controlled comparisons, but it also raises concerns about whether models trained on smaller datasets can generalize effectively to different prompts. Moreover, the presence of diverse reasoning domains enables a detailed analysis of prompt optimization strategies, but ensuring balanced dataset splits across various reasoning subdomains remains a key challenge for fair and interpretable evaluations.
\section{Evaluation of Prompt Optimization Strategies Across Pretrained Language Models}

\begin{landscape}
\begin{figure}
    \centering
    \includegraphics[width=1\linewidth]{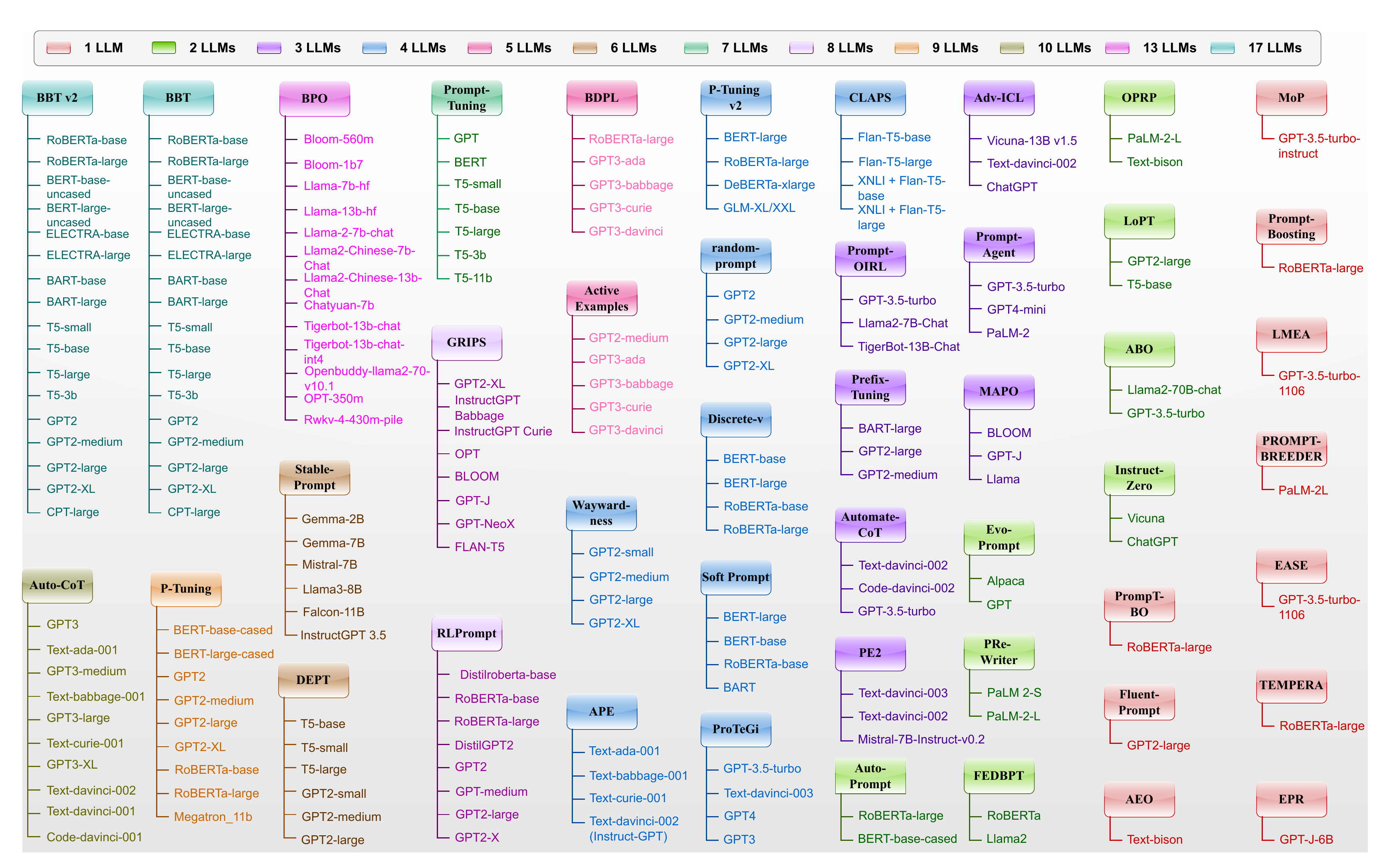}
    \caption{Graphical representation of pretrained language models (PLMs) employed across 45 distinct prompt optimization strategies}
    \label{LMs}
\end{figure}
\end{landscape}

\normalsize
%%%%%%%%%%%%%%%%%%%%%%%%%%%%%%%%%%%%%%%%%%%%%%%%%%%%%%%%%%%%%%%%%%%%%%%%
%Trends
Figure \ref{LMs} presents a comprehensive overview of pre-trained language models used across 45 unique prompt optimization strategies. It is evident from Table \ref{LMs} that numerous methods explore a wide range of pre-trained language models, a significant portion approximately about 21\% of the approaches rely on just a single PLM thus limiting their adaptability. In contrast, 34\% utilize 2 models, while an equal proportion (34\%) utilize 4 language models, 7\% of the methods utilize 3 models, 8\% leverage 5 models, and 18\% incorporate between 6 to 17 pre-trained language models. Despite this diversity, only PROPANE was used for large-scale benchmarking, accounting for just 2\% of the total. This limited adoption of extensive benchmarking raises concerns about the external validity and generalizability of findings across the prompt optimization field. The reliance on smaller-scale evaluations may fail to adequately capture the complexities of prompt optimization in real-world scenarios. Hence, it is essential to standardized efforts towards large-scale benchmarks to advance the field and facilitate robust and transferable insights.

Moreover, based on architecture, pre-trained language models can be categorized into three types: (1) encoder-based models, (2) decoder-based models, and (3) encoder-decoder models. A critical analysis of Table \ref{LMs} reveals that decoder-based models are the most prevalent, utilized by approximately 23 approaches, including EPR, EASE, LMEA, MoP, FluentPrompt, AEO \cite{hsieh2023automatic}, ABO, InstructZero, EvoPrompt, Adv-ICL, MAPO, PE2, Automate-CoT, Prompt-OIRL, ProTeGi, random-prompt, APE, Waywardness, Active Examples, -tuning, Auto-CoT, BPO, and PROPANE. In contrast, five approaches employ encoder-based models namely TEMPERA, PromptBoosting, PrompT-BO, AUTOPROMPT, and Discrete-v \cite{ju2023continuous}. While only three approaches leverage encoder-decoder models: PROMPTBREEDER, PRewriter and RLPromp. Beyond these distinct categories, a substantial number of approaches, approximately 11, explored combinations of models. Notably, seven approaches utilize both encoder-based and decoder-based models: OPRO, PROMPTAGENT, Prefix-Tuning, DEPT, PROMST, StablePRompt, and GRIPS. While only Soft Prompt employ both encoder-based and encoder-decoder models. Furthermore, three approaches namely Prompt-Tuning, BBT, and BBTv2, employ all three types of models, encoder-based, decoder-based, and encoder-decoder models. While the predominance of decoder-based models highlights their current popularity, it also raises concerns that the potential of encoder and encoder-decoder models may be under-explored. This disparity suggests a need for more balanced exploration across all model types to fully understand their potential and limitations. \\

Moreover, Figure \ref{LMs} highlights the dominance of GPT-based models in prompt-based approaches, as 31 out of 45 unique methods incorporating them. These include EPR, EASE, LMEA, MoP, FluentPrompt, InstructZero, LoPT, ABO, EvoPrompt, PROMPTAGENT, MAPO, Adv-ICL, Automate-CoT, Prompt-OIRL, Prefix-Tuning, ProTeGi, random-prompt, APE, Active Examples, DEPT, StablePRompt, Waywardness, BDPL, PROMST, Prompt-Tuning, RLPrompt, P-tuning, GRIPS, Auto-CoT, BBT, and BBTv2. Additionally, BERT family, including BERT, RoBERTa, DistilBERT, DeBERTa, and Electra, has been utilized in 14 approaches, such as TEMPERA, PromptBoosting, PrompT-BO, FEDBT, AUTOPROMPT, SoftPrompt, Discrete-v, P-tuning v2, BDPL, Prompt-Tuning, RLPrompt, P-tuning, BBT, and BBTv2. In contrast, only seven approaches including LoPT, CLAPS, DEPT, Prompt-Tuning, GRIPS, BBT, and BBTv2, utilize T5-based models (T5, Flan-T5). Similarly, LLaMA models are leveraged by seven methods including ABO, FEDBT, MAPO, Prompt-OIRL, StablePRompt, BPO, and PROPANE. Meanwhile, models such as PaLM, OPO, BLOOM, Mistral, Gemma, and Vicuna are explored on a smaller scale. \\

Furthermore, among 45 unique approaches, 22 approaches including PROMPTTAGENT, PRewriter, Prefix-Tuning, CLAPS, ProTeGi, Soft Prompt, Discrete-v, random-prompt, Waywardness, Active Examples, BDPL, DEPT, PROMST, StablePRompt, Prompt-Tuning, RLPrompt, P-tuning, Auto-CoT, BPO, BBT, BBTv2 and PROPANE, utilized multiple sizes of the same model (such as GPT, T5, BERT, RoBERTa, PaLM-2, Gemma, LLaMA, Vicuna, TigerBot), which highlights the effect of model scalability on the efficiency of prompting techniques. Few-shot and zero-shot methods including Auto-CoT, RLPrompt, Automate-CoT, EvoPrompt, APE, PE2, MoP and PROMPTAGENT favor large language models (GPT-3.5, GPT-4, FLAN-T5) because they rely on in-context learning rather than fine-tuning. While smaller-scale tuning-based methods, such as Soft Prompt, PromptBoosting, PrompT-BO, AUTOPROMPT and Discrete-v tend to test on fewer and smaller models (e.g., BERT, RoBERTa).

%%%%%%%%%%%%%%%%%%%%%%%%%%%%%%%%%%%%%%%%%%%%%%%%%%%%%%%%%%%%%%%%%%%%%%%%

%%%%%%%%%%%%%%%%%%%%%%%%%%%%%%%%%%%%%%%%%%%%%%%%%%%%%%%%%%%%%%%%%%%%%%%%
% Encoder only models 5 (TEMPERA, PromptBoosting, PrompT-BO, AUTOPROMPT, Discrete-v, 

% Decoder only models 22 (EPR, EASE , LMEA, MoP, FluentPrompt, AEO, ABO, InstructZero, EvoPrompt, Adv-ICL, MAPO, PE2, Automate-CoT, Prompt-OIRL, ProTeGi, random-prompt, APE, Waywardness, Active Examples, Auto-CoT, BPO, PROPANE

% Encoder + Decoder + Encoder-Decoder models 3 (Prompt-Tuning, BBT, BBTv2)
%%%%%%%%%%%%%%%%%%%%%%%%%%%%%%%%%%%%%%%%%%%%%%%%%%%%%%%%%%%%%%%%%%%%%%%%

% 1 method utilized 7 pre-trained models (Prompt-Tuning)
% 2 (4%) methods utilized 8 pre-trained models (RLPrompt, GRIPS)
% 1 method utilized 9 pre-trained models (P-tuning )
% 1 method utilized 10 pre-trained models (Auto-CoT)
% 1 method utilized 13 pre-trained models (BPO)
% 2 methods utilized 17 pre-trained models (BBT, BBTv2)
% 1 method utilized 31 pre-trained models (PROPANE)

%%%%%%%%%%%%%%%%%%%%%%%%%%%%%%%%%%%%%%%%%%%%%%%%%%%%%%%%%%%%%%%%%%%%%
% 14 (30) Encoder-based models (RLPrompt, BDPL, TEMPERA, P-tuning v2, P-tuning, BBT, BBTv2, FEDBPT, Soft Prompt, Discrete-v, PromptBoosting, Prompt-BO, AUTOPROMPT, Prompt-Tuning)

\section{Discussion and Conclusion}
This study analyzes 45 different prompt optimization strategies in the field of NLP. This analysis rigorously structured these methods based on working paradigms, (e.g, single-layer or multi-layer interactions, reinforcement learning,  evolutionary strategies,  Bayesian optimization). Furthermore, the review also classified methods based on the specific pre-trained language models (PLMs) employed, the types of targeted tasks, and benchmark datasets to offer a holistic overview of the field. The evolution of prompt optimization reveals a clear shift from basic prompt tuning methods to complex frameworks that employ deep model-layer interacions, black-box optimization, interpretable prompts, and collaboration between humans and LLMs.

A critical finding of this review underscores the inconsistency in existing benchmarking practices across prompt optimization research. This lack of standardization significantly hinders the ability to directly compare different studies, primarily because they employ divergent datasets, evaluation metrics, or formulate tasks in dissimilar ways. Consequently, the absence of a unified evaluation framework makes it difficult to achieve comparable results and to measure the real progress of prompt optimization field. Furthermore, numerous methods have been validated on only a limited array of tasks typically such as sentiment analysis, question answering, or classification and have yet to be rigorously tested on more diverse, open-ended, or cross-lingual benchmarks.

Another key limitation identified through this review is the limited diversity PLMs used to evaluate prompt optimization strategies. While a few well-known models such as GPT-2, GPT-3 and BERT variants are frequently tested, many recent and diverse PLMs including multilingual, instruction-tuned, domain-specific and lightweight models remain largely unexplored. This narrow focus constrains our understanding of how well prompt optimization strategies generalize across architectures with different tokenization schemes, training objectives, and parameter scales. Exploring a broader range of PLMs would help uncover the strengths and limitations of each method and support the development of more robust, model-agnostic prompting strategies. From a methodological perspective, a shift is observed toward hybrid strategies that blend human knowledge (e.g., through instruction tuning or human-in-the-loop evaluation) with automated search techniques such as reinforcement learning and Bayesian optimization. These collaborative models offer a promising direction, especially when dealing with black-box LLMs or high-stakes decision-making tasks.

\subsection{Future Directions}
 Future work should focus on establishing standardized benchmark datasets and evaluation protocols for prompt optimization. A common suite of tasks and metrics would foster fair comparison and meaningful performance insights. Moreover, new methods should be evaluated across diverse NLP tasks, including generation, reasoning, dialogue, and multilingual settings, to assess their robustness and adaptability beyond simple classification tasks. Many current methods are computationally expensive or require fine-tuning at scale. Research should aim at developing more lightweight frameworks that can generalize with minimal training overhead. There is room to explore the full capabilities of PLMs including prompt chaining, model feedback loops, and more interpretable prompt designs that align with the underlying model architecture.
 The integration of human expertise with model-based optimization should be investigated comprehensively. Systems that allow human-guided prompt shaping with automated feedback loops are likely to yield more controllable and aligned model behaviour.  Ultimately, the utility of prompt optimization strategies should be validated in real-world applications such as biomedical NLP, multilingual summarization and educational dialogue systems.

\backmatter

\bibliography{sn-bibliography}


\begin{thebibliography}{149}
\ifx \bisbn   \undefined \def \bisbn  #1{ISBN #1}\fi
\ifx \binits  \undefined \def \binits#1{#1}\fi
\ifx \bauthor  \undefined \def \bauthor#1{#1}\fi
\ifx \batitle  \undefined \def \batitle#1{#1}\fi
\ifx \bjtitle  \undefined \def \bjtitle#1{#1}\fi
\ifx \bvolume  \undefined \def \bvolume#1{\textbf{#1}}\fi
\ifx \byear  \undefined \def \byear#1{#1}\fi
\ifx \bissue  \undefined \def \bissue#1{#1}\fi
\ifx \bfpage  \undefined \def \bfpage#1{#1}\fi
\ifx \blpage  \undefined \def \blpage #1{#1}\fi
\ifx \burl  \undefined \def \burl#1{\textsf{#1}}\fi
\ifx \doiurl  \undefined \def \doiurl#1{\url{https://doi.org/#1}}\fi
\ifx \betal  \undefined \def \betal{\textit{et al.}}\fi
\ifx \binstitute  \undefined \def \binstitute#1{#1}\fi
\ifx \binstitutionaled  \undefined \def \binstitutionaled#1{#1}\fi
\ifx \bctitle  \undefined \def \bctitle#1{#1}\fi
\ifx \beditor  \undefined \def \beditor#1{#1}\fi
\ifx \bpublisher  \undefined \def \bpublisher#1{#1}\fi
\ifx \bbtitle  \undefined \def \bbtitle#1{#1}\fi
\ifx \bedition  \undefined \def \bedition#1{#1}\fi
\ifx \bseriesno  \undefined \def \bseriesno#1{#1}\fi
\ifx \blocation  \undefined \def \blocation#1{#1}\fi
\ifx \bsertitle  \undefined \def \bsertitle#1{#1}\fi
\ifx \bsnm \undefined \def \bsnm#1{#1}\fi
\ifx \bsuffix \undefined \def \bsuffix#1{#1}\fi
\ifx \bparticle \undefined \def \bparticle#1{#1}\fi
\ifx \barticle \undefined \def \barticle#1{#1}\fi
\bibcommenthead
\ifx \bconfdate \undefined \def \bconfdate #1{#1}\fi
\ifx \botherref \undefined \def \botherref #1{#1}\fi
\ifx \url \undefined \def \url#1{\textsf{#1}}\fi
\ifx \bchapter \undefined \def \bchapter#1{#1}\fi
\ifx \bbook \undefined \def \bbook#1{#1}\fi
\ifx \bcomment \undefined \def \bcomment#1{#1}\fi
\ifx \oauthor \undefined \def \oauthor#1{#1}\fi
\ifx \citeauthoryear \undefined \def \citeauthoryear#1{#1}\fi
\ifx \endbibitem  \undefined \def \endbibitem {}\fi
\ifx \bconflocation  \undefined \def \bconflocation#1{#1}\fi
\ifx \arxivurl  \undefined \def \arxivurl#1{\textsf{#1}}\fi
\csname PreBibitemsHook\endcsname

%%% 1
\bibitem[\protect\citeauthoryear{Fakhoury et~al.}{2024}]{fakhoury2024llm}
\begin{botherref}
\oauthor{\bsnm{Fakhoury}, \binits{S.}},
\oauthor{\bsnm{Naik}, \binits{A.}},
\oauthor{\bsnm{Sakkas}, \binits{G.}},
\oauthor{\bsnm{Chakraborty}, \binits{S.}},
\oauthor{\bsnm{Lahiri}, \binits{S.K.}}:
Llm-based test-driven interactive code generation: User study and empirical evaluation.
IEEE Transactions on Software Engineering
(2024)
\end{botherref}
\endbibitem

%%% 2
\bibitem[\protect\citeauthoryear{Huang et~al.}{2024}]{huang2024bias}
\begin{botherref}
\oauthor{\bsnm{Huang}, \binits{D.}},
\oauthor{\bsnm{Zhang}, \binits{J.M.}},
\oauthor{\bsnm{Bu}, \binits{Q.}},
\oauthor{\bsnm{Xie}, \binits{X.}},
\oauthor{\bsnm{Chen}, \binits{J.}},
\oauthor{\bsnm{Cui}, \binits{H.}}:
Bias testing and mitigation in llm-based code generation.
ACM Transactions on Software Engineering and Methodology
(2024)
\end{botherref}
\endbibitem

%%% 3
\bibitem[\protect\citeauthoryear{Enis and Hopkins}{2024}]{enis2024llm}
\begin{botherref}
\oauthor{\bsnm{Enis}, \binits{M.}},
\oauthor{\bsnm{Hopkins}, \binits{M.}}:
From llm to nmt: Advancing low-resource machine translation with claude.
arXiv preprint arXiv:2404.13813
(2024)
\end{botherref}
\endbibitem

%%% 4
\bibitem[\protect\citeauthoryear{Elshin et~al.}{2024}]{elshin2024general}
\begin{bchapter}
\bauthor{\bsnm{Elshin}, \binits{D.}},
\bauthor{\bsnm{Karpachev}, \binits{N.}},
\bauthor{\bsnm{Gruzdev}, \binits{B.}},
\bauthor{\bsnm{Golovanov}, \binits{I.}},
\bauthor{\bsnm{Ivanov}, \binits{G.}},
\bauthor{\bsnm{Antonov}, \binits{A.}},
\bauthor{\bsnm{Skachkov}, \binits{N.}},
\bauthor{\bsnm{Latypova}, \binits{E.}},
\bauthor{\bsnm{Layner}, \binits{V.}},
\bauthor{\bsnm{Enikeeva}, \binits{E.}}, \betal:
\bctitle{From general llm to translation: How we dramatically improve translation quality using human evaluation data for llm finetuning}.
In: \bbtitle{Proceedings of the Ninth Conference on Machine Translation},
pp. \bfpage{247}--\blpage{252}
(\byear{2024})
\end{bchapter}
\endbibitem

%%% 5
\bibitem[\protect\citeauthoryear{Wu and Deng}{2025}]{wu2025implementing}
\begin{botherref}
\oauthor{\bsnm{Wu}, \binits{Y.}},
\oauthor{\bsnm{Deng}, \binits{X.}}:
Implementing long text style transfer with llms through dual-layered sentence and paragraph structure extraction and mapping.
arXiv preprint arXiv:2505.07888
(2025)
\end{botherref}
\endbibitem

%%% 6
\bibitem[\protect\citeauthoryear{Asim et~al.}{2022}]{asim2022lgca}
\begin{barticle}
\bauthor{\bsnm{Asim}, \binits{M.N.}},
\bauthor{\bsnm{Ibrahim}, \binits{M.A.}},
\bauthor{\bsnm{Malik}, \binits{M.I.}},
\bauthor{\bsnm{Dengel}, \binits{A.}},
\bauthor{\bsnm{Ahmed}, \binits{S.}}:
\batitle{Lgca-vhppi: A local-global residue context aware viral-host protein-protein interaction predictor}.
\bjtitle{Plos one}
\bvolume{17}(\bissue{7}),
\bfpage{0270275}
(\byear{2022})
\end{barticle}
\endbibitem

%%% 7
\bibitem[\protect\citeauthoryear{Saleem et~al.}{2025}]{saleem2025generative}
\begin{barticle}
\bauthor{\bsnm{Saleem}, \binits{S.}},
\bauthor{\bsnm{Asim}, \binits{M.N.}},
\bauthor{\bsnm{Elst}, \binits{L.V.}},
\bauthor{\bsnm{Dengel}, \binits{A.}}:
\batitle{Generative language models potential for requirement engineering applications: insights into current strengths and limitations}.
\bjtitle{Complex \& Intelligent Systems}
\bvolume{11}(\bissue{6}),
\bfpage{1}--\blpage{22}
(\byear{2025})
\end{barticle}
\endbibitem

%%% 8
\bibitem[\protect\citeauthoryear{Saleem et~al.}{2024}]{saleem2024passionnet}
\begin{botherref}
\oauthor{\bsnm{Saleem}, \binits{S.}},
\oauthor{\bsnm{Asim}, \binits{M.N.}},
\oauthor{\bsnm{Dengel}, \binits{A.}}:
Passionnet: An innovative framework for duplicate and conflicting requirements identification.
arXiv preprint arXiv:2412.01657
(2024)
\end{botherref}
\endbibitem

%%% 9
\bibitem[\protect\citeauthoryear{Raiaan et~al.}{2024}]{raiaan2024review}
\begin{barticle}
\bauthor{\bsnm{Raiaan}, \binits{M.A.K.}},
\bauthor{\bsnm{Mukta}, \binits{M.S.H.}},
\bauthor{\bsnm{Fatema}, \binits{K.}},
\bauthor{\bsnm{Fahad}, \binits{N.M.}},
\bauthor{\bsnm{Sakib}, \binits{S.}},
\bauthor{\bsnm{Mim}, \binits{M.M.J.}},
\bauthor{\bsnm{Ahmad}, \binits{J.}},
\bauthor{\bsnm{Ali}, \binits{M.E.}},
\bauthor{\bsnm{Azam}, \binits{S.}}:
\batitle{A review on large language models: Architectures, applications, taxonomies, open issues and challenges}.
\bjtitle{IEEE access}
\bvolume{12},
\bfpage{26839}--\blpage{26874}
(\byear{2024})
\end{barticle}
\endbibitem

%%% 10
\bibitem[\protect\citeauthoryear{Devlin}{2018}]{devlin2018bert}
\begin{botherref}
\oauthor{\bsnm{Devlin}, \binits{J.}}:
Bert: Pre-training of deep bidirectional transformers for language understanding.
arXiv preprint arXiv:1810.04805
(2018)
\end{botherref}
\endbibitem

%%% 11
\bibitem[\protect\citeauthoryear{{NVIDIA Developer}}{2021}]{nvidiamegatronturing}
\begin{botherref}
\oauthor{\bsnm{{NVIDIA Developer}}}:
Using {DeepSpeed} and {Megatron} to Train {Megatron-Turing NLG} 530B, the World's Largest and Most Powerful Generative Language Model.
\url{https://developer.nvidia.com/blog/using-deepspeed-and-megatron-to-train-megatron-turing-nlg-530b-the-worlds-largest-and-most-powerful-generative-language-model/}.
Accessed: 2024-12-23
(2021)
\end{botherref}
\endbibitem

%%% 12
\bibitem[\protect\citeauthoryear{Bonfigli et~al.}{2024}]{bonfigli2024pre}
\begin{barticle}
\bauthor{\bsnm{Bonfigli}, \binits{A.}},
\bauthor{\bsnm{Bacco}, \binits{L.}},
\bauthor{\bsnm{Merone}, \binits{M.}},
\bauthor{\bsnm{Dell’Orletta}, \binits{F.}}:
\batitle{From pre-training to fine-tuning: An in-depth analysis of large language models in the biomedical domain}.
\bjtitle{Artificial Intelligence in Medicine}
\bvolume{157},
\bfpage{103003}
(\byear{2024})
\end{barticle}
\endbibitem

%%% 13
\bibitem[\protect\citeauthoryear{Council}{2024}]{forbes_ai_carbon}
\begin{botherref}
\oauthor{\bsnm{Council}, \binits{F.T.}}:
The Untold Story Of AI's Huge Carbon Footprint.
\url{https://www.forbes.com/councils/forbestechcouncil/2024/04/26/the-untold-story-of-ais-huge-carbon-footprint/}
(2024).
\url{https://www.forbes.com/councils/forbestechcouncil/2024/04/26/the-untold-story-of-ais-huge-carbon-footprint/}
Accessed 2024-12-23
\end{botherref}
\endbibitem

%%% 14
\bibitem[\protect\citeauthoryear{Chen et~al.}{2023}]{chen2023unleashing}
\begin{botherref}
\oauthor{\bsnm{Chen}, \binits{B.}},
\oauthor{\bsnm{Zhang}, \binits{Z.}},
\oauthor{\bsnm{Langren{\'e}}, \binits{N.}},
\oauthor{\bsnm{Zhu}, \binits{S.}}:
Unleashing the potential of prompt engineering in large language models: a comprehensive review.
arXiv preprint arXiv:2310.14735
(2023)
\end{botherref}
\endbibitem

%%% 15
\bibitem[\protect\citeauthoryear{Qin and Eisner}{2021}]{qin2021learning}
\begin{botherref}
\oauthor{\bsnm{Qin}, \binits{G.}},
\oauthor{\bsnm{Eisner}, \binits{J.}}:
Learning how to ask: Querying lms with mixtures of soft prompts.
arXiv preprint arXiv:2104.06599
(2021)
\end{botherref}
\endbibitem

%%% 16
\bibitem[\protect\citeauthoryear{Wen et~al.}{2023}]{wen2023hard}
\begin{barticle}
\bauthor{\bsnm{Wen}, \binits{Y.}},
\bauthor{\bsnm{Jain}, \binits{N.}},
\bauthor{\bsnm{Kirchenbauer}, \binits{J.}},
\bauthor{\bsnm{Goldblum}, \binits{M.}},
\bauthor{\bsnm{Geiping}, \binits{J.}},
\bauthor{\bsnm{Goldstein}, \binits{T.}}:
\batitle{Hard prompts made easy: Gradient-based discrete optimization for prompt tuning and discovery}.
\bjtitle{Advances in Neural Information Processing Systems}
\bvolume{36},
\bfpage{51008}--\blpage{51025}
(\byear{2023})
\end{barticle}
\endbibitem

%%% 17
\bibitem[\protect\citeauthoryear{Dalibor et~al.}{2022}]{dalibor2022generating}
\begin{barticle}
\bauthor{\bsnm{Dalibor}, \binits{M.}},
\bauthor{\bsnm{Heithoff}, \binits{M.}},
\bauthor{\bsnm{Michael}, \binits{J.}},
\bauthor{\bsnm{Netz}, \binits{L.}},
\bauthor{\bsnm{Pfeiffer}, \binits{J.}},
\bauthor{\bsnm{Rumpe}, \binits{B.}},
\bauthor{\bsnm{Varga}, \binits{S.}},
\bauthor{\bsnm{Wortmann}, \binits{A.}}:
\batitle{Generating customized low-code development platforms for digital twins}.
\bjtitle{Journal of Computer Languages}
\bvolume{70},
\bfpage{101117}
(\byear{2022})
\end{barticle}
\endbibitem

%%% 18
\bibitem[\protect\citeauthoryear{Di~Ruscio et~al.}{2022}]{di2022low}
\begin{barticle}
\bauthor{\bsnm{Di~Ruscio}, \binits{D.}},
\bauthor{\bsnm{Kolovos}, \binits{D.}},
\bauthor{\bsnm{Lara}, \binits{J.}},
\bauthor{\bsnm{Pierantonio}, \binits{A.}},
\bauthor{\bsnm{Tisi}, \binits{M.}},
\bauthor{\bsnm{Wimmer}, \binits{M.}}:
\batitle{Low-code development and model-driven engineering: Two sides of the same coin?}
\bjtitle{Software and Systems Modeling}
\bvolume{21}(\bissue{2}),
\bfpage{437}--\blpage{446}
(\byear{2022})
\end{barticle}
\endbibitem

%%% 19
\bibitem[\protect\citeauthoryear{Shin et~al.}{2020}]{shin2020autoprompt}
\begin{botherref}
\oauthor{\bsnm{Shin}, \binits{T.}},
\oauthor{\bsnm{Razeghi}, \binits{Y.}},
\oauthor{\bsnm{Logan~IV}, \binits{R.L.}},
\oauthor{\bsnm{Wallace}, \binits{E.}},
\oauthor{\bsnm{Singh}, \binits{S.}}:
Autoprompt: Eliciting knowledge from language models with automatically generated prompts.
arXiv preprint arXiv:2010.15980
(2020)
\end{botherref}
\endbibitem

%%% 20
\bibitem[\protect\citeauthoryear{Shi et~al.}{2022}]{shi2022toward}
\begin{botherref}
\oauthor{\bsnm{Shi}, \binits{W.}},
\oauthor{\bsnm{Han}, \binits{X.}},
\oauthor{\bsnm{Gonen}, \binits{H.}},
\oauthor{\bsnm{Holtzman}, \binits{A.}},
\oauthor{\bsnm{Tsvetkov}, \binits{Y.}},
\oauthor{\bsnm{Zettlemoyer}, \binits{L.}}:
Toward human readable prompt tuning: Kubrick's the shining is a good movie, and a good prompt too?
arXiv preprint arXiv:2212.10539
(2022)
\end{botherref}
\endbibitem

%%% 21
\bibitem[\protect\citeauthoryear{Lester et~al.}{2021}]{lester2021power}
\begin{botherref}
\oauthor{\bsnm{Lester}, \binits{B.}},
\oauthor{\bsnm{Al-Rfou}, \binits{R.}},
\oauthor{\bsnm{Constant}, \binits{N.}}:
The power of scale for parameter-efficient prompt tuning.
arXiv preprint arXiv:2104.08691
(2021)
\end{botherref}
\endbibitem

%%% 22
\bibitem[\protect\citeauthoryear{Liu et~al.}{2024}]{liu2024gpt}
\begin{barticle}
\bauthor{\bsnm{Liu}, \binits{X.}},
\bauthor{\bsnm{Zheng}, \binits{Y.}},
\bauthor{\bsnm{Du}, \binits{Z.}},
\bauthor{\bsnm{Ding}, \binits{M.}},
\bauthor{\bsnm{Qian}, \binits{Y.}},
\bauthor{\bsnm{Yang}, \binits{Z.}},
\bauthor{\bsnm{Tang}, \binits{J.}}:
\batitle{Gpt understands, too}.
\bjtitle{AI Open}
\bvolume{5},
\bfpage{208}--\blpage{215}
(\byear{2024})
\end{barticle}
\endbibitem

%%% 23
\bibitem[\protect\citeauthoryear{Sun et~al.}{2022}]{sun2022black}
\begin{bchapter}
\bauthor{\bsnm{Sun}, \binits{T.}},
\bauthor{\bsnm{Shao}, \binits{Y.}},
\bauthor{\bsnm{Qian}, \binits{H.}},
\bauthor{\bsnm{Huang}, \binits{X.}},
\bauthor{\bsnm{Qiu}, \binits{X.}}:
\bctitle{Black-box tuning for language-model-as-a-service}.
In: \bbtitle{International Conference on Machine Learning},
pp. \bfpage{20841}--\blpage{20855}
(\byear{2022}).
\bcomment{PMLR}
\end{bchapter}
\endbibitem

%%% 24
\bibitem[\protect\citeauthoryear{Sun et~al.}{2023}]{sun2023fedbpt}
\begin{botherref}
\oauthor{\bsnm{Sun}, \binits{J.}},
\oauthor{\bsnm{Xu}, \binits{Z.}},
\oauthor{\bsnm{Yin}, \binits{H.}},
\oauthor{\bsnm{Yang}, \binits{D.}},
\oauthor{\bsnm{Xu}, \binits{D.}},
\oauthor{\bsnm{Chen}, \binits{Y.}},
\oauthor{\bsnm{Roth}, \binits{H.R.}}:
Fedbpt: Efficient federated black-box prompt tuning for large language models.
arXiv preprint arXiv:2310.01467
(2023)
\end{botherref}
\endbibitem

%%% 25
\bibitem[\protect\citeauthoryear{Shi and Lipani}{2023}]{shi2023dept}
\begin{botherref}
\oauthor{\bsnm{Shi}, \binits{Z.}},
\oauthor{\bsnm{Lipani}, \binits{A.}}:
Dept: Decomposed prompt tuning for parameter-efficient fine-tuning.
arXiv preprint arXiv:2309.05173
(2023)
\end{botherref}
\endbibitem

%%% 26
\bibitem[\protect\citeauthoryear{Guo et~al.}{2024}]{guo2024lopt}
\begin{botherref}
\oauthor{\bsnm{Guo}, \binits{S.}},
\oauthor{\bsnm{Damani}, \binits{S.}},
\oauthor{\bsnm{Chang}, \binits{K.-h.}}:
Lopt: Low-rank prompt tuning for parameter efficient language models.
arXiv preprint arXiv:2406.19486
(2024)
\end{botherref}
\endbibitem

%%% 27
\bibitem[\protect\citeauthoryear{Li and Liang}{2021}]{li2021prefix}
\begin{botherref}
\oauthor{\bsnm{Li}, \binits{X.L.}},
\oauthor{\bsnm{Liang}, \binits{P.}}:
Prefix-tuning: Optimizing continuous prompts for generation.
arXiv preprint arXiv:2101.00190
(2021)
\end{botherref}
\endbibitem

%%% 28
\bibitem[\protect\citeauthoryear{Liu et~al.}{2021}]{liu2021p}
\begin{botherref}
\oauthor{\bsnm{Liu}, \binits{X.}},
\oauthor{\bsnm{Ji}, \binits{K.}},
\oauthor{\bsnm{Fu}, \binits{Y.}},
\oauthor{\bsnm{Tam}, \binits{W.L.}},
\oauthor{\bsnm{Du}, \binits{Z.}},
\oauthor{\bsnm{Yang}, \binits{Z.}},
\oauthor{\bsnm{Tang}, \binits{J.}}:
P-tuning v2: Prompt tuning can be comparable to fine-tuning universally across scales and tasks.
arXiv preprint arXiv:2110.07602
(2021)
\end{botherref}
\endbibitem

%%% 29
\bibitem[\protect\citeauthoryear{Sun et~al.}{2022}]{sun2022bbtv2}
\begin{botherref}
\oauthor{\bsnm{Sun}, \binits{T.}},
\oauthor{\bsnm{He}, \binits{Z.}},
\oauthor{\bsnm{Qian}, \binits{H.}},
\oauthor{\bsnm{Zhou}, \binits{Y.}},
\oauthor{\bsnm{Huang}, \binits{X.}},
\oauthor{\bsnm{Qiu}, \binits{X.}}:
Bbtv2: Towards a gradient-free future with large language models.
arXiv preprint arXiv:2205.11200
(2022)
\end{botherref}
\endbibitem

%%% 30
\bibitem[\protect\citeauthoryear{Khashabi et~al.}{2022}]{khashabi-etal-2022-prompt}
\begin{bchapter}
\bauthor{\bsnm{Khashabi}, \binits{D.}},
\bauthor{\bsnm{Lyu}, \binits{X.}},
\bauthor{\bsnm{Min}, \binits{S.}},
\bauthor{\bsnm{Qin}, \binits{L.}},
\bauthor{\bsnm{Richardson}, \binits{K.}},
\bauthor{\bsnm{Welleck}, \binits{S.}},
\bauthor{\bsnm{Hajishirzi}, \binits{H.}},
\bauthor{\bsnm{Khot}, \binits{T.}},
\bauthor{\bsnm{Sabharwal}, \binits{A.}},
\bauthor{\bsnm{Singh}, \binits{S.}},
\bauthor{\bsnm{Choi}, \binits{Y.}}:
\bctitle{Prompt waywardness: The curious case of discretized interpretation of continuous prompts}.
In: \bbtitle{Proceedings of the 2022 Conference of the North American Chapter of the Association for Computational Linguistics: Human Language Technologies},
pp. \bfpage{3631}--\blpage{3643}.
\bpublisher{Association for Computational Linguistics},
\blocation{Seattle, United States}
(\byear{2022}).
\doiurl{10.18653/v1/2022.naacl-main.266} .
\burl{https://aclanthology.org/2022.naacl-main.266/}
\end{bchapter}
\endbibitem

%%% 31
\bibitem[\protect\citeauthoryear{Deng et~al.}{2022}]{deng2022rlprompt}
\begin{botherref}
\oauthor{\bsnm{Deng}, \binits{M.}},
\oauthor{\bsnm{Wang}, \binits{J.}},
\oauthor{\bsnm{Hsieh}, \binits{C.-P.}},
\oauthor{\bsnm{Wang}, \binits{Y.}},
\oauthor{\bsnm{Guo}, \binits{H.}},
\oauthor{\bsnm{Shu}, \binits{T.}},
\oauthor{\bsnm{Song}, \binits{M.}},
\oauthor{\bsnm{Xing}, \binits{E.P.}},
\oauthor{\bsnm{Hu}, \binits{Z.}}:
Rlprompt: Optimizing discrete text prompts with reinforcement learning.
arXiv preprint arXiv:2205.12548
(2022)
\end{botherref}
\endbibitem

%%% 32
\bibitem[\protect\citeauthoryear{Diao et~al.}{2023}]{diao2023black}
\begin{botherref}
\oauthor{\bsnm{Diao}, \binits{S.}},
\oauthor{\bsnm{Huang}, \binits{Z.}},
\oauthor{\bsnm{Xu}, \binits{R.}},
\oauthor{\bsnm{Li}, \binits{X.}},
\oauthor{\bsnm{Lin}, \binits{Y.}},
\oauthor{\bsnm{Zhou}, \binits{X.}},
\oauthor{\bsnm{Zhang}, \binits{T.}}:
Black-box prompt learning for pre-trained language models.
Transactions on Machine Learning Research
\textbf{2023}
(2023)
\end{botherref}
\endbibitem

%%% 33
\bibitem[\protect\citeauthoryear{Zhang et~al.}{2022}]{zhang2022tempera}
\begin{botherref}
\oauthor{\bsnm{Zhang}, \binits{T.}},
\oauthor{\bsnm{Wang}, \binits{X.}},
\oauthor{\bsnm{Zhou}, \binits{D.}},
\oauthor{\bsnm{Schuurmans}, \binits{D.}},
\oauthor{\bsnm{Gonzalez}, \binits{J.E.}}:
Tempera: Test-time prompting via reinforcement learning.
arXiv preprint arXiv:2211.11890
(2022)
\end{botherref}
\endbibitem

%%% 34
\bibitem[\protect\citeauthoryear{Chen et~al.}{2024}]{chen2024mapo}
\begin{botherref}
\oauthor{\bsnm{Chen}, \binits{Y.}},
\oauthor{\bsnm{Wen}, \binits{Z.}},
\oauthor{\bsnm{Fan}, \binits{G.}},
\oauthor{\bsnm{Chen}, \binits{Z.}},
\oauthor{\bsnm{Wu}, \binits{W.}},
\oauthor{\bsnm{Liu}, \binits{D.}},
\oauthor{\bsnm{Li}, \binits{Z.}},
\oauthor{\bsnm{Liu}, \binits{B.}},
\oauthor{\bsnm{Xiao}, \binits{Y.}}:
Mapo: Boosting large language model performance with model-adaptive prompt optimization.
arXiv preprint arXiv:2407.04118
(2024)
\end{botherref}
\endbibitem

%%% 35
\bibitem[\protect\citeauthoryear{Kong et~al.}{2024}]{kong2024prewrite}
\begin{botherref}
\oauthor{\bsnm{Kong}, \binits{W.}},
\oauthor{\bsnm{Hombaiah}, \binits{S.A.}},
\oauthor{\bsnm{Zhang}, \binits{M.}},
\oauthor{\bsnm{Mei}, \binits{Q.}},
\oauthor{\bsnm{Bendersky}, \binits{M.}}:
Prewrite: Prompt rewriting with reinforcement learning.
arXiv preprint arXiv:2401.08189
(2024)
\end{botherref}
\endbibitem

%%% 36
\bibitem[\protect\citeauthoryear{Kwon et~al.}{2024}]{kwon2024stableprompt}
\begin{botherref}
\oauthor{\bsnm{Kwon}, \binits{M.}},
\oauthor{\bsnm{Kim}, \binits{G.}},
\oauthor{\bsnm{Kim}, \binits{J.}},
\oauthor{\bsnm{Lee}, \binits{H.}},
\oauthor{\bsnm{Kim}, \binits{J.}}:
Stableprompt: Automatic prompt tuning using reinforcement learning for large language models.
arXiv preprint arXiv:2410.07652
(2024)
\end{botherref}
\endbibitem

%%% 37
\bibitem[\protect\citeauthoryear{Sun et~al.}{2023}]{sun2023query}
\begin{botherref}
\oauthor{\bsnm{Sun}, \binits{H.}},
\oauthor{\bsnm{H{\"u}y{\"u}k}, \binits{A.}},
\oauthor{\bsnm{Schaar}, \binits{M.}}:
Query-dependent prompt evaluation and optimization with offline inverse rl.
arXiv preprint arXiv:2309.06553
(2023)
\end{botherref}
\endbibitem

%%% 38
\bibitem[\protect\citeauthoryear{Wang et~al.}{2023}]{wang2023promptagent}
\begin{botherref}
\oauthor{\bsnm{Wang}, \binits{X.}},
\oauthor{\bsnm{Li}, \binits{C.}},
\oauthor{\bsnm{Wang}, \binits{Z.}},
\oauthor{\bsnm{Bai}, \binits{F.}},
\oauthor{\bsnm{Luo}, \binits{H.}},
\oauthor{\bsnm{Zhang}, \binits{J.}},
\oauthor{\bsnm{Jojic}, \binits{N.}},
\oauthor{\bsnm{Xing}, \binits{E.P.}},
\oauthor{\bsnm{Hu}, \binits{Z.}}:
Promptagent: Strategic planning with language models enables expert-level prompt optimization.
arXiv preprint arXiv:2310.16427
(2023)
\end{botherref}
\endbibitem

%%% 39
\bibitem[\protect\citeauthoryear{Prasad et~al.}{2022}]{prasad2022grips}
\begin{botherref}
\oauthor{\bsnm{Prasad}, \binits{A.}},
\oauthor{\bsnm{Hase}, \binits{P.}},
\oauthor{\bsnm{Zhou}, \binits{X.}},
\oauthor{\bsnm{Bansal}, \binits{M.}}:
Grips: Gradient-free, edit-based instruction search for prompting large language models.
arXiv preprint arXiv:2203.07281
(2022)
\end{botherref}
\endbibitem

%%% 40
\bibitem[\protect\citeauthoryear{Zhou et~al.}{2023}]{zhou2023survival}
\begin{botherref}
\oauthor{\bsnm{Zhou}, \binits{H.}},
\oauthor{\bsnm{Wan}, \binits{X.}},
\oauthor{\bsnm{Vuli{\'c}}, \binits{I.}},
\oauthor{\bsnm{Korhonen}, \binits{A.}}:
Survival of the most influential prompts: Efficient black-box prompt search via clustering and pruning.
arXiv preprint arXiv:2310.12774
(2023)
\end{botherref}
\endbibitem

%%% 41
\bibitem[\protect\citeauthoryear{Fernando et~al.}{2023}]{fernando2023promptbreeder}
\begin{botherref}
\oauthor{\bsnm{Fernando}, \binits{C.}},
\oauthor{\bsnm{Banarse}, \binits{D.}},
\oauthor{\bsnm{Michalewski}, \binits{H.}},
\oauthor{\bsnm{Osindero}, \binits{S.}},
\oauthor{\bsnm{Rockt{\"a}schel}, \binits{T.}}:
Promptbreeder: Self-referential self-improvement via prompt evolution.
arXiv preprint arXiv:2309.16797
(2023)
\end{botherref}
\endbibitem

%%% 42
\bibitem[\protect\citeauthoryear{Guo12 et~al.}{}]{guo12connecting}
\begin{botherref}
\oauthor{\bsnm{Guo12}, \binits{Q.}},
\oauthor{\bsnm{Wang}, \binits{R.}},
\oauthor{\bsnm{Guo}, \binits{J.}},
\oauthor{\bsnm{Li23}, \binits{B.}},
\oauthor{\bsnm{Song}, \binits{K.}},
\oauthor{\bsnm{Tan}, \binits{X.}},
\oauthor{\bsnm{Liu}, \binits{G.}},
\oauthor{\bsnm{Bian}, \binits{J.}},
\oauthor{\bsnm{Yang}, \binits{Y.}}:
Connecting large language models with evo-lutionary algorithms yields powerful prompt optimizers
\end{botherref}
\endbibitem

%%% 43
\bibitem[\protect\citeauthoryear{Liu et~al.}{2024}]{liu2024large}
\begin{bchapter}
\bauthor{\bsnm{Liu}, \binits{S.}},
\bauthor{\bsnm{Chen}, \binits{C.}},
\bauthor{\bsnm{Qu}, \binits{X.}},
\bauthor{\bsnm{Tang}, \binits{K.}},
\bauthor{\bsnm{Ong}, \binits{Y.-S.}}:
\bctitle{Large language models as evolutionary optimizers}.
In: \bbtitle{2024 IEEE Congress on Evolutionary Computation (CEC)},
pp. \bfpage{1}--\blpage{8}
(\byear{2024}).
\bcomment{IEEE}
\end{bchapter}
\endbibitem

%%% 44
\bibitem[\protect\citeauthoryear{Rubin et~al.}{2021}]{rubin2021learning}
\begin{botherref}
\oauthor{\bsnm{Rubin}, \binits{O.}},
\oauthor{\bsnm{Herzig}, \binits{J.}},
\oauthor{\bsnm{Berant}, \binits{J.}}:
Learning to retrieve prompts for in-context learning.
arXiv preprint arXiv:2112.08633
(2021)
\end{botherref}
\endbibitem

%%% 45
\bibitem[\protect\citeauthoryear{Zhang et~al.}{2022a}]{zhang2022active}
\begin{botherref}
\oauthor{\bsnm{Zhang}, \binits{Y.}},
\oauthor{\bsnm{Feng}, \binits{S.}},
\oauthor{\bsnm{Tan}, \binits{C.}}:
Active example selection for in-context learning.
arXiv preprint arXiv:2211.04486
(2022)
\end{botherref}
\endbibitem

%%% 46
\bibitem[\protect\citeauthoryear{Zhang et~al.}{2022b}]{zhang2022automatic}
\begin{botherref}
\oauthor{\bsnm{Zhang}, \binits{Z.}},
\oauthor{\bsnm{Zhang}, \binits{A.}},
\oauthor{\bsnm{Li}, \binits{M.}},
\oauthor{\bsnm{Smola}, \binits{A.}}:
Automatic chain of thought prompting in large language models.
arXiv preprint arXiv:2210.03493
(2022)
\end{botherref}
\endbibitem

%%% 47
\bibitem[\protect\citeauthoryear{Shum et~al.}{2023}]{shum2023automatic}
\begin{botherref}
\oauthor{\bsnm{Shum}, \binits{K.}},
\oauthor{\bsnm{Diao}, \binits{S.}},
\oauthor{\bsnm{Zhang}, \binits{T.}}:
Automatic prompt augmentation and selection with chain-of-thought from labeled data.
arXiv preprint arXiv:2302.12822
(2023)
\end{botherref}
\endbibitem

%%% 48
\bibitem[\protect\citeauthoryear{Hou et~al.}{2023}]{hou2023promptboosting}
\begin{bchapter}
\bauthor{\bsnm{Hou}, \binits{B.}},
\bauthor{\bsnm{O’connor}, \binits{J.}},
\bauthor{\bsnm{Andreas}, \binits{J.}},
\bauthor{\bsnm{Chang}, \binits{S.}},
\bauthor{\bsnm{Zhang}, \binits{Y.}}:
\bctitle{Promptboosting: Black-box text classification with ten forward passes}.
In: \bbtitle{International Conference on Machine Learning},
pp. \bfpage{13309}--\blpage{13324}
(\byear{2023}).
\bcomment{PMLR}
\end{bchapter}
\endbibitem

%%% 49
\bibitem[\protect\citeauthoryear{Wang et~al.}{2023}]{wang2023mixture}
\begin{botherref}
\oauthor{\bsnm{Wang}, \binits{R.}},
\oauthor{\bsnm{An}, \binits{S.}},
\oauthor{\bsnm{Cheng}, \binits{M.}},
\oauthor{\bsnm{Zhou}, \binits{T.}},
\oauthor{\bsnm{Hwang}, \binits{S.J.}},
\oauthor{\bsnm{Hsieh}, \binits{C.-J.}}:
Mixture-of-experts in prompt optimization
(2023)
\end{botherref}
\endbibitem

%%% 50
\bibitem[\protect\citeauthoryear{Wu et~al.}{2024}]{wu2024prompt}
\begin{botherref}
\oauthor{\bsnm{Wu}, \binits{Z.}},
\oauthor{\bsnm{Lin}, \binits{X.}},
\oauthor{\bsnm{Dai}, \binits{Z.}},
\oauthor{\bsnm{Hu}, \binits{W.}},
\oauthor{\bsnm{Shu}, \binits{Y.}},
\oauthor{\bsnm{Ng}, \binits{S.-K.}},
\oauthor{\bsnm{Jaillet}, \binits{P.}},
\oauthor{\bsnm{Low}, \binits{B.K.H.}}:
Prompt optimization with ease? efficient ordering-aware automated selection of exemplars.
arXiv preprint arXiv:2405.16122
(2024)
\end{botherref}
\endbibitem

%%% 51
\bibitem[\protect\citeauthoryear{Long et~al.}{2024}]{long2024prompt}
\begin{bchapter}
\bauthor{\bsnm{Long}, \binits{D.}},
\bauthor{\bsnm{Zhao}, \binits{Y.}},
\bauthor{\bsnm{Brown}, \binits{H.}},
\bauthor{\bsnm{Xie}, \binits{Y.}},
\bauthor{\bsnm{Zhao}, \binits{J.}},
\bauthor{\bsnm{Chen}, \binits{N.}},
\bauthor{\bsnm{Kawaguchi}, \binits{K.}},
\bauthor{\bsnm{Shieh}, \binits{M.}},
\bauthor{\bsnm{He}, \binits{J.}}:
\bctitle{Prompt optimization via adversarial in-context learning}.
In: \bbtitle{Proceedings of the 62nd Annual Meeting of the Association for Computational Linguistics (Volume 1: Long Papers)},
pp. \bfpage{7308}--\blpage{7327}
(\byear{2024})
\end{bchapter}
\endbibitem

%%% 52
\bibitem[\protect\citeauthoryear{Pryzant et~al.}{2023}]{pryzant2023automatic}
\begin{botherref}
\oauthor{\bsnm{Pryzant}, \binits{R.}},
\oauthor{\bsnm{Iter}, \binits{D.}},
\oauthor{\bsnm{Li}, \binits{J.}},
\oauthor{\bsnm{Lee}, \binits{Y.T.}},
\oauthor{\bsnm{Zhu}, \binits{C.}},
\oauthor{\bsnm{Zeng}, \binits{M.}}:
Automatic prompt optimization with" gradient descent" and beam search.
arXiv preprint arXiv:2305.03495
(2023)
\end{botherref}
\endbibitem

%%% 53
\bibitem[\protect\citeauthoryear{Yang et~al.}{2023}]{yang2023large}
\begin{botherref}
\oauthor{\bsnm{Yang}, \binits{C.}},
\oauthor{\bsnm{Wang}, \binits{X.}},
\oauthor{\bsnm{Lu}, \binits{Y.}},
\oauthor{\bsnm{Liu}, \binits{H.}},
\oauthor{\bsnm{Le}, \binits{Q.V.}},
\oauthor{\bsnm{Zhou}, \binits{D.}},
\oauthor{\bsnm{Chen}, \binits{X.}}:
Large language models as optimizers.
arXiv preprint arXiv:2309.03409
(2023)
\end{botherref}
\endbibitem

%%% 54
\bibitem[\protect\citeauthoryear{Ye et~al.}{2023}]{ye2023prompt}
\begin{botherref}
\oauthor{\bsnm{Ye}, \binits{Q.}},
\oauthor{\bsnm{Axmed}, \binits{M.}},
\oauthor{\bsnm{Pryzant}, \binits{R.}},
\oauthor{\bsnm{Khani}, \binits{F.}}:
Prompt engineering a prompt engineer.
arXiv preprint arXiv:2311.05661
(2023)
\end{botherref}
\endbibitem

%%% 55
\bibitem[\protect\citeauthoryear{Lu et~al.}{2023}]{lu2023strings}
\begin{botherref}
\oauthor{\bsnm{Lu}, \binits{Y.}},
\oauthor{\bsnm{Wang}, \binits{J.}},
\oauthor{\bsnm{Tang}, \binits{R.}},
\oauthor{\bsnm{Riedel}, \binits{S.}},
\oauthor{\bsnm{Stenetorp}, \binits{P.}}:
Strings from the library of babel: Random sampling as a strong baseline for prompt optimisation.
arXiv preprint arXiv:2311.09569
(2023)
\end{botherref}
\endbibitem

%%% 56
\bibitem[\protect\citeauthoryear{Cheng et~al.}{2023}]{cheng2023black}
\begin{botherref}
\oauthor{\bsnm{Cheng}, \binits{J.}},
\oauthor{\bsnm{Liu}, \binits{X.}},
\oauthor{\bsnm{Zheng}, \binits{K.}},
\oauthor{\bsnm{Ke}, \binits{P.}},
\oauthor{\bsnm{Wang}, \binits{H.}},
\oauthor{\bsnm{Dong}, \binits{Y.}},
\oauthor{\bsnm{Tang}, \binits{J.}},
\oauthor{\bsnm{Huang}, \binits{M.}}:
Black-box prompt optimization: Aligning large language models without model training.
arXiv preprint arXiv:2311.04155
(2023)
\end{botherref}
\endbibitem

%%% 57
\bibitem[\protect\citeauthoryear{Sabbatella et~al.}{2024}]{sabbatella2024prompt}
\begin{barticle}
\bauthor{\bsnm{Sabbatella}, \binits{A.}},
\bauthor{\bsnm{Ponti}, \binits{A.}},
\bauthor{\bsnm{Giordani}, \binits{I.}},
\bauthor{\bsnm{Candelieri}, \binits{A.}},
\bauthor{\bsnm{Archetti}, \binits{F.}}:
\batitle{Prompt optimization in large language models}.
\bjtitle{Mathematics}
\bvolume{12}(\bissue{6}),
\bfpage{929}
(\byear{2024})
\end{barticle}
\endbibitem

%%% 58
\bibitem[\protect\citeauthoryear{Chen et~al.}{2023}]{chen2023instructzero}
\begin{botherref}
\oauthor{\bsnm{Chen}, \binits{L.}},
\oauthor{\bsnm{Chen}, \binits{J.}},
\oauthor{\bsnm{Goldstein}, \binits{T.}},
\oauthor{\bsnm{Huang}, \binits{H.}},
\oauthor{\bsnm{Zhou}, \binits{T.}}:
Instructzero: Efficient instruction optimization for black-box large language models.
arXiv preprint arXiv:2306.03082
(2023)
\end{botherref}
\endbibitem

%%% 59
\bibitem[\protect\citeauthoryear{Zhou et~al.}{2022}]{zhou2022large}
\begin{bchapter}
\bauthor{\bsnm{Zhou}, \binits{Y.}},
\bauthor{\bsnm{Muresanu}, \binits{A.I.}},
\bauthor{\bsnm{Han}, \binits{Z.}},
\bauthor{\bsnm{Paster}, \binits{K.}},
\bauthor{\bsnm{Pitis}, \binits{S.}},
\bauthor{\bsnm{Chan}, \binits{H.}},
\bauthor{\bsnm{Ba}, \binits{J.}}:
\bctitle{Large language models are human-level prompt engineers}.
In: \bbtitle{The Eleventh International Conference on Learning Representations}
(\byear{2022})
\end{bchapter}
\endbibitem

%%% 60
\bibitem[\protect\citeauthoryear{Ma et~al.}{2024}]{ma2024large}
\begin{botherref}
\oauthor{\bsnm{Ma}, \binits{R.}},
\oauthor{\bsnm{Wang}, \binits{X.}},
\oauthor{\bsnm{Zhou}, \binits{X.}},
\oauthor{\bsnm{Li}, \binits{J.}},
\oauthor{\bsnm{Du}, \binits{N.}},
\oauthor{\bsnm{Gui}, \binits{T.}},
\oauthor{\bsnm{Zhang}, \binits{Q.}},
\oauthor{\bsnm{Huang}, \binits{X.}}:
Are large language models good prompt optimizers?
arXiv preprint arXiv:2402.02101
(2024)
\end{botherref}
\endbibitem

%%% 61
\bibitem[\protect\citeauthoryear{Wang}{2017}]{wang2017liar}
\begin{botherref}
\oauthor{\bsnm{Wang}, \binits{W.Y.}}:
" liar, liar pants on fire": A new benchmark dataset for fake news detection.
arXiv preprint arXiv:1705.00648
(2017)
\end{botherref}
\endbibitem

%%% 62
\bibitem[\protect\citeauthoryear{Levesque et~al.}{2012}]{levesque2012winograd}
\begin{bchapter}
\bauthor{\bsnm{Levesque}, \binits{H.}},
\bauthor{\bsnm{Davis}, \binits{E.}},
\bauthor{\bsnm{Morgenstern}, \binits{L.}}:
\bctitle{The winograd schema challenge}.
In: \bbtitle{Thirteenth International Conference on the Principles of Knowledge Representation and Reasoning}
(\byear{2012})
\end{bchapter}
\endbibitem

%%% 63
\bibitem[\protect\citeauthoryear{Sakaguchi et~al.}{2021}]{sakaguchi2021winogrande}
\begin{barticle}
\bauthor{\bsnm{Sakaguchi}, \binits{K.}},
\bauthor{\bsnm{Bras}, \binits{R.L.}},
\bauthor{\bsnm{Bhagavatula}, \binits{C.}},
\bauthor{\bsnm{Choi}, \binits{Y.}}:
\batitle{Winogrande: An adversarial winograd schema challenge at scale}.
\bjtitle{Communications of the ACM}
\bvolume{64}(\bissue{9}),
\bfpage{99}--\blpage{106}
(\byear{2021})
\end{barticle}
\endbibitem

%%% 64
\bibitem[\protect\citeauthoryear{Suzgun et~al.}{2022}]{suzgun2022bigbenchhard}
\begin{botherref}
\oauthor{\bsnm{Suzgun}, \binits{M.}}, et al.:
BIG-Bench Hard (BBH).
Accessed: 2025-02-11
(2022).
\url{https://github.com/suzgunmirac/BIG-Bench-Hard}
\end{botherref}
\endbibitem

%%% 65
\bibitem[\protect\citeauthoryear{Mollas et~al.}{2022}]{mollas2022ethos}
\begin{barticle}
\bauthor{\bsnm{Mollas}, \binits{I.}},
\bauthor{\bsnm{Chrysopoulou}, \binits{Z.}},
\bauthor{\bsnm{Karlos}, \binits{S.}},
\bauthor{\bsnm{Tsoumakas}, \binits{G.}}:
\batitle{Ethos: a multi-label hate speech detection dataset}.
\bjtitle{Complex \& Intelligent Systems}
\bvolume{8}(\bissue{6}),
\bfpage{4663}--\blpage{4678}
(\byear{2022})
\end{barticle}
\endbibitem

%%% 66
\bibitem[\protect\citeauthoryear{Pang and Lee}{2004}]{pang2004sentimental}
\begin{botherref}
\oauthor{\bsnm{Pang}, \binits{B.}},
\oauthor{\bsnm{Lee}, \binits{L.}}:
A sentimental education: Sentiment analysis using subjectivity summarization based on minimum cuts.
arXiv preprint cs/0409058
(2004)
\end{botherref}
\endbibitem

%%% 67
\bibitem[\protect\citeauthoryear{Farha and Magdy}{2020}]{farha2020arabic}
\begin{bchapter}
\bauthor{\bsnm{Farha}, \binits{I.A.}},
\bauthor{\bsnm{Magdy}, \binits{W.}}:
\bctitle{From arabic sentiment analysis to sarcasm detection: The arsarcasm dataset}.
In: \bbtitle{Proceedings of the 4th Workshop on Open-Source Arabic Corpora and Processing Tools, with a Shared Task on Offensive Language Detection},
pp. \bfpage{32}--\blpage{39}
(\byear{2020})
\end{bchapter}
\endbibitem

%%% 68
\bibitem[\protect\citeauthoryear{Kiesel et~al.}{2019}]{kiesel2019semeval}
\begin{bchapter}
\bauthor{\bsnm{Kiesel}, \binits{J.}},
\bauthor{\bsnm{Mestre}, \binits{M.}},
\bauthor{\bsnm{Shukla}, \binits{R.}},
\bauthor{\bsnm{Vincent}, \binits{E.}},
\bauthor{\bsnm{Adineh}, \binits{P.}},
\bauthor{\bsnm{Corney}, \binits{D.}},
\bauthor{\bsnm{Stein}, \binits{B.}},
\bauthor{\bsnm{Potthast}, \binits{M.}}:
\bctitle{Semeval-2019 task 4: Hyperpartisan news detection}.
In: \bbtitle{Proceedings of the 13th International Workshop on Semantic Evaluation},
pp. \bfpage{829}--\blpage{839}
(\byear{2019})
\end{bchapter}
\endbibitem

%%% 69
\bibitem[\protect\citeauthoryear{Warstadt}{2019}]{warstadt2019neural}
\begin{botherref}
\oauthor{\bsnm{Warstadt}, \binits{A.}}:
Neural network acceptability judgments.
arXiv preprint arXiv:1805.12471
(2019)
\end{botherref}
\endbibitem

%%% 70
\bibitem[\protect\citeauthoryear{Pilehvar and Camacho-Collados}{2018}]{pilehvar2018wic}
\begin{botherref}
\oauthor{\bsnm{Pilehvar}, \binits{M.T.}},
\oauthor{\bsnm{Camacho-Collados}, \binits{J.}}:
Wic: the word-in-context dataset for evaluating context-sensitive meaning representations.
arXiv preprint arXiv:1808.09121
(2018)
\end{botherref}
\endbibitem

%%% 71
\bibitem[\protect\citeauthoryear{Honovich et~al.}{2022}]{honovich2022instruction}
\begin{botherref}
\oauthor{\bsnm{Honovich}, \binits{O.}},
\oauthor{\bsnm{Shaham}, \binits{U.}},
\oauthor{\bsnm{Bowman}, \binits{S.R.}},
\oauthor{\bsnm{Levy}, \binits{O.}}:
Instruction induction: From few examples to natural language task descriptions.
arXiv preprint arXiv:2205.10782
(2022)
\end{botherref}
\endbibitem

%%% 72
\bibitem[\protect\citeauthoryear{Dernoncourt and Lee}{2017}]{dernoncourt2017pubmed}
\begin{botherref}
\oauthor{\bsnm{Dernoncourt}, \binits{F.}},
\oauthor{\bsnm{Lee}, \binits{J.Y.}}:
Pubmed 200k rct: a dataset for sequential sentence classification in medical abstracts.
arXiv preprint arXiv:1710.06071
(2017)
\end{botherref}
\endbibitem

%%% 73
\bibitem[\protect\citeauthoryear{Zhang et~al.}{2015}]{zhang2015character}
\begin{botherref}
\oauthor{\bsnm{Zhang}, \binits{X.}},
\oauthor{\bsnm{Zhao}, \binits{J.}},
\oauthor{\bsnm{LeCun}, \binits{Y.}}:
Character-level convolutional networks for text classification.
Advances in neural information processing systems
\textbf{28}
(2015)
\end{botherref}
\endbibitem

%%% 74
\bibitem[\protect\citeauthoryear{Voorhees and Tice}{2000}]{voorhees2000building}
\begin{bchapter}
\bauthor{\bsnm{Voorhees}, \binits{E.M.}},
\bauthor{\bsnm{Tice}, \binits{D.M.}}:
\bctitle{Building a question answering test collection}.
In: \bbtitle{Proceedings of the 23rd Annual International ACM SIGIR Conference on Research and Development in Information Retrieval},
pp. \bfpage{200}--\blpage{207}
(\byear{2000})
\end{bchapter}
\endbibitem

%%% 75
\bibitem[\protect\citeauthoryear{Lehmann et~al.}{2015}]{lehmann2015dbpedia}
\begin{barticle}
\bauthor{\bsnm{Lehmann}, \binits{J.}},
\bauthor{\bsnm{Isele}, \binits{R.}},
\bauthor{\bsnm{Jakob}, \binits{M.}},
\bauthor{\bsnm{Jentzsch}, \binits{A.}},
\bauthor{\bsnm{Kontokostas}, \binits{D.}},
\bauthor{\bsnm{Mendes}, \binits{P.N.}},
\bauthor{\bsnm{Hellmann}, \binits{S.}},
\bauthor{\bsnm{Morsey}, \binits{M.}},
\bauthor{\bsnm{Van~Kleef}, \binits{P.}},
\bauthor{\bsnm{Auer}, \binits{S.}}, \betal:
\batitle{Dbpedia--a large-scale, multilingual knowledge base extracted from wikipedia}.
\bjtitle{Semantic web}
\bvolume{6}(\bissue{2}),
\bfpage{167}--\blpage{195}
(\byear{2015})
\end{barticle}
\endbibitem

%%% 76
\bibitem[\protect\citeauthoryear{Socher et~al.}{2013}]{socher2013recursive}
\begin{bchapter}
\bauthor{\bsnm{Socher}, \binits{R.}},
\bauthor{\bsnm{Perelygin}, \binits{A.}},
\bauthor{\bsnm{Wu}, \binits{J.}},
\bauthor{\bsnm{Chuang}, \binits{J.}},
\bauthor{\bsnm{Manning}, \binits{C.D.}},
\bauthor{\bsnm{Ng}, \binits{A.Y.}},
\bauthor{\bsnm{Potts}, \binits{C.}}:
\bctitle{Recursive deep models for semantic compositionality over a sentiment treebank}.
In: \bbtitle{Proceedings of the 2013 Conference on Empirical Methods in Natural Language Processing},
pp. \bfpage{1631}--\blpage{1642}
(\byear{2013})
\end{bchapter}
\endbibitem

%%% 77
\bibitem[\protect\citeauthoryear{Pang and Lee}{2005}]{pang2005seeing}
\begin{botherref}
\oauthor{\bsnm{Pang}, \binits{B.}},
\oauthor{\bsnm{Lee}, \binits{L.}}:
Seeing stars: Exploiting class relationships for sentiment categorization with respect to rating scales.
arXiv preprint cs/0506075
(2005)
\end{botherref}
\endbibitem

%%% 78
\bibitem[\protect\citeauthoryear{Hu and Liu}{2004}]{hu2004mining}
\begin{bchapter}
\bauthor{\bsnm{Hu}, \binits{M.}},
\bauthor{\bsnm{Liu}, \binits{B.}}:
\bctitle{Mining and summarizing customer reviews}.
In: \bbtitle{Proceedings of the Tenth ACM SIGKDD International Conference on Knowledge Discovery and Data Mining},
pp. \bfpage{168}--\blpage{177}
(\byear{2004})
\end{bchapter}
\endbibitem

%%% 79
\bibitem[\protect\citeauthoryear{McAuley and Leskovec}{2013}]{mcauley2013hidden}
\begin{bchapter}
\bauthor{\bsnm{McAuley}, \binits{J.}},
\bauthor{\bsnm{Leskovec}, \binits{J.}}:
\bctitle{Hidden factors and hidden topics: understanding rating dimensions with review text}.
In: \bbtitle{Proceedings of the 7th ACM Conference on Recommender Systems},
pp. \bfpage{165}--\blpage{172}
(\byear{2013})
\end{bchapter}
\endbibitem

%%% 80
\bibitem[\protect\citeauthoryear{Zhang et~al.}{2018}]{zhang2018record}
\begin{botherref}
\oauthor{\bsnm{Zhang}, \binits{S.}},
\oauthor{\bsnm{Liu}, \binits{X.}},
\oauthor{\bsnm{Liu}, \binits{J.}},
\oauthor{\bsnm{Gao}, \binits{J.}},
\oauthor{\bsnm{Duh}, \binits{K.}},
\oauthor{\bsnm{Van~Durme}, \binits{B.}}:
Record: Bridging the gap between human and machine commonsense reading comprehension.
arXiv preprint arXiv:1810.12885
(2018)
\end{botherref}
\endbibitem

%%% 81
\bibitem[\protect\citeauthoryear{Rajpurkar}{2016}]{rajpurkar2016squad}
\begin{botherref}
\oauthor{\bsnm{Rajpurkar}, \binits{P.}}:
Squad: 100,000+ questions for machine comprehension of text.
arXiv preprint arXiv:1606.05250
(2016)
\end{botherref}
\endbibitem

%%% 82
\bibitem[\protect\citeauthoryear{Kwiatkowski et~al.}{2019}]{kwiatkowski2019natural}
\begin{barticle}
\bauthor{\bsnm{Kwiatkowski}, \binits{T.}},
\bauthor{\bsnm{Palomaki}, \binits{J.}},
\bauthor{\bsnm{Redfield}, \binits{O.}},
\bauthor{\bsnm{Collins}, \binits{M.}},
\bauthor{\bsnm{Parikh}, \binits{A.}},
\bauthor{\bsnm{Alberti}, \binits{C.}},
\bauthor{\bsnm{Epstein}, \binits{D.}},
\bauthor{\bsnm{Polosukhin}, \binits{I.}},
\bauthor{\bsnm{Devlin}, \binits{J.}},
\bauthor{\bsnm{Lee}, \binits{K.}}, \betal:
\batitle{Natural questions: a benchmark for question answering research}.
\bjtitle{Transactions of the Association for Computational Linguistics}
\bvolume{7},
\bfpage{453}--\blpage{466}
(\byear{2019})
\end{barticle}
\endbibitem

%%% 83
\bibitem[\protect\citeauthoryear{Yang et~al.}{2018}]{yang2018hotpotqa}
\begin{botherref}
\oauthor{\bsnm{Yang}, \binits{Z.}},
\oauthor{\bsnm{Qi}, \binits{P.}},
\oauthor{\bsnm{Zhang}, \binits{S.}},
\oauthor{\bsnm{Bengio}, \binits{Y.}},
\oauthor{\bsnm{Cohen}, \binits{W.W.}},
\oauthor{\bsnm{Salakhutdinov}, \binits{R.}},
\oauthor{\bsnm{Manning}, \binits{C.D.}}:
Hotpotqa: A dataset for diverse, explainable multi-hop question answering.
arXiv preprint arXiv:1809.09600
(2018)
\end{botherref}
\endbibitem

%%% 84
\bibitem[\protect\citeauthoryear{Dunn et~al.}{2017}]{dunn2017searchqa}
\begin{botherref}
\oauthor{\bsnm{Dunn}, \binits{M.}},
\oauthor{\bsnm{Sagun}, \binits{L.}},
\oauthor{\bsnm{Higgins}, \binits{M.}},
\oauthor{\bsnm{Guney}, \binits{V.U.}},
\oauthor{\bsnm{Cirik}, \binits{V.}},
\oauthor{\bsnm{Cho}, \binits{K.}}:
Searchqa: A new q\&a dataset augmented with context from a search engine. arxiv.
arXiv preprint cs.CL/1704.05179
(2017)
\end{botherref}
\endbibitem

%%% 85
\bibitem[\protect\citeauthoryear{Trischler et~al.}{2016}]{trischler2016newsqa}
\begin{botherref}
\oauthor{\bsnm{Trischler}, \binits{A.}},
\oauthor{\bsnm{Wang}, \binits{T.}},
\oauthor{\bsnm{Yuan}, \binits{X.}},
\oauthor{\bsnm{Harris}, \binits{J.}},
\oauthor{\bsnm{Sordoni}, \binits{A.}},
\oauthor{\bsnm{Bachman}, \binits{P.}},
\oauthor{\bsnm{Suleman}, \binits{K.}}:
Newsqa: A machine comprehension dataset.
arXiv preprint arXiv:1611.09830
(2016)
\end{botherref}
\endbibitem

%%% 86
\bibitem[\protect\citeauthoryear{Clark et~al.}{2019}]{clark2019boolq}
\begin{botherref}
\oauthor{\bsnm{Clark}, \binits{C.}},
\oauthor{\bsnm{Lee}, \binits{K.}},
\oauthor{\bsnm{Chang}, \binits{M.-W.}},
\oauthor{\bsnm{Kwiatkowski}, \binits{T.}},
\oauthor{\bsnm{Collins}, \binits{M.}},
\oauthor{\bsnm{Toutanova}, \binits{K.}}:
Boolq: Exploring the surprising difficulty of natural yes/no questions.
arXiv preprint arXiv:1905.10044
(2019)
\end{botherref}
\endbibitem

%%% 87
\bibitem[\protect\citeauthoryear{Roemmele et~al.}{2011}]{roemmele2011choice}
\begin{bchapter}
\bauthor{\bsnm{Roemmele}, \binits{M.}},
\bauthor{\bsnm{Bejan}, \binits{C.A.}},
\bauthor{\bsnm{Gordon}, \binits{A.S.}}:
\bctitle{Choice of plausible alternatives: An evaluation of commonsense causal reasoning}.
In: \bbtitle{2011 AAAI Spring Symposium Series}
(\byear{2011})
\end{bchapter}
\endbibitem

%%% 88
\bibitem[\protect\citeauthoryear{Khashabi et~al.}{2018}]{khashabi2018looking}
\begin{bchapter}
\bauthor{\bsnm{Khashabi}, \binits{D.}},
\bauthor{\bsnm{Chaturvedi}, \binits{S.}},
\bauthor{\bsnm{Roth}, \binits{M.}},
\bauthor{\bsnm{Upadhyay}, \binits{S.}},
\bauthor{\bsnm{Roth}, \binits{D.}}:
\bctitle{Looking beyond the surface: A challenge set for reading comprehension over multiple sentences}.
In: \bbtitle{Proceedings of the 2018 Conference of the North American Chapter of the Association for Computational Linguistics: Human Language Technologies, Volume 1 (Long Papers)},
pp. \bfpage{252}--\blpage{262}
(\byear{2018})
\end{bchapter}
\endbibitem

%%% 89
\bibitem[\protect\citeauthoryear{Saha et~al.}{2018}]{saha2018duorc}
\begin{botherref}
\oauthor{\bsnm{Saha}, \binits{A.}},
\oauthor{\bsnm{Aralikatte}, \binits{R.}},
\oauthor{\bsnm{Khapra}, \binits{M.M.}},
\oauthor{\bsnm{Sankaranarayanan}, \binits{K.}}:
Duorc: Towards complex language understanding with paraphrased reading comprehension.
arXiv preprint arXiv:1804.07927
(2018)
\end{botherref}
\endbibitem

%%% 90
\bibitem[\protect\citeauthoryear{Dua et~al.}{2019}]{dua2019drop}
\begin{botherref}
\oauthor{\bsnm{Dua}, \binits{D.}},
\oauthor{\bsnm{Wang}, \binits{Y.}},
\oauthor{\bsnm{Dasigi}, \binits{P.}},
\oauthor{\bsnm{Stanovsky}, \binits{G.}},
\oauthor{\bsnm{Singh}, \binits{S.}},
\oauthor{\bsnm{Gardner}, \binits{M.}}:
Drop: A reading comprehension benchmark requiring discrete reasoning over paragraphs.
arXiv preprint arXiv:1903.00161
(2019)
\end{botherref}
\endbibitem

%%% 91
\bibitem[\protect\citeauthoryear{Kembhavi et~al.}{2017}]{kembhavi2017you}
\begin{bchapter}
\bauthor{\bsnm{Kembhavi}, \binits{A.}},
\bauthor{\bsnm{Seo}, \binits{M.}},
\bauthor{\bsnm{Schwenk}, \binits{D.}},
\bauthor{\bsnm{Choi}, \binits{J.}},
\bauthor{\bsnm{Farhadi}, \binits{A.}},
\bauthor{\bsnm{Hajishirzi}, \binits{H.}}:
\bctitle{Are you smarter than a sixth grader? textbook question answering for multimodal machine comprehension}.
In: \bbtitle{Proceedings of the IEEE Conference on Computer Vision and Pattern Recognition},
pp. \bfpage{4999}--\blpage{5007}
(\byear{2017})
\end{bchapter}
\endbibitem

%%% 92
\bibitem[\protect\citeauthoryear{}{}]{BioASQ}
\begin{botherref}
BioASQ: A Challenge on Large-Scale Biomedical Semantic Indexing and Question Answering.
\url{https://www.bioasq.org/}.
Accessed: 2025-02-10
\end{botherref}
\endbibitem

%%% 93
\bibitem[\protect\citeauthoryear{Lai et~al.}{2017}]{lai2017race}
\begin{botherref}
\oauthor{\bsnm{Lai}, \binits{G.}},
\oauthor{\bsnm{Xie}, \binits{Q.}},
\oauthor{\bsnm{Liu}, \binits{H.}},
\oauthor{\bsnm{Yang}, \binits{Y.}},
\oauthor{\bsnm{Hovy}, \binits{E.}}:
Race: Large-scale reading comprehension dataset from examinations.
arXiv preprint arXiv:1704.04683
(2017)
\end{botherref}
\endbibitem

%%% 94
\bibitem[\protect\citeauthoryear{Mihaylov et~al.}{2018}]{mihaylov2018can}
\begin{botherref}
\oauthor{\bsnm{Mihaylov}, \binits{T.}},
\oauthor{\bsnm{Clark}, \binits{P.}},
\oauthor{\bsnm{Khot}, \binits{T.}},
\oauthor{\bsnm{Sabharwal}, \binits{A.}}:
Can a suit of armor conduct electricity? a new dataset for open book question answering.
arXiv preprint arXiv:1809.02789
(2018)
\end{botherref}
\endbibitem

%%% 95
\bibitem[\protect\citeauthoryear{Lin et~al.}{2021}]{lin2021truthfulqa}
\begin{botherref}
\oauthor{\bsnm{Lin}, \binits{S.}},
\oauthor{\bsnm{Hilton}, \binits{J.}},
\oauthor{\bsnm{Evans}, \binits{O.}}:
Truthfulqa: Measuring how models mimic human falsehoods.
arXiv preprint arXiv:2109.07958
(2021)
\end{botherref}
\endbibitem

%%% 96
\bibitem[\protect\citeauthoryear{Jin et~al.}{2021}]{jin2021disease}
\begin{barticle}
\bauthor{\bsnm{Jin}, \binits{D.}},
\bauthor{\bsnm{Pan}, \binits{E.}},
\bauthor{\bsnm{Oufattole}, \binits{N.}},
\bauthor{\bsnm{Weng}, \binits{W.-H.}},
\bauthor{\bsnm{Fang}, \binits{H.}},
\bauthor{\bsnm{Szolovits}, \binits{P.}}:
\batitle{What disease does this patient have? a large-scale open domain question answering dataset from medical exams}.
\bjtitle{Applied Sciences}
\bvolume{11}(\bissue{14}),
\bfpage{6421}
(\byear{2021})
\end{barticle}
\endbibitem

%%% 97
\bibitem[\protect\citeauthoryear{Williams et~al.}{2017}]{williams2017broad}
\begin{botherref}
\oauthor{\bsnm{Williams}, \binits{A.}},
\oauthor{\bsnm{Nangia}, \binits{N.}},
\oauthor{\bsnm{Bowman}, \binits{S.R.}}:
A broad-coverage challenge corpus for sentence understanding through inference.
arXiv preprint arXiv:1704.05426
(2017)
\end{botherref}
\endbibitem

%%% 98
\bibitem[\protect\citeauthoryear{Wang et~al.}{2018}]{wang2018glue}
\begin{botherref}
\oauthor{\bsnm{Wang}, \binits{A.}},
\oauthor{\bsnm{Singh}, \binits{A.}},
\oauthor{\bsnm{Michael}, \binits{J.}},
\oauthor{\bsnm{Hill}, \binits{F.}},
\oauthor{\bsnm{Levy}, \binits{O.}},
\oauthor{\bsnm{Bowman}, \binits{S.}}:
Glue: A multi-task benchmark and analysis platform for natural language understanding. arXiv preprint arXiv: 180407461
(2018)
\end{botherref}
\endbibitem

%%% 99
\bibitem[\protect\citeauthoryear{Giampiccolo et~al.}{2007}]{giampiccolo2007third}
\begin{bchapter}
\bauthor{\bsnm{Giampiccolo}, \binits{D.}},
\bauthor{\bsnm{Magnini}, \binits{B.}},
\bauthor{\bsnm{Dagan}, \binits{I.}},
\bauthor{\bsnm{Dolan}, \binits{W.B.}}:
\bctitle{The third pascal recognizing textual entailment challenge}.
In: \bbtitle{Proceedings of the ACL-PASCAL Workshop on Textual Entailment and Paraphrasing},
pp. \bfpage{1}--\blpage{9}
(\byear{2007})
\end{bchapter}
\endbibitem

%%% 100
\bibitem[\protect\citeauthoryear{Blunsom et~al.}{2018}]{blunsom2018snli}
\begin{botherref}
\oauthor{\bsnm{Blunsom}, \binits{P.}},
\oauthor{\bsnm{Camburu}, \binits{O.-M.}},
\oauthor{\bsnm{Lukasiewicz}, \binits{T.}},
\oauthor{\bsnm{Rockt{\"a}schel}, \binits{T.}}:
e- snli: Natural language inference with natural language explanations
(2018)
\end{botherref}
\endbibitem

%%% 101
\bibitem[\protect\citeauthoryear{De~Marneffe et~al.}{2019}]{de2019commitmentbank}
\begin{bchapter}
\bauthor{\bsnm{De~Marneffe}, \binits{M.-C.}},
\bauthor{\bsnm{Simons}, \binits{M.}},
\bauthor{\bsnm{Tonhauser}, \binits{J.}}:
\bctitle{The commitmentbank: Investigating projection in naturally occurring discourse}.
In: \bbtitle{Proceedings of Sinn und Bedeutung},
vol. \bseriesno{23},
pp. \bfpage{107}--\blpage{124}
(\byear{2019})
\end{bchapter}
\endbibitem

%%% 102
\bibitem[\protect\citeauthoryear{Khot et~al.}{2018}]{khot2018scitail}
\begin{bchapter}
\bauthor{\bsnm{Khot}, \binits{T.}},
\bauthor{\bsnm{Sabharwal}, \binits{A.}},
\bauthor{\bsnm{Clark}, \binits{P.}}:
\bctitle{Scitail: A textual entailment dataset from science question answering}.
In: \bbtitle{Proceedings of the AAAI Conference on Artificial Intelligence},
vol. \bseriesno{32}
(\byear{2018})
\end{bchapter}
\endbibitem

%%% 103
\bibitem[\protect\citeauthoryear{Bowman et~al.}{2015}]{bowman2015large}
\begin{botherref}
\oauthor{\bsnm{Bowman}, \binits{S.R.}},
\oauthor{\bsnm{Angeli}, \binits{G.}},
\oauthor{\bsnm{Potts}, \binits{C.}},
\oauthor{\bsnm{Manning}, \binits{C.D.}}:
A large annotated corpus for learning natural language inference.
arXiv preprint arXiv:1508.05326
(2015)
\end{botherref}
\endbibitem

%%% 104
\bibitem[\protect\citeauthoryear{Marelli et~al.}{2014}]{marelli2014semeval}
\begin{bchapter}
\bauthor{\bsnm{Marelli}, \binits{M.}},
\bauthor{\bsnm{Bentivogli}, \binits{L.}},
\bauthor{\bsnm{Baroni}, \binits{M.}},
\bauthor{\bsnm{Bernardi}, \binits{R.}},
\bauthor{\bsnm{Menini}, \binits{S.}},
\bauthor{\bsnm{Zamparelli}, \binits{R.}}:
\bctitle{Semeval-2014 task 1: Evaluation of compositional distributional semantic models on full sentences through semantic relatedness and textual entailment}.
In: \bbtitle{Proceedings of the 8th International Workshop on Semantic Evaluation (SemEval 2014)},
pp. \bfpage{1}--\blpage{8}
(\byear{2014})
\end{bchapter}
\endbibitem

%%% 105
\bibitem[\protect\citeauthoryear{Conneau et~al.}{2018}]{conneau2018xnli}
\begin{botherref}
\oauthor{\bsnm{Conneau}, \binits{A.}},
\oauthor{\bsnm{Lample}, \binits{G.}},
\oauthor{\bsnm{Rinott}, \binits{R.}},
\oauthor{\bsnm{Williams}, \binits{A.}},
\oauthor{\bsnm{Bowman}, \binits{S.R.}},
\oauthor{\bsnm{Schwenk}, \binits{H.}},
\oauthor{\bsnm{Stoyanov}, \binits{V.}}:
Xnli: Evaluating cross-lingual sentence representations.
arXiv preprint arXiv:1809.05053
(2018)
\end{botherref}
\endbibitem

%%% 106
\bibitem[\protect\citeauthoryear{Gliwa et~al.}{2019}]{gliwa2019samsum}
\begin{botherref}
\oauthor{\bsnm{Gliwa}, \binits{B.}},
\oauthor{\bsnm{Mochol}, \binits{I.}},
\oauthor{\bsnm{Biesek}, \binits{M.}},
\oauthor{\bsnm{Wawer}, \binits{A.}}:
Samsum corpus: A human-annotated dialogue dataset for abstractive summarization.
arXiv preprint arXiv:1911.12237
(2019)
\end{botherref}
\endbibitem

%%% 107
\bibitem[\protect\citeauthoryear{Narayan et~al.}{2018}]{narayan2018don}
\begin{botherref}
\oauthor{\bsnm{Narayan}, \binits{S.}},
\oauthor{\bsnm{Cohen}, \binits{S.B.}},
\oauthor{\bsnm{Lapata}, \binits{M.}}:
Don't give me the details, just the summary! topic-aware convolutional neural networks for extreme summarization.
arXiv preprint arXiv:1808.08745
(2018)
\end{botherref}
\endbibitem

%%% 108
\bibitem[\protect\citeauthoryear{Nallapati et~al.}{2016}]{nallapati2016abstractive}
\begin{botherref}
\oauthor{\bsnm{Nallapati}, \binits{R.}},
\oauthor{\bsnm{Zhou}, \binits{B.}},
\oauthor{\bsnm{Gulcehre}, \binits{C.}},
\oauthor{\bsnm{Xiang}, \binits{B.}}, et al.:
Abstractive text summarization using sequence-to-sequence rnns and beyond.
arXiv preprint arXiv:1602.06023
(2016)
\end{botherref}
\endbibitem

%%% 109
\bibitem[\protect\citeauthoryear{Novikova et~al.}{2017}]{novikova2017e2e}
\begin{botherref}
\oauthor{\bsnm{Novikova}, \binits{J.}},
\oauthor{\bsnm{Du{\v{s}}ek}, \binits{O.}},
\oauthor{\bsnm{Rieser}, \binits{V.}}:
The e2e dataset: New challenges for end-to-end generation.
arXiv preprint arXiv:1706.09254
(2017)
\end{botherref}
\endbibitem

%%% 110
\bibitem[\protect\citeauthoryear{Gardent et~al.}{2017}]{gardent2017webnlg}
\begin{bchapter}
\bauthor{\bsnm{Gardent}, \binits{C.}},
\bauthor{\bsnm{Shimorina}, \binits{A.}},
\bauthor{\bsnm{Narayan}, \binits{S.}},
\bauthor{\bsnm{Perez-Beltrachini}, \binits{L.}}:
\bctitle{The webnlg challenge: Generating text from rdf data}.
In: \bbtitle{10th International Conference on Natural Language Generation},
pp. \bfpage{124}--\blpage{133}
(\byear{2017}).
\bcomment{ACL Anthology}
\end{bchapter}
\endbibitem

%%% 111
\bibitem[\protect\citeauthoryear{Nan et~al.}{2020}]{nan2020dart}
\begin{botherref}
\oauthor{\bsnm{Nan}, \binits{L.}},
\oauthor{\bsnm{Radev}, \binits{D.}},
\oauthor{\bsnm{Zhang}, \binits{R.}},
\oauthor{\bsnm{Rau}, \binits{A.}},
\oauthor{\bsnm{Sivaprasad}, \binits{A.}},
\oauthor{\bsnm{Hsieh}, \binits{C.}},
\oauthor{\bsnm{Tang}, \binits{X.}},
\oauthor{\bsnm{Vyas}, \binits{A.}},
\oauthor{\bsnm{Verma}, \binits{N.}},
\oauthor{\bsnm{Krishna}, \binits{P.}}, et al.:
Dart: Open-domain structured data record to text generation.
arXiv preprint arXiv:2007.02871
(2020)
\end{botherref}
\endbibitem

%%% 112
\bibitem[\protect\citeauthoryear{Alva-Manchego et~al.}{2020}]{alva2020asset}
\begin{botherref}
\oauthor{\bsnm{Alva-Manchego}, \binits{F.}},
\oauthor{\bsnm{Martin}, \binits{L.}},
\oauthor{\bsnm{Bordes}, \binits{A.}},
\oauthor{\bsnm{Scarton}, \binits{C.}},
\oauthor{\bsnm{Sagot}, \binits{B.}},
\oauthor{\bsnm{Specia}, \binits{L.}}:
Asset: A dataset for tuning and evaluation of sentence simplification models with multiple rewriting transformations.
arXiv preprint arXiv:2005.00481
(2020)
\end{botherref}
\endbibitem

%%% 113
\bibitem[\protect\citeauthoryear{Shen et~al.}{2017}]{shen2017style}
\begin{botherref}
\oauthor{\bsnm{Shen}, \binits{T.}},
\oauthor{\bsnm{Lei}, \binits{T.}},
\oauthor{\bsnm{Barzilay}, \binits{R.}},
\oauthor{\bsnm{Jaakkola}, \binits{T.}}:
Style transfer from non-parallel text by cross-alignment.
Advances in neural information processing systems
\textbf{30}
(2017)
\end{botherref}
\endbibitem

%%% 114
\bibitem[\protect\citeauthoryear{Xu et~al.}{2012}]{xu2012paraphrasing}
\begin{bchapter}
\bauthor{\bsnm{Xu}, \binits{W.}},
\bauthor{\bsnm{Ritter}, \binits{A.}},
\bauthor{\bsnm{Dolan}, \binits{W.B.}},
\bauthor{\bsnm{Grishman}, \binits{R.}},
\bauthor{\bsnm{Cherry}, \binits{C.}}:
\bctitle{Paraphrasing for style}.
In: \bbtitle{Proceedings of COLING 2012},
pp. \bfpage{2899}--\blpage{2914}
(\byear{2012})
\end{bchapter}
\endbibitem

%%% 115
\bibitem[\protect\citeauthoryear{Dumitrescu et~al.}{2021}]{dumitrescu2021liro}
\begin{bchapter}
\bauthor{\bsnm{Dumitrescu}, \binits{S.D.}},
\bauthor{\bsnm{Rebeja}, \binits{P.}},
\bauthor{\bsnm{Lorincz}, \binits{B.}},
\bauthor{\bsnm{Gaman}, \binits{M.}},
\bauthor{\bsnm{Avram}, \binits{A.}},
\bauthor{\bsnm{Ilie}, \binits{M.}},
\bauthor{\bsnm{Pruteanu}, \binits{A.}},
\bauthor{\bsnm{Stan}, \binits{A.}},
\bauthor{\bsnm{Rosia}, \binits{L.}},
\bauthor{\bsnm{Iacobescu}, \binits{C.}}, \betal:
\bctitle{Liro: Benchmark and leaderboard for romanian language tasks}.
In: \bbtitle{Thirty-fifth Conference on Neural Information Processing Systems Datasets and Benchmarks Track (Round 1)}
(\byear{2021})
\end{bchapter}
\endbibitem

%%% 116
\bibitem[\protect\citeauthoryear{Qi et~al.}{2018}]{qi2018and}
\begin{botherref}
\oauthor{\bsnm{Qi}, \binits{Y.}},
\oauthor{\bsnm{Sachan}, \binits{D.S.}},
\oauthor{\bsnm{Felix}, \binits{M.}},
\oauthor{\bsnm{Padmanabhan}, \binits{S.J.}},
\oauthor{\bsnm{Neubig}, \binits{G.}}:
When and why are pre-trained word embeddings useful for neural machine translation?
arXiv preprint arXiv:1804.06323
(2018)
\end{botherref}
\endbibitem

%%% 117
\bibitem[\protect\citeauthoryear{Wolfson et~al.}{2020}]{wolfson2020break}
\begin{barticle}
\bauthor{\bsnm{Wolfson}, \binits{T.}},
\bauthor{\bsnm{Geva}, \binits{M.}},
\bauthor{\bsnm{Gupta}, \binits{A.}},
\bauthor{\bsnm{Gardner}, \binits{M.}},
\bauthor{\bsnm{Goldberg}, \binits{Y.}},
\bauthor{\bsnm{Deutch}, \binits{D.}},
\bauthor{\bsnm{Berant}, \binits{J.}}:
\batitle{Break it down: A question understanding benchmark}.
\bjtitle{Transactions of the Association for Computational Linguistics}
\bvolume{8},
\bfpage{183}--\blpage{198}
(\byear{2020})
\end{barticle}
\endbibitem

%%% 118
\bibitem[\protect\citeauthoryear{Li et~al.}{2020}]{li2020mtop}
\begin{botherref}
\oauthor{\bsnm{Li}, \binits{H.}},
\oauthor{\bsnm{Arora}, \binits{A.}},
\oauthor{\bsnm{Chen}, \binits{S.}},
\oauthor{\bsnm{Gupta}, \binits{A.}},
\oauthor{\bsnm{Gupta}, \binits{S.}},
\oauthor{\bsnm{Mehdad}, \binits{Y.}}:
Mtop: A comprehensive multilingual task-oriented semantic parsing benchmark.
arXiv preprint arXiv:2008.09335
(2020)
\end{botherref}
\endbibitem

%%% 119
\bibitem[\protect\citeauthoryear{Andreas et~al.}{2020}]{andreas2020task}
\begin{barticle}
\bauthor{\bsnm{Andreas}, \binits{J.}},
\bauthor{\bsnm{Bufe}, \binits{J.}},
\bauthor{\bsnm{Burkett}, \binits{D.}},
\bauthor{\bsnm{Chen}, \binits{C.}},
\bauthor{\bsnm{Clausman}, \binits{J.}},
\bauthor{\bsnm{Crawford}, \binits{J.}},
\bauthor{\bsnm{Crim}, \binits{K.}},
\bauthor{\bsnm{DeLoach}, \binits{J.}},
\bauthor{\bsnm{Dorner}, \binits{L.}},
\bauthor{\bsnm{Eisner}, \binits{J.}}, \betal:
\batitle{Task-oriented dialogue as dataflow synthesis}.
\bjtitle{Transactions of the Association for Computational Linguistics}
\bvolume{8},
\bfpage{556}--\blpage{571}
(\byear{2020})
\end{barticle}
\endbibitem

%%% 120
\bibitem[\protect\citeauthoryear{So{\u{g}}anc{\i}o{\u{g}}lu et~al.}{2017}]{souganciouglu2017biosses}
\begin{barticle}
\bauthor{\bsnm{So{\u{g}}anc{\i}o{\u{g}}lu}, \binits{G.}},
\bauthor{\bsnm{{\"O}zt{\"u}rk}, \binits{H.}},
\bauthor{\bsnm{{\"O}zg{\"u}r}, \binits{A.}}:
\batitle{Biosses: a semantic sentence similarity estimation system for the biomedical domain}.
\bjtitle{Bioinformatics}
\bvolume{33}(\bissue{14}),
\bfpage{49}--\blpage{58}
(\byear{2017})
\end{barticle}
\endbibitem

%%% 121
\bibitem[\protect\citeauthoryear{Cer et~al.}{2017}]{cer2017semeval}
\begin{botherref}
\oauthor{\bsnm{Cer}, \binits{D.}},
\oauthor{\bsnm{Diab}, \binits{M.}},
\oauthor{\bsnm{Agirre}, \binits{E.}},
\oauthor{\bsnm{Lopez-Gazpio}, \binits{I.}},
\oauthor{\bsnm{Specia}, \binits{L.}}:
Semeval-2017 task 1: Semantic textual similarity-multilingual and cross-lingual focused evaluation.
arXiv preprint arXiv:1708.00055
(2017)
\end{botherref}
\endbibitem

%%% 122
\bibitem[\protect\citeauthoryear{Dolan and Brockett}{2005}]{dolan2005automatically}
\begin{bchapter}
\bauthor{\bsnm{Dolan}, \binits{B.}},
\bauthor{\bsnm{Brockett}, \binits{C.}}:
\bctitle{Automatically constructing a corpus of sentential paraphrases}.
In: \bbtitle{Third International Workshop on Paraphrasing (IWP2005)}
(\byear{2005})
\end{bchapter}
\endbibitem

%%% 123
\bibitem[\protect\citeauthoryear{Quora}{2017}]{quora2017questionpairs}
\begin{botherref}
\oauthor{\bsnm{Quora}}:
First Quora Dataset Release: Question Pairs.
Accessed: 2025-02-11
(2017).
\url{https://quoradata.quora.com/First-Quora-Dataset-Release-Question-Pairs}
\end{botherref}
\endbibitem

%%% 124
\bibitem[\protect\citeauthoryear{Zhang et~al.}{2019}]{zhang2019paws}
\begin{botherref}
\oauthor{\bsnm{Zhang}, \binits{Y.}},
\oauthor{\bsnm{Baldridge}, \binits{J.}},
\oauthor{\bsnm{He}, \binits{L.}}:
Paws: Paraphrase adversaries from word scrambling.
arXiv preprint arXiv:1904.01130
(2019)
\end{botherref}
\endbibitem

%%% 125
\bibitem[\protect\citeauthoryear{Liu et~al.}{2018}]{liu2018lcqmc}
\begin{bchapter}
\bauthor{\bsnm{Liu}, \binits{X.}},
\bauthor{\bsnm{Chen}, \binits{Q.}},
\bauthor{\bsnm{Deng}, \binits{C.}},
\bauthor{\bsnm{Zeng}, \binits{H.}},
\bauthor{\bsnm{Chen}, \binits{J.}},
\bauthor{\bsnm{Li}, \binits{D.}},
\bauthor{\bsnm{Tang}, \binits{B.}}:
\bctitle{Lcqmc: A large-scale chinese question matching corpus}.
In: \bbtitle{Proceedings of the 27th International Conference on Computational Linguistics},
pp. \bfpage{1952}--\blpage{1962}
(\byear{2018})
\end{bchapter}
\endbibitem

%%% 126
\bibitem[\protect\citeauthoryear{Sang and De~Meulder}{2003}]{sang2003introduction}
\begin{botherref}
\oauthor{\bsnm{Sang}, \binits{E.F.}},
\oauthor{\bsnm{De~Meulder}, \binits{F.}}:
Introduction to the conll-2003 shared task: Language-independent named entity recognition.
arXiv preprint cs/0306050
(2003)
\end{botherref}
\endbibitem

%%% 127
\bibitem[\protect\citeauthoryear{Carreras and M{\`a}rquez}{2004}]{carreras-marquez-2004-introduction}
\begin{bchapter}
\bauthor{\bsnm{Carreras}, \binits{X.}},
\bauthor{\bsnm{M{\`a}rquez}, \binits{L.}}:
\bctitle{Introduction to the {C}o{NLL}-2004 shared task: Semantic role labeling}.
In: \bbtitle{Proceedings of the Eighth Conference on Computational Natural Language Learning ({C}o{NLL}-2004) at {HLT}-{NAACL} 2004},
pp. \bfpage{89}--\blpage{97}.
\bpublisher{Association for Computational Linguistics},
\blocation{Boston, Massachusetts, USA}
(\byear{2004}).
\burl{https://aclanthology.org/W04-2412/}
\end{bchapter}
\endbibitem

%%% 128
\bibitem[\protect\citeauthoryear{Weischedel et~al.}{2013}]{weischedel2013ontonotes}
\begin{botherref}
\oauthor{\bsnm{Weischedel}, \binits{R.}}, et al.:
OntoNotes Release 5.0.
Linguistic Data Consortium,
Philadelphia.
LDC2013T19
(2013)
\end{botherref}
\endbibitem

%%% 129
\bibitem[\protect\citeauthoryear{Do{\u{g}}an et~al.}{2014}]{dougan2014ncbi}
\begin{barticle}
\bauthor{\bsnm{Do{\u{g}}an}, \binits{R.I.}},
\bauthor{\bsnm{Leaman}, \binits{R.}},
\bauthor{\bsnm{Lu}, \binits{Z.}}:
\batitle{Ncbi disease corpus: a resource for disease name recognition and concept normalization}.
\bjtitle{Journal of biomedical informatics}
\bvolume{47},
\bfpage{1}--\blpage{10}
(\byear{2014})
\end{barticle}
\endbibitem

%%% 130
\bibitem[\protect\citeauthoryear{Pradhan et~al.}{2012}]{pradhan2012conll}
\begin{bchapter}
\bauthor{\bsnm{Pradhan}, \binits{S.}},
\bauthor{\bsnm{Moschitti}, \binits{A.}},
\bauthor{\bsnm{Xue}, \binits{N.}},
\bauthor{\bsnm{Uryupina}, \binits{O.}},
\bauthor{\bsnm{Zhang}, \binits{Y.}}:
\bctitle{Conll-2012 shared task: Modeling multilingual unrestricted coreference in ontonotes}.
In: \bbtitle{Joint Conference on EMNLP and CoNLL-shared Task},
pp. \bfpage{1}--\blpage{40}
(\byear{2012})
\end{bchapter}
\endbibitem

%%% 131
\bibitem[\protect\citeauthoryear{Carreras and M{\`a}rquez}{2005}]{carreras2005introduction}
\begin{bchapter}
\bauthor{\bsnm{Carreras}, \binits{X.}},
\bauthor{\bsnm{M{\`a}rquez}, \binits{L.}}:
\bctitle{Introduction to the conll-2005 shared task: Semantic role labeling}.
In: \bbtitle{Proceedings of the Ninth Conference on Computational Natural Language Learning (CoNLL-2005)},
pp. \bfpage{152}--\blpage{164}
(\byear{2005})
\end{bchapter}
\endbibitem

%%% 132
\bibitem[\protect\citeauthoryear{Elsahar et~al.}{2018}]{elsahar2018t}
\begin{bchapter}
\bauthor{\bsnm{Elsahar}, \binits{H.}},
\bauthor{\bsnm{Vougiouklis}, \binits{P.}},
\bauthor{\bsnm{Remaci}, \binits{A.}},
\bauthor{\bsnm{Gravier}, \binits{C.}},
\bauthor{\bsnm{Hare}, \binits{J.}},
\bauthor{\bsnm{Laforest}, \binits{F.}},
\bauthor{\bsnm{Simperl}, \binits{E.}}:
\bctitle{T-rex: A large scale alignment of natural language with knowledge base triples}.
In: \bbtitle{Proceedings of the Eleventh International Conference on Language Resources and Evaluation (LREC 2018)}
(\byear{2018})
\end{bchapter}
\endbibitem

%%% 133
\bibitem[\protect\citeauthoryear{Research}{2021}]{google_relation_extraction}
\begin{botherref}
\oauthor{\bsnm{Research}, \binits{G.}}:
Relation Extraction Corpus.
Accessed: 2025-02-11
(2021).
\url{https://github.com/google-research-datasets/relation-extraction-corpus}
\end{botherref}
\endbibitem

%%% 134
\bibitem[\protect\citeauthoryear{Speer et~al.}{2017}]{speer2017conceptnet}
\begin{bchapter}
\bauthor{\bsnm{Speer}, \binits{R.}},
\bauthor{\bsnm{Chin}, \binits{J.}},
\bauthor{\bsnm{Havasi}, \binits{C.}}:
\bctitle{Conceptnet 5.5: An open multilingual graph of general knowledge}.
In: \bbtitle{Proceedings of the AAAI Conference on Artificial Intelligence},
vol. \bseriesno{31}
(\byear{2017})
\end{bchapter}
\endbibitem

%%% 135
\bibitem[\protect\citeauthoryear{Luan et~al.}{2018}]{luan2018multi}
\begin{botherref}
\oauthor{\bsnm{Luan}, \binits{Y.}},
\oauthor{\bsnm{He}, \binits{L.}},
\oauthor{\bsnm{Ostendorf}, \binits{M.}},
\oauthor{\bsnm{Hajishirzi}, \binits{H.}}:
Multi-task identification of entities, relations, and coreference for scientific knowledge graph construction.
arXiv preprint arXiv:1808.09602
(2018)
\end{botherref}
\endbibitem

%%% 136
\bibitem[\protect\citeauthoryear{Levy et~al.}{2017}]{levy2017zero}
\begin{botherref}
\oauthor{\bsnm{Levy}, \binits{O.}},
\oauthor{\bsnm{Seo}, \binits{M.}},
\oauthor{\bsnm{Choi}, \binits{E.}},
\oauthor{\bsnm{Zettlemoyer}, \binits{L.}}:
Zero-shot relation extraction via reading comprehension.
arXiv preprint arXiv:1706.04115
(2017)
\end{botherref}
\endbibitem

%%% 137
\bibitem[\protect\citeauthoryear{Petroni et~al.}{2019}]{petroni2019language}
\begin{botherref}
\oauthor{\bsnm{Petroni}, \binits{F.}},
\oauthor{\bsnm{Rockt{\"a}schel}, \binits{T.}},
\oauthor{\bsnm{Lewis}, \binits{P.}},
\oauthor{\bsnm{Bakhtin}, \binits{A.}},
\oauthor{\bsnm{Wu}, \binits{Y.}},
\oauthor{\bsnm{Miller}, \binits{A.H.}},
\oauthor{\bsnm{Riedel}, \binits{S.}}:
Language models as knowledge bases?
arXiv preprint arXiv:1909.01066
(2019)
\end{botherref}
\endbibitem

%%% 138
\bibitem[\protect\citeauthoryear{Talmor et~al.}{2018}]{talmor2018commonsenseqa}
\begin{botherref}
\oauthor{\bsnm{Talmor}, \binits{A.}},
\oauthor{\bsnm{Herzig}, \binits{J.}},
\oauthor{\bsnm{Lourie}, \binits{N.}},
\oauthor{\bsnm{Berant}, \binits{J.}}:
Commonsenseqa: A question answering challenge targeting commonsense knowledge.
arXiv preprint arXiv:1811.00937
(2018)
\end{botherref}
\endbibitem

%%% 139
\bibitem[\protect\citeauthoryear{Geva et~al.}{2021}]{geva2021did}
\begin{barticle}
\bauthor{\bsnm{Geva}, \binits{M.}},
\bauthor{\bsnm{Khashabi}, \binits{D.}},
\bauthor{\bsnm{Segal}, \binits{E.}},
\bauthor{\bsnm{Khot}, \binits{T.}},
\bauthor{\bsnm{Roth}, \binits{D.}},
\bauthor{\bsnm{Berant}, \binits{J.}}:
\batitle{Did aristotle use a laptop? a question answering benchmark with implicit reasoning strategies}.
\bjtitle{Transactions of the Association for Computational Linguistics}
\bvolume{9},
\bfpage{346}--\blpage{361}
(\byear{2021})
\end{barticle}
\endbibitem

%%% 140
\bibitem[\protect\citeauthoryear{Wei et~al.}{2022}]{wei2022chain}
\begin{barticle}
\bauthor{\bsnm{Wei}, \binits{J.}},
\bauthor{\bsnm{Wang}, \binits{X.}},
\bauthor{\bsnm{Schuurmans}, \binits{D.}},
\bauthor{\bsnm{Bosma}, \binits{M.}},
\bauthor{\bsnm{Xia}, \binits{F.}},
\bauthor{\bsnm{Chi}, \binits{E.}},
\bauthor{\bsnm{Le}, \binits{Q.V.}},
\bauthor{\bsnm{Zhou}, \binits{D.}}, \betal:
\batitle{Chain-of-thought prompting elicits reasoning in large language models}.
\bjtitle{Advances in neural information processing systems}
\bvolume{35},
\bfpage{24824}--\blpage{24837}
(\byear{2022})
\end{barticle}
\endbibitem

%%% 141
\bibitem[\protect\citeauthoryear{Cobbe et~al.}{2021}]{cobbe2021training}
\begin{botherref}
\oauthor{\bsnm{Cobbe}, \binits{K.}},
\oauthor{\bsnm{Kosaraju}, \binits{V.}},
\oauthor{\bsnm{Bavarian}, \binits{M.}},
\oauthor{\bsnm{Chen}, \binits{M.}},
\oauthor{\bsnm{Jun}, \binits{H.}},
\oauthor{\bsnm{Kaiser}, \binits{L.}},
\oauthor{\bsnm{Plappert}, \binits{M.}},
\oauthor{\bsnm{Tworek}, \binits{J.}},
\oauthor{\bsnm{Hilton}, \binits{J.}},
\oauthor{\bsnm{Nakano}, \binits{R.}}, et al.:
Training verifiers to solve math word problems.
arXiv preprint arXiv:2110.14168
(2021)
\end{botherref}
\endbibitem

%%% 142
\bibitem[\protect\citeauthoryear{Roy and Roth}{2016}]{roy2016solving}
\begin{botherref}
\oauthor{\bsnm{Roy}, \binits{S.}},
\oauthor{\bsnm{Roth}, \binits{D.}}:
Solving general arithmetic word problems.
arXiv preprint arXiv:1608.01413
(2016)
\end{botherref}
\endbibitem

%%% 143
\bibitem[\protect\citeauthoryear{Patel et~al.}{2021}]{patel2021nlp}
\begin{botherref}
\oauthor{\bsnm{Patel}, \binits{A.}},
\oauthor{\bsnm{Bhattamishra}, \binits{S.}},
\oauthor{\bsnm{Goyal}, \binits{N.}}:
Are nlp models really able to solve simple math word problems?
arXiv preprint arXiv:2103.07191
(2021)
\end{botherref}
\endbibitem

%%% 144
\bibitem[\protect\citeauthoryear{Ling et~al.}{2017}]{ling2017program}
\begin{botherref}
\oauthor{\bsnm{Ling}, \binits{W.}},
\oauthor{\bsnm{Yogatama}, \binits{D.}},
\oauthor{\bsnm{Dyer}, \binits{C.}},
\oauthor{\bsnm{Blunsom}, \binits{P.}}:
Program induction by rationale generation: Learning to solve and explain algebraic word problems.
arXiv preprint arXiv:1705.04146
(2017)
\end{botherref}
\endbibitem

%%% 145
\bibitem[\protect\citeauthoryear{Hosseini et~al.}{2014}]{hosseini2014learning}
\begin{bchapter}
\bauthor{\bsnm{Hosseini}, \binits{M.J.}},
\bauthor{\bsnm{Hajishirzi}, \binits{H.}},
\bauthor{\bsnm{Etzioni}, \binits{O.}},
\bauthor{\bsnm{Kushman}, \binits{N.}}:
\bctitle{Learning to solve arithmetic word problems with verb categorization}.
In: \bbtitle{Proceedings of the 2014 Conference on Empirical Methods in Natural Language Processing (EMNLP)},
pp. \bfpage{523}--\blpage{533}
(\byear{2014})
\end{bchapter}
\endbibitem

%%% 146
\bibitem[\protect\citeauthoryear{Koncel-Kedziorski et~al.}{2015}]{koncel2015parsing}
\begin{barticle}
\bauthor{\bsnm{Koncel-Kedziorski}, \binits{R.}},
\bauthor{\bsnm{Hajishirzi}, \binits{H.}},
\bauthor{\bsnm{Sabharwal}, \binits{A.}},
\bauthor{\bsnm{Etzioni}, \binits{O.}},
\bauthor{\bsnm{Ang}, \binits{S.D.}}:
\batitle{Parsing algebraic word problems into equations}.
\bjtitle{Transactions of the Association for Computational Linguistics}
\bvolume{3},
\bfpage{585}--\blpage{597}
(\byear{2015})
\end{barticle}
\endbibitem

%%% 147
\bibitem[\protect\citeauthoryear{Hendrycks et~al.}{2021}]{hendrycks2021measuring}
\begin{botherref}
\oauthor{\bsnm{Hendrycks}, \binits{D.}},
\oauthor{\bsnm{Burns}, \binits{C.}},
\oauthor{\bsnm{Kadavath}, \binits{S.}},
\oauthor{\bsnm{Arora}, \binits{A.}},
\oauthor{\bsnm{Basart}, \binits{S.}},
\oauthor{\bsnm{Tang}, \binits{E.}},
\oauthor{\bsnm{Song}, \binits{D.}},
\oauthor{\bsnm{Steinhardt}, \binits{J.}}:
Measuring mathematical problem solving with the math dataset.
arXiv preprint arXiv:2103.03874
(2021)
\end{botherref}
\endbibitem

%%% 148
\bibitem[\protect\citeauthoryear{Hsieh et~al.}{2023}]{hsieh2023automatic}
\begin{botherref}
\oauthor{\bsnm{Hsieh}, \binits{C.-J.}},
\oauthor{\bsnm{Si}, \binits{S.}},
\oauthor{\bsnm{Yu}, \binits{F.X.}},
\oauthor{\bsnm{Dhillon}, \binits{I.S.}}:
Automatic engineering of long prompts.
arXiv preprint arXiv:2311.10117
(2023)
\end{botherref}
\endbibitem

%%% 149
\bibitem[\protect\citeauthoryear{Ju et~al.}{2023}]{ju2023continuous}
\begin{bchapter}
\bauthor{\bsnm{Ju}, \binits{T.}},
\bauthor{\bsnm{Zheng}, \binits{Y.}},
\bauthor{\bsnm{Wang}, \binits{H.}},
\bauthor{\bsnm{Zhao}, \binits{H.}},
\bauthor{\bsnm{Liu}, \binits{G.}}:
\bctitle{Is continuous prompt a combination of discrete prompts? towards a novel view for interpreting continuous prompts}.
In: \bbtitle{Findings of the Association for Computational Linguistics: ACL 2023},
pp. \bfpage{7804}--\blpage{7819}
(\byear{2023})
\end{bchapter}
\endbibitem

\end{thebibliography}

\end{document}